\pdfoutput=1

\documentclass[11pt]{article}

\usepackage[final]{acl}

\usepackage{times}
\usepackage{latexsym}

\usepackage[T1]{fontenc}

\usepackage[utf8]{inputenc}

\usepackage{microtype}

\usepackage{inconsolata}

\usepackage{hyperref}
\usepackage{url}
\usepackage{graphicx}
\usepackage{wrapfig}
\usepackage{subcaption}
\usepackage{float}
\usepackage{amsfonts}
\usepackage{comment}
\usepackage{amsmath}
\usepackage{booktabs}
\usepackage{siunitx}
\usepackage{amssymb}
\usepackage{graphicx}
\usepackage{dsfont}
\usepackage{multirow}
\usepackage{dcolumn}
\usepackage{algorithm}
\usepackage{algpseudocode}
\usepackage{mathtools}
\usepackage{soul}
\usepackage{enumitem}

\usepackage{amsmath,amsfonts,bm}

\def\eqref#1{equation~\ref{#1}}

\def\1{\bm{1}}

\def\vtheta{{\bm{\theta}}}

\def\vl{{\bm{l}}}
\def\vm{{\bm{m}}}

\def\vs{{\bm{s}}}

\DeclareMathAlphabet{\mathsfit}{\encodingdefault}{\sfdefault}{m}{sl}
\SetMathAlphabet{\mathsfit}{bold}{\encodingdefault}{\sfdefault}{bx}{n}

\def\gU{{\mathcal{U}}}

\def\sR{{\mathbb{R}}}

\newcommand{\sigmoid}{\sigma}

\newcommand{\Loss}{\mathcal{L}}
\newcommand{\ie}{\textit{i.e.}}
\newcommand{\eg}{\textit{e.g.}}
\newcommand{\indic}{\mathds{1}}

\newcommand{\mask}{\vm}
\newcommand{\inversemask}{\tilde{\mask}}
\newcommand{\score}{\vs}
\newcommand{\logit}{\vl}
\newcommand{\param}{\vtheta}
\newcommand{\plm}{f(x, \param)}
\newcommand{\subnetwork}{f(x, \mask \odot \param)}
\newcommand{\inverse}{f(x, \inversemask \odot \param)}

\newcommand{\plmprob}{f(x, \param)}
\newcommand{\subnetworkprob}{f(x, \mask \odot \param)}
\newcommand{\inverseprob}{f(x, \inversemask \odot \param)}

\newcommand{\allgraph}{K}
\newcommand{\targetgraph}{K_{T}}
\newcommand{\controlgraph}{K_{C}}
\newcommand{\lmdataset}{D_{LM}}

\newcommand{\targetkg}{\textsc{TargetKG}}
\newcommand{\controlkg}{\textsc{ControlKG}}
\newcommand{\lmodeling}{\textsc{ControlLM}}

\newcommand{\forgetting}{\textbf{suppression}}
\newcommand{\maintenancekg}{\textbf{maintenance-KG}}
\newcommand{\maintenancelm}{\textbf{maintenance-LM}}
\newcommand{\sparsity}{\textbf{sparsity}}

\newcommand{\deltappl}{$\Delta$PPL}
\newcommand{\deltarank}{$\Delta$Rank}

\algnewcommand{\algorithmicand}{\textbf{ and }}
\algnewcommand{\algorithmicor}{\textbf{ or }}
\algnewcommand{\OR}{\algorithmicor}
\algnewcommand{\AND}{\algorithmicand}
\algnewcommand{\var}{\texttt}
\algnewcommand\algorithmicforeach{\textbf{for each}}
\algdef{S}[FOR]{ForEach}[1]{\algorithmicforeach\ #1\ \algorithmicdo}

\newcommand*{\thead}[1]{\multicolumn{1}{c}{\bfseries #1}}

\DeclarePairedDelimiterX{\infdivx}[2]{(}{)}{%
  #1\;\delimsize\|\;#2%
}
\newcommand{\kldiv}{D_\text{KL}\infdivx}

\title{Discovering Knowledge-Critical Subnetworks\\in Pretrained Language Models}

\author{
  Deniz Bayazit, Negar Foroutan, Zeming Chen, Gail Weiss, Antoine Bosselut  \\
  EPFL\\
  \texttt{\{deniz.bayazit,antoine.bosselut\}@epfl.ch}
}

\begin{document}
\maketitle
\begin{abstract}
Pretrained language models (LMs) encode implicit representations of knowledge in their parameters. However, localizing these representations and disentangling them from each other remains an open problem. In this work, we investigate whether pretrained language models contain various \textit{knowledge-critical} subnetworks: particular sparse computational subgraphs that can, if removed, precisely suppress specific knowledge the model has memorized. We propose a multi-objective differentiable masking scheme that can be applied to both weights and neurons to discover such subnetworks and show that we can use them to precisely remove specific knowledge from models while minimizing adverse effects on the behavior of the original model. We demonstrate our method on multiple GPT2 variants, uncovering highly sparse subnetworks~(98\%+ sparsity) that are critical for expressing specific collections of relational knowledge. When these subnetworks are removed, the remaining network maintains most of its initial abilities but struggles to represent the suppressed knowledge.\footnote{The code is made available at \url{https://github.com/bayazitdeniz/know-subnet}}
\end{abstract}

\section{Introduction}

\begin{figure}[t]
\centering
\includegraphics[width=0.95\linewidth]{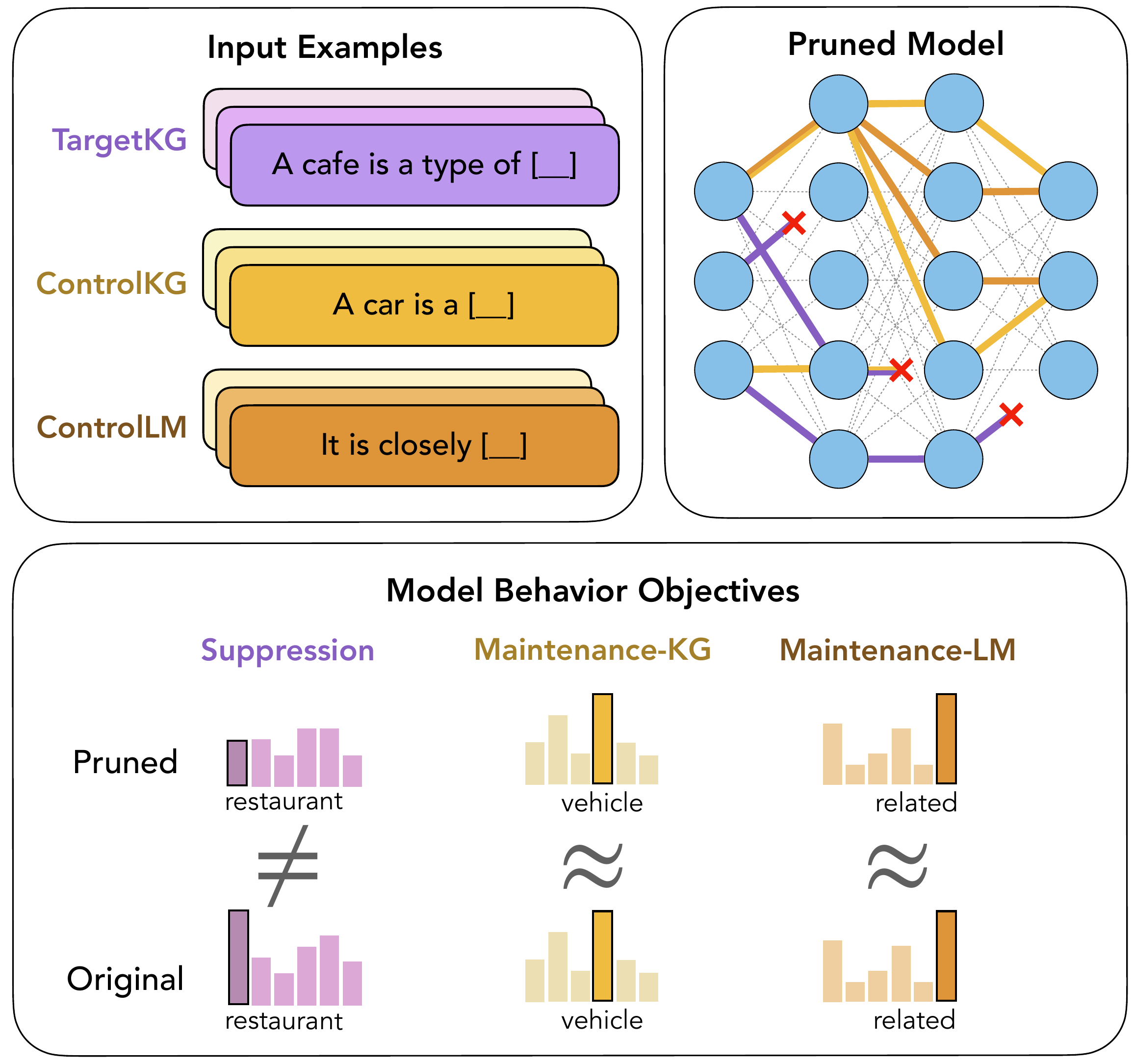}
\caption{\textit{Knowledge-critical} subnetworks are necessary for expressing target knowledge triplets (\targetkg{}) in LMs. When removed, the remaining model no longer expresses the specific triplets, but maintains its ability to express other relational knowledge (\controlkg{}) and its language modeling abilities (\lmodeling{}).}
\label{fig:first-fig}
\end{figure}

Large-scale language models (LLMs) encode large amounts of relational knowledge \cite{petroni-etal-2019-language, carlini2023quantifying, liu2021prompt}, which they transfer to successfully adapt to downstream tasks \cite{wang2019glue, wang2019superglue}. Following this success, considerable research focuses on better understanding the extent to which LLMs capture this knowledge \cite{liu-etal-2019-linguistic, safavi-koutra-2021-relational, da2021analyzing, huang2022language}. In these works, relational triplets (\eg{}, \texttt{(car, IsA, vehicle)}) are converted to natural language (\eg{}, \texttt{``A car is a vehicle.''}) before being presented to a model. Key tokens in these input sequences are masked, and the model demonstrates its knowledge of the relations by recovering these tokens.

With the body of work studying LLMs as knowledge bases, a subset of works focuses on \textit{where} and \textit{how} this knowledge may be encoded by the models that capture it. The answer to these questions could potentially facilitate the development of more effective finetuning methods, which can be useful for rectifying factual errors made by language models, updating models with evolving knowledge, and preventing ethically undesirable behavior. 

Considerable work in model probing \cite{belinkov-glass-2019-analysis, durrani-etal-2020-analyzing, antverg-ben-david-and-yonatan-belinkov-2022-idani, belinkov-2022-probing} and mechanistic interpretability \cite{geva-etal-2021-transformer,geva-etal-2022-transformer,geva-etal-2022-lm} explores these questions, discovering hidden representations, neurons, and layers that are responsible for the expression of knowledge from these systems. However, these works typically do not localize the knowledge accessing behavior to individual parameters.
Another line of work in model editing explores whether knowledge in the model can be changed \cite{de-cao-etal-2021-editing, dai-etal-2022-knowledge, hase-etal-2023-methods, mitchell-2022-fast, mitchell2022memory, meng2022locating, meng2023massediting, hase-2023-localizationediting, gupta-etal-2023-editing, jang-etal-2023-knowledge, chen-2023-reckoning}. However, the goal of these methods is also typically not to precisely localize the parameters responsible for expressing knowledge, but instead to broadly edit model parameters such that a new desired behavior overwrites the model's preference for the old one.

In this work, we hypothesize that any piece of relational knowledge expressed by a language model is encoded by a limited subset of its parameters. We search for these parameters by identifying sparse subnetworks that, when removed, suppress the model's ability to express the knowledge of interest while not affecting other abilities of the model. As the model cannot express target knowledge without these subnetworks, we refer to them as \textit{knowledge-critical}. In Figure~\ref{fig:first-fig}, we illustrate this concept -- when the weights marked with a red cross are removed from the original network, the expression of the triplet \texttt{(cafe, IsA, restaurant)} is suppressed, whereas other triplets are not.

To discover knowledge-critical subnetworks, we propose training differentiable masks over weights or neurons of the original pretrained model, such that the mask can identify and remove a knowledge-critical subnetwork for the targeted knowledge graph. Specifically, we train the mask to: (1) suppress the expression of the target knowledge triplets, (2) maintain the ability to express generic relational knowledge and language, and (3) remove only a minimal subset of weights. After training, the remaining pruned model can no longer express the target knowledge, but maintains its performance on other behaviors, thereby identifying the \textit{knowledge-critical} subnetwork as the masked portion of the original model.

Our results --- across multiple target knowledge graphs (constructed from WordNet and ConceptNet) and LLMs at multiple scales (from the family of GPT2 models) --- show that weight masking consistently identifies sparse subnetworks (an average sparsity of $\sim$98.6\%) that satisfy our objectives. When these subnetworks are removed, the remaining model's perplexity on the target knowledge associated with the subnetwork largely increases (an average relative perplexity increase of 253\% - 5589\% for different GPT2 models), indicating that the expression of the target knowledge is successfully suppressed. However, the remaining network's ability to model generic relational knowledge and natural language negligibly changes. Finally, in a study on CommonsenseQA, we show that once these subnetworks are removed, models finetuned using parameter-efficient methods struggle with questions that require the knowledge encoded by the removed subnetworks.
\section{Related Work}

\paragraph{LLMs as Knowledge Bases}
Our work builds on prior research that demonstrates the knowledge memorization abilities of large language models (LLMs; \citealp{carlini2021extracting, alkhamissi2022review}). Multiple studies have shown that LLMs encode various types of knowledge \cite{liu-etal-2019-linguistic, chen2022linguistic, safavi-koutra-2021-relational, huang2022language}.
In these works, parametric knowledge in LLMs is typically expressed by conditioning on a natural language context to complete or infill a sequence that expresses the knowledge \cite{petroni-etal-2019-language, jiang-etal-2020-know, shin-etal-2020-autoprompt, cao-etal-2021-knowledgeable, zhong-etal-2021-factual, qin-eisner-2021-learning, liu2021prompt, yu2023generate}. Other methods also fine-tune models to create an interface to parametric knowledge \cite{bosselut-etal-2019-comet, roberts-etal-2020-much, jiang-etal-2021-im, hwang2021comet}. In contrast, our work investigates \textit{where} knowledge is encoded by LLMs and localizes the critical subnetworks for expressing these facts.

\vspace{-5pt}
\paragraph{Function-Specific Subnetworks}
Methodologically, our work draws inspiration from studies that identify task-specific subnetworks in neural networks. Perhaps most known, \citet{frankle-2018-lotteryticket} propose the \emph{Lottery Ticket Hypothesis}, showing that learned subnetworks could achieve test accuracy similar to that of original networks. Other works prune subnetworks for the purpose of efficient finetuning \cite{mallya-2018-piggyback, zhao-etal-2020-masking, sanh-2020-movementpruning, guo-etal-2021-parameter}, or identifying function-specific subnetworks \cite{cao-etal-2021-low,sanh-2020-movementpruning,zhang-etal-2021-disentangling, csordas-2021-modular}. Identifying function-specific subnetworks also leads to useful applications, such as disentangling representations to reduce model susceptibility to spurious correlations \cite{zhang-etal-2021-disentangling}, probing models for linguistic properties \cite{cao-etal-2021-low, de-cao-etal-2020-decisions}, identifying and removing a toxic behavior or bias \cite{li2023circuit, chintam-etal-2023-identifying}, and finding subnetworks specialized for different languages \cite{foroutan-etal-2022-discovering}. Most similar to our work is that of \citet{ren-zhu-2022-specializing}, which learns coarse subnetworks that encoded large portions of ConceptNet. We also adopt a differentiable weight masking scheme, but use it to identify highly sparse subnetworks critical for particular expressions of knowledge.

\vspace{-5pt}
\paragraph{Mechanistic Interpretability}
Mechanistic interpretability tackles the problem of understanding model behavior by reverse-engineering computations performed by transformer models. \citet{elhage2021mathematical} discovered algorithmic patterns and frameworks in simplified transformer models. Following this, researchers discovered induction heads \cite{olsson2022incontext}, \ie{}, specific attention heads involved in in-context learning in LLMs. Similarly, with interventions on attention and MLP sublayers, \citet{geva-etal-2023-dissecting} identified critical points where the model propagates information, as well as the internal mechanism for attribute extraction.
Other work focuses on knowledge tracing and localization in model parameters for the goal of model editing \cite{dai-etal-2022-knowledge, meng2022locating, meng2023massediting, gupta-etal-2023-editing, hernandez2023remedi}. Activation patching with corrupted tokens \cite{meng2022locating} or corrupted prompts \cite{wang2023interpretability} use causal intervention to identify activations responsible for flipping the model's output. In contrast, our work focuses on preserving the original model to precisely locate individual model \textit{weights} responsible for expressing a given set of target knowledge without counterfactuals.
Our work is closer to path patching \cite{goldowskydill2023localizing} and automatic circuit discovery \cite{conmy2023automated} to localize behaviors in network subgraphs but focuses specifically on identifying subnetworks associated with knowledge relationships. 
Our work is also similar to \citet{lo-etal-2024-large}, which shows that models can re-learn removed concepts via neurons. In contrast, we focus on individual parameter pruning.

\section{Background \& Considerations}
\label{sec:background}
To find a knowledge-critical subnetwork in a pretrained language model, we learn a differentiable parameter mask (\S\ref{sec:methodology}) using a prediction task where the LM is prompted for relational knowledge.

\paragraph{Prompting LMs with KGs}
We define a global relational knowledge graph (KG) as the set of knowledge triplets, 
\[
\allgraph = \{(h_1, r_1, t_1), ... (h_n, r_n, t_n)\}
\] 
where $h$ and $t$ are head and tail entity nodes, respectively, and $r$ is the relation that holds between the two entities. 
To input relational knowledge to an LM, triplets are verbalized using a natural language template. For example, the triplet \texttt{(house, IsA, building)}, can be verbalized with the template \texttt{``\{article\} \{$h$\} is \{article\} \{$t$\}''} as \texttt{``A house is a building.''} 
A typical way to prompt for knowledge is to mask the tail entity \texttt{``A house is a \_\_\_''} \cite{petroni-etal-2019-language}. To approximate an autoregressive model's confidence in a given triplet, we compute a distribution over the missing token and calculate the perplexity of the model for the correct token \texttt{building}.

\paragraph{Differentiable Weight Masking for Function-Specific Parameter Search}
To localize parameters that are critical for modeling specific knowledge, we learn a binary mask over each network parameter. For a language model $\plm{}$ with pretrained parameters $\param$ that takes as input $x$, we learn a set of binary parameters $\mask \in \{0, 1\}^{|\param|}$ that is element-wise multiplied with the frozen $\param$, such that our subnetwork is formulated as $\subnetwork{}$. Similar to other binary mask learning methods \cite{cao-etal-2021-low,sanh-2020-movementpruning}, our method models each parameter mask $\mask_i$ with the concrete (\ie{}, Gumbel-Softmax) distribution, a differentiable approach to learn continuous mask scores $\score_i \in [0, 1]$ from real-valued parameters $\logit_i \in \sR$ \cite{maddison-2016-concrete,jang-2017-CatergoricalReparameterization}:
\begin{equation}
    \score_i = \sigmoid ((\logit_i - \log ( \log \gU_1 / \log \gU_2 )) / \tau ) \label{eq:gumbel}
\end{equation}
\noindent where $\gU_1, \gU_2 \sim \gU(0,1)$ and $\sigmoid$ is a sigmoid function.
We use the approach of \citet{csordas-2021-modular}, which uses a straight-through estimator that thresholds the continuous score \cite{bengio-2013-EstimatingPropagating}:
\begin{equation}
    \mask_i = [\indic_{\score_i > 0.5} - \score_i]_{\text{detach}} + \score_i \label{eq:st_estimator}
\end{equation}  
\noindent where $\indic$ is an indicator function that thresholds the scores at 0.5 and $[]_{\text{detach}}$ is an operation that prevents back-propagation. This way, we back-propagate through the non-detached continuous mask scores $\score_i$ and still calculate loss with the overall binarized mask score $\mask_i$. 

\paragraph{Mask Granularity} Discovering subnetworks requires selecting the granularity of the parameter mask, reflecting the granularity at which we hypothesize separable knowledge representations can be discovered in the model. Most prior work selects neurons \citep{elhage2022superposition} or layers \citep{zhou-etal-2023-rome} as the basic structural unit for localizing model behaviors. While these representations have been shown to encode knowledge behaviors \citep{dai-etal-2022-knowledge,lo-etal-2024-large}, they are perhaps too broad for reliably disentangling specific knowledge, as they are typically polysemantic (\ie{}, they jointly encode multiple behaviors; \citealp{olah2020zoom}). Conversely, localizing knowledge representations as an unconstrained combination of individual parameters is likely more separable, but may be noisy, as many parameters may be largely redundant, and individual parameters may suffer from overfitting. With no clear choice, in this work, we explore both parameter-level and neuron-level masking to provide complementary insights for mechanistic knowledge localization.
 
\section{Methodology}
\label{sec:methodology}
This section defines our methodology for discovering \textit{knowledge-critical} subnetworks using differentiable weight or neuron masking.

\vspace{-5pt}
\paragraph{Notation} We define a subnetwork as in \S\ref{sec:background}: $\subnetwork$, where $\param$ is the set of parameters of the network $f$ and $\mask$ is the mask over a portion of that network's parameters. {To learn a mask over neurons, we jointly mask all the weights connecting to the same neuron.} We assume a target set of knowledge  $\targetgraph \subset \allgraph$ (\targetkg{}) for which we want to identify the critical parameters. 

\subsection{Knowledge-Critical Subnetworks}
Our goal is to find \textit{knowledge-critical} subnetworks: the essential parameters to express a given set of target knowledge. When knowledge-critical subnetworks are removed, the expression of the target triplets should be suppressed, and the expression of irrelevant triplets should be unaffected.

\paragraph{Suppression}
For $\subnetwork$ to be \textit{critical} in expressing $\targetgraph$, its removal from the original network should also suppress the model's ability to express the knowledge in $\targetgraph$. More formally, the inversely masked subnetwork (\ie{}, remaining model), $\inverse$, where $\inversemask = 1 - \mask$, should have difficulty expressing $\targetgraph$. We define this as the \forgetting{} criterion, as it encourages that the remaining model cannot represent knowledge in $\targetgraph$. If we find such a disentanglement, we consider that the pretrained model heavily relies on the removed subnetwork to perform a task related to $\targetgraph$.

\paragraph{Maintenance}
However, if only optimized for \forgetting{}, our method may discover subnetworks that are \textit{critical} to all expressions of knowledge, or all expressions of coherent sequences of language. As the model should retain its initial capacities, we also define \textbf{maintenance} criteria for \textit{knowledge-critical} subnetworks. They should: (1) not affect the model's ability to express other relational knowledge $\controlgraph = \allgraph \setminus \targetgraph$ (\controlkg{}), and (2) not affect the model's original language modeling abilities (\lmodeling{}).
These criteria are referred to as \maintenancekg{} and \maintenancelm{}, respectively.

\paragraph{Sparsity}
Finally, we aim to keep the knowledge-critical subnetwork as sparse as possible to discover the parameters that predominantly encode the expression of $\targetgraph$. Without imposing a high sparsity level, parameters unrelated to the expression of $\targetgraph$ or $\controlgraph$ might persist within the subnetwork.

\begin{table*}[th!]
\centering
\resizebox{0.75\linewidth}{!}{
    \begin{tabular}{clcccccccc}
    \toprule
    \multicolumn{2}{c}{\multirow{2}{*}{\textbf{Knowledge Graph}}}  & \multirow{2}{*}{\textbf{\# triplets}} & \multirow{2}{*}{\textbf{\# heads}}  & \multirow{2}{*}{\textbf{\# tails}} & \multirow{2}{*}{\textbf{\# rels}}  & \multicolumn{4}{c}{\textbf{GPT-2 PPL}}    \\ 
     & & & & & & \textbf{Small} & \textbf{Med} & \textbf{Large} & \textbf{XL} \\ \midrule
    \multirow{9}{*}{WordNet} & \controlkg{} train      &   9751       &    9707    &   2709    &   1 & 63.6 & 32.8 & 27.4 & 24.5 \\
    & \controlkg{} val.       &   50         &      50  &    50      & 1 & 73.2 & 37.5 & 31.3 & 30.8  \\ \cmidrule{2-10}
    & \texttt{building} &      11      &        11    &     11       & 1 & 51.9 & - & - & - \\
    & \texttt{communication} &       16     &         16   &       9      & 1 & 96.3  & 65.2 & 69.2 & 59.8 \\
    & \texttt{change} &      13      &         13   &      13      &  1 & 109.7 & - & - & - \\
    & \texttt{statement}       &      16      &        16    &      16      & 1 & 170.2  & - & - \\
    & \texttt{location}   &      19      &       19     &      7       & 1 &  198.0 & 119.0 & 125.5 & 81.4   \\
    & \texttt{representation}  &      12      &    12        &      12      & 1 &  210.7 & 106.8 & 108.7 & 85.0  \\
    & \texttt{magnitude} &        12    &         12   &       7      & 1 & 299.9 & - & - & -  \\  \midrule
    \multirow{5}{*}{ConceptNet} & \controlkg{} train      &   5455       &    2898    &   2129    &   16 & 373.0 & - & - & -  \\
    & \controlkg{} val.       &   606         &  522      &    482      & 16 &  172.3  & - & - & -  \\ \cmidrule{2-10}
    & \texttt{fruit} & 36 & 11  & 37 & 12 & 381.6 & - & - & - \\
    & \texttt{sun} & 36 & 11  & 36 & 12 & 387.5 & - & - & - \\ 
    & \texttt{swimming} & 40 & 14 & 40 & 15 & 517.8 & - & - & - \\\bottomrule
    \end{tabular}
}
\caption{\textbf{Statistics on sampled KGs and their verbalization.} The graph statistics show the number of triplets and the unique number of heads, tails, and relations. The average perplexity is calculated with the gold tail token cross-entropy loss. The perplexity for certain KGs in the Medium, Large and XL columns are not included as we do not evaluate them in our study on model scale.}
\label{tab:kg-stats}
\vspace{-5pt}
\end{table*}

\subsection{Mask Learning}
\label{sec:loss-terms}
To learn a weight mask for knowledge-critical subnetworks, we define a joint objective that optimizes for the criteria defined above.

\paragraph{Suppression Loss}
To fulfill the \forgetting{} criterion, the remaining model, $\inverse$, should be less confident in the expression of knowledge in $\targetgraph$. We propose to minimize the KL divergence between the remaining model's predicted distribution over possible tail entities of a knowledge triplet and a uniform reference distribution $\mathcal{U_V}$ over the tokens in the model's vocabulary. For $x \in \targetgraph$:
\begin{equation}
    \Loss_{\text{suppress}} = \kldiv{\mathcal{U_V}}{\inverse}
\end{equation}
\paragraph{Maintenance Losses}
As there are multiple ways a model could learn to suppress the expression $\targetgraph$, namely (1) suppressing all knowledge that is in the same format or (2) suppressing all language expressions, we define two regularization objectives.
To encourage the rest of the model to keep its original performance on the control knowledge $\controlgraph$ and a standard language modeling dataset $\lmdataset$, we calculate the KL divergence of $\inverse$ with the pretrained model's distribution $\plm$ as a reference.
Thus, for any $x \in \controlgraph$ or $x \in \lmdataset$:  
\begin{equation}
    \Loss_{\text{maintain}} = \kldiv{\plm}{\inverse}
\end{equation}
\noindent We define two such loss terms, one for each of \maintenancekg{} and \maintenancelm{}.

\paragraph{Sparsity Regularization}
To promote the subnetwork containing only parameters critical for modeling \targetkg{}, we encourage sparsity by minimizing the average subnetwork density (\ie{}, sigmoid of the masking parameters $\logit_i$ from Eq.~\ref{eq:gumbel}):
\begin{equation}
    \Loss_{\text{sparsity}} = \frac{1}{|\param|} \sum\limits_{i=1}^{|\param|}{\sigma(\logit_i)}
\end{equation}
\noindent \textbf{Final Loss}
Our final loss is a mixture of these losses with weights $\lambda_{i}$:
\begin{equation}
    \begin{split}
    \Loss_{\text{final}} =  \lambda_{1} \Loss_{\text{suppress}} + \lambda_{2} \Loss_{\text{maintain-KG}} \\ + \lambda_{3} \Loss_{\text{maintain-LM}} + \lambda_{4} \Loss_{\text{sparsity}} 
    \label{eq:objective}
    \end{split}
\end{equation} 
\section{Experimental Setup}
\label{sec:eval-setup}

\paragraph{Models \& Training}
To test whether our method can scale to various model sizes, we discover knowledge subnetwork masks for GPT2-small, {(117M parameters, 12 layers),}
GPT2-medium, {(345M parameters, 24 layers),}
GPT2-large, {(774M parameters, 36 layers),} 
and GPT2-XL. {(1.5B parameters, 42 layers; \citealp{radford-2019-gpt2})}. 
During mask learning, we do not mask the embedding, language modeling head, layer-normalization, and bias parameters,\footnote{Prior work has not observed an advantage to masking these components for general tasks \cite{zhao-etal-2020-masking}.} and only learn masks for the top 50\% of transformer layers.\footnote{Multiple layer-wise analyses have shown that the first layers of transformer LMs encode low-level linguistic features that may be a prerequisite for knowledge modeling \cite{tenney-etal-2019-bert, liu-etal-2019-linguistic}. We also perform a masked layer choice study that confirms this intuition (Appendix~\ref{sec:layer-study}).} Further implementation details on masking, hyperparameter, and checkpoint selection are in Appendix~\ref{sec:train-implement}.

\paragraph{Datasets} 
To create \targetkg{} and \controlkg{}s, we sample hypernym triplets from WordNet \cite{miller-1995-wordnet}, as well as triplets from the LAMA subset of ConceptNet \cite{speer2017conceptnet, petroni-etal-2019-language}. For simplicity, we only use triplets with single-token tail entities. We sample 7 \targetkg{}s for WordNet, and 3 for ConceptNet (statistics shown in Table~\ref{tab:kg-stats}) by randomly selecting an initial node and sampling knowledge triplets by performing 3-hop random walks in both the parent and child direction of the KG. To create \controlkg{}, we prioritize not leaking \targetkg{} counterfactuals and having a shared \controlkg{} across different \targetkg{}s, and remove from the complete KG any triplet that shares the same entities as the union of the \targetkg{}s shown in Table~\ref{tab:kg-stats}. For all triplets, to suppress and maintain knowledge that the model is already confident about, we select the verbalization for each triplet with the lowest perplexity on the tail token. 
For the \lmodeling{} dataset, we use WikiText-2 \cite{merity-2017-pointer}. We refer to \controlkg{} and \lmodeling{} together as maintenance datasets. All maintenance results are on a held-out validation set. 
Further information on data preprocessing is in Appendices~\ref{sec:data-creation} and \ref{sec:train-implement}.

\begin{table*}[t]
\centering
\resizebox{1.0\linewidth}{!}{
    \begin{tabular}{ccc r@{\hspace{2pt}} r r@{\hspace{2pt}}l r@{\hspace{2pt}}l r@{\hspace{2pt}}l r@{\hspace{2pt}}l}
        \toprule
        \textbf{Knowledge} & \textbf{Mask}   & \textbf{Sparsity}        & \multicolumn{2}{c}{\textbf{\targetkg{}}}                & \multicolumn{2}{c}{\textbf{\controlkg{}}}               & \multicolumn{2}{c}{\textbf{\lmodeling{}}}               & \multicolumn{2}{c}{\textbf{\targetkg{}}}                 & \multicolumn{2}{c}{\textbf{\controlkg{}}} \\
         \textbf{Graph}    & \textbf{Method} & \textbf{($\uparrow$)}    &  \multicolumn{2}{c}{\textbf{\deltappl{} ($\uparrow$)}}  & \multicolumn{2}{c}{\textbf{\deltappl{} ($\downarrow$)}} & \multicolumn{2}{c}{\textbf{\deltappl{} ($\downarrow$)}} & \multicolumn{2}{c}{\textbf{\deltarank{} ($\uparrow$)}}    & \multicolumn{2}{c}{\textbf{\deltarank{} ($\downarrow$)}} \\ \midrule
        \multirow{4}{*}{WordNet} 
        & Weight Masking            & 98.6  & 590.9 & \multirow{4}{*}{(162.4)}  & -0.2 & \multirow{4}{*}{(73.2)}     & 0.5 & \multirow{4}{*}{(37.7)}    & 320.9 & \multirow{4}{*}{(45.4)}  & 1.4  & \multirow{4}{*}{(9.7)} \\
        & Neuron Masking   & 95.3  & 715.9 &      & 22.2 &     & 4.1 &       & 288.7   &    & 7.1  &  \\
        & Random Weights     & 98.6  & 24.3 &       & 14.6 &     & 2.2 &      & 12.0     &   & 2.7 & \\
        & Random Neurons     & 95.3  & 23.8 &       & 8.3 &     & 8.3 &      & 17.0     &   & 4.5 & \\
        \bottomrule\toprule
        \multirow{4}{*}{ConceptNet} 
        & Weight Masking           & 99.1  & 636.4 & \multirow{4}{*}{(429.0)}      & 2.8 & \multirow{4}{*}{(172.3)}     & 0.2 & \multirow{4}{*}{(37.7)}     & 636.1  & \multirow{4}{*}{(290.6)}      & 1.6 & \multirow{4}{*}{(47.5)}  \\
        & Neuron Masking            & 94.9  & 22422.0 &    & 71.9 &         & 5.4  &      & 8720.2   &   & 29.4  &  \\
        & Random Weights     & 99.1  & 21.0 &        & 14.6 &          & 1.5 &       & 11.4   &     & 5.5  & \\ 
        & Random Neurons     & 94.9  & 110.7 &       & 70.4 &     & 11.2 &      & 35.9     &   & 28.5 & \\
        \bottomrule
    \end{tabular} 
}
\caption{\textbf{Subnetwork discovery for GPT2-small,} averaged over three seeds and seven KGs for WordNet, and three KGs for ConceptNet. \deltappl{} = PPL($\inverseprob$) - PPL($\plmprob$) and similarly for \deltarank{} results. {The values in parenthesis are the average metric (PPL or Rank) for the pretrained model (\ie{}, the base from which the $\Delta$ is computed).} The arrows ($\uparrow$,$\downarrow$) show the desired direction for the metric. Random is an average of randomly masked baselines at the same sparsity levels as the discovered knowledge-critical subnetworks for each KG-seed pair.}
\label{tab:success-criteria}
\end{table*}

\paragraph{Metrics}
We follow prior work \cite{hase-2023-localizationediting} that considers perplexity (PPL) as a proxy for a model's confidence in the expression of knowledge, and calculate the perplexity difference between the remaining and original models, \deltappl{} = PPL($\inverseprob$)  - PPL($\plmprob$). 
We also report \deltarank{}, the tail token rank difference between the remaining and original models. For the \forgetting{} and \maintenancekg{} criteria, we calculate \deltappl{} using the loss on the masked tail entity for triplets in the \targetkg{} and \controlkg{} datasets. For a knowledge-critical subnetwork, we expect \deltappl{} and \deltarank{} values to be high for \targetkg{} and low for \controlkg{}. For the \maintenancelm{} criterion, we calculate \deltappl{} as the average perplexity on all tokens in a sequence, which should be low if removing the critical subnetwork does not affect the model's general language modeling ability.\footnote{We do not report \deltarank{} for \maintenancelm{} as the average rank of all tokens in an open-ended sentence is not as informative as the single tail token rank.} For the \sparsity{} criterion, we calculate the percentage of parameters that were not pruned. 
The denominator is the number of masked parameters, meaning the total size of dense layers in the upper half of the model. Ideally, the sparsity should be as high as possible to keep the majority of parameters (\ie{}, near 99\%).

\paragraph{Baseline} We use weight and neuron masking to localize knowledge-critical subnetworks. As a control baseline, we create randomly masked models at the same sparsity level as the knowledge-critical subnetwork. If the discovered subnetwork is critical for expressing \targetkg{}, then removing a random subnetwork at the same weight or neuron sparsity should yield lower corruption for expressing \targetkg{} (\ie, lower \deltappl{}) than removing the critical subnetwork. Similarly, if the critical subnetwork successfully preserves the \textbf{maintenance} criteria, a random subnetwork should be more likely to prune useful weights for expressing \controlkg{} and \lmodeling{}, which should lead to a higher \deltappl{} on maintenance datasets. {Further information on the implementation of the random masking baseline is in Appendix~\ref{sec:train-implement}.} 
\section{Experimental Results}

We first evaluate the degree to which discovered subnetworks are knowledge-critical.

\begin{table*}[t]
\centering
\resizebox{0.95\linewidth}{!}{
    \begin{tabular}{l r@{\hspace{2pt}}l r@{\hspace{2pt}}l r@{\hspace{2pt}}l r@{\hspace{2pt}}l}
    \toprule
    \multirow{2}{*}{\textbf{Ablation}} &
    \multicolumn{2}{c}{\textbf{Sparsity}} &
    \multicolumn{2}{c}{\textbf{\targetkg{}}} &
    \multicolumn{2}{c}{\textbf{\controlkg{}}} &
    \multicolumn{2}{c}{\textbf{\lmodeling{}}} \\
    &
    \multicolumn{2}{c}{($\uparrow$)}      &
    \multicolumn{2}{c}{\textbf{\deltappl{} ($\uparrow$)}}  &
    \multicolumn{2}{c}{\textbf{\deltappl{} ($\downarrow$)}} &
    \multicolumn{2}{c}{\textbf{\deltappl{} ($\downarrow$)}}
    \\ \midrule
            No Suppression       & 99.5 & [99.5, 99.5] & -7.2 & [-11.9, -3.7] & -3.2 & [-3.2, -3.2] & 0.2 & [0.2, 0.2] \\
            No Maintenance-LM & 99.2 & [99.0, 99.3] & 259.8 & [-1.5, 401.7] & 9.0 & [-3.6, 25.1] & 25.9 & [24.7, 27.3] \\
            No Maintenance-KG  & 99.8 & [99.8, 99.8] & 21141.1 & [16885.9, 25471.8] & 1697.5 & [1334.6, 2180.1] & 0.2 & [0.2, 0.2] \\ \midrule 
            Our Method    & 98.6 & [97.8, 99.1] & 378.1 & [74.3, 834.9] & 1.6 & [-0.7, 4.0] & 0.5 & [0.3, 0.8] \\
            \bottomrule
    \end{tabular}
}
\caption{\textbf{Ablation study for the multi-objective loss on GPT2-small {using weight masking,}} with [min, max] boundaries, averaged across three KGs and two seeds.}
\label{tab:ablation}
\end{table*}

\paragraph{Weight-masked Subnetworks} In Table~\ref{tab:success-criteria}, we observe that across seven different knowledge graphs (\targetkg{}s) and three random seeds, the subnetworks {found with weight masking} consistently achieve a notably high sparsity ($>$ 98\%).\footnote{Table~\ref{tab:success-criteria-boundaries} provides individual KG results for the averaged weight masking results in Table~\ref{tab:success-criteria}.} 
For the \forgetting{} criterion, we notice a high \deltappl{} on \targetkg{} for both approaches, meaning that the perplexity of the remaining model on \targetkg{} is significantly higher than the pretrained model's perplexity.
In contrast, removing a random subnetwork at the same sparsity yields a smaller perplexity increase, meaning the discovered subnetworks are significantly more critical for expressing \targetkg{}.
At the same time, we find little change in perplexity on the maintenance datasets for relational knowledge (\controlkg{}) and language modeling (\lmodeling{}), demonstrated by the negligible \deltappl{} on both datasets and the small \deltarank{} value on \controlkg{}. \footnote{Note that the lower average PPL of \controlkg{} compared to \targetkg{} is due to \controlkg{} being larger, which minimizes the impact of outliers and reduces average perplexity.} 
We note that a negative \deltappl{} here may result from the remaining model slightly overfitting to the \controlkg{} distribution, although it is never too significant. 

We observe similar results for knowledge-critical subnetworks for larger models. For three \targetkg{}s: \texttt{communication}, \texttt{representation}, and \texttt{location}, we observe an average increase in \targetkg{} perplexity of 256 for GPT2-medium, 5780 for GPT2-large, 536 for GPT2-XL, and a negligible maintenance \deltappl{} (Table~\ref{tab:additional-success-criteria-scaled}).

\paragraph{Neuron-masked Subnetworks}
On the other hand, neuron masking does not reliably fulfill the conditions of discovering knowledge-critical subnetworks. While removing neuron-masked subnetworks yields greater suppression of \targetkg{} than weight masking, it also significantly impacts \controlkg{} \deltappl{} and \deltarank{} (more than randomly removing neurons at the same sparsity), indicating that other behaviors of the model are not robustly maintained. They also tend to be less sparse, frequently keeping $\sim$5\% of the parameters of the original model.\footnote{Appendix Table \ref{tab:success-criteria-boundaries-neuron} provides individual KG results for the averaged neuron masking results in Table~\ref{tab:success-criteria}.} We hypothesize that this observation is potentially related to neuron superposition \citep{elhage2022superposition}, where the neurons that represent \targetkg{} cannot be fully disentangled from representations that encode general relational knowledge. While weights may also be polysemantic, they are more fine-grained, potentially encoding knowledge in a more separable manner. 

\vspace{1pt}
\paragraph{Ablation Study}
As our method relies on a joint objective combining multiple loss functions, we perform an ablation study of the loss terms presented in \S\ref{sec:loss-terms} for weight masking and remove each objective (i.e., No Suppression, No Maintenance-KG, No Maintenance-LM) to validate whether these losses accomplish their goals.\footnote{We do not ablate the \sparsity{} term. Without it, the subnetwork search stagnates at the initial sparsity.} 
In Table~\ref{tab:ablation}, we observe that the \forgetting{} loss is necessary to increase \targetkg{} perplexity (and suppress the knowledge). Without it, the model only optimizes for retaining \controlkg{}, and generalizes this improvement to \targetkg{} as well (as indicated by the negative \deltappl{}).
We also find that removing the maintenance losses significantly affects \controlkg{} and \lmodeling{} perplexity differences. Without these controls, our method learns to suppress the knowledge from the model by suppressing \textit{general abilities}. The \forgetting{} objective, a minimization of the KL divergence between the output distribution and a uniform distribution, affects the prediction of tail entities for all relational knowledge rather than affecting only \targetkg{}. We present additional ablations related to varying the training objectives in Appendices~\ref{sec:train-implement} (varying $\lambda_i$ in Eq.~\ref{eq:objective}) and \ref{sec:expression-loss} (adding additional loss terms).

\paragraph{Paraphrase Generalization}
To assess whether our subnetworks generalize to other verbalizations of \targetkg{} and \controlkg{}, we evaluate the pruned models on 20 other distinct relation paraphrases that are not used during training. {Specifically, we vary the tokens representing the relation and the format of the head and tail entities while still ensuring grammatical correctness.\footnote{For further details and examples on the creation of these verbalizations, please refer to Appendix~\ref{sec:data-creation}.}} 
For weight masking, our conclusions do not change when using other prompt styles, as seen in Table~\ref{tab:qual-paraphrase}. Interestingly, the \deltappl{} for \controlkg{} paraphrases is sometimes lower than for the format used for training, likely because the starting perplexity is higher on other templates,\footnote{Recall that for every triplet, we use the verbalization with the lowest original model perplexity for training. The average perplexity of GPT2-small on the \textit{worse} paraphrases is 3231 for \targetkg{} and 2368 for \controlkg{}.} and the maintenance of \controlkg{} generalizes to a greater degree on these suboptimal templates. The neuron masking approach generalizes well to \targetkg{} paraphrases, but poorly for \controlkg{} templates, reinforcing our previous observations.

\begin{table}
\centering
\resizebox{1.0\linewidth}{!}{
    \begin{tabular}{ccrr}
        \toprule
        \textbf{Mask} & \textbf{Knowledge}    & \textbf{\targetkg{}}                  & \textbf{\controlkg{}}              \\
        \textbf{Method} & \textbf{Graph}        & \textbf{\deltappl{} ($\uparrow$)}     & \textbf{\deltappl{} ($\downarrow$)} \\ \midrule
        \multirow{3}{*}{Weight} &\texttt{communication} & 492.4 & 3.0 \\ 
        &\texttt{location} & 684.9 & -33.0 \\ 
        &\texttt{representation} & 916.4 & -6.9  \\ \midrule
        \multirow{3}{*}{Neuron} &\texttt{communication} & 866.8 & -325.0 \\ 
        &\texttt{location} & 941.6 & 284.5 \\ 
        &\texttt{representation} & 1928.9 & 654.5  \\ \bottomrule
    \end{tabular}
}
\caption{\textbf{Paraphrase results on GPT2-small.}}
\label{tab:qual-paraphrase}
\vspace{-10pt} 
\end{table}
 
\subsection{Subnetwork Analysis}

\paragraph{Subnetwork Structure} 
To better understand how \textit{knowledge-critical} subnetworks interact with the rest of the model, we explore their structure in the parameter space of the original model. For three WordNet \targetkg{}s and three random seeds, we find that GPT2-small subnetworks are relatively denser in the first and final masked transformer blocks. {For weight masking, more density is} observed in the attention sublayers (Figure~\ref{fig:structure-layer}). Interestingly, much of the density of the subnetworks in the attention sublayers is tied to individual attention heads (Figure~\ref{fig:structure-head}), supporting prior conclusions that particular attention heads encode semantic relationships \citep{clark-etal-2019-bert, geva-etal-2023-dissecting}.

However, despite being dense around similar portions of the model across different \targetkg{}s and random seeds, the subnetworks are quite distinct. When we calculate the Jaccard similarity (\ie{}, IoU) of the individual parameters across subnetworks for different random seeds for the same \targetkg{}, the result is quite low on average {for weight-masked subnetworks} (3-4\%) --- though higher for the final attention output sublayer (10-12\%) --- indicating the \textit{knowledge-critical} subnetworks are quite disjoint, even when discovered by suppressing the same information (Figure ~\ref{fig:layer-jaccard}).

Neuron masking led to a much higher density in the second feedforward layers of the transformer blocks and attention layers (Figure~\ref{fig:structure-layer-input-neuron}). We find that the IoU of neuron-masked subnetworks are also 10$\times$ higher (34-44\%; Figure~\ref{fig:layer-jaccard-neuron-inputs}), partially due to their reduced sparsity, but also perhaps indicating that neuron masking yields more unique subnetworks across seeds, though they are also less reliably knowledge-critical.

\begin{table}
\centering
\resizebox{1.0\linewidth}{!}{
    \begin{tabular}{cccccc}
        \toprule
        \multicolumn{2}{c}{\textbf{Mask Composition}}   & \textbf{Sparsity} & \textbf{\targetkg{}} & \textbf{\controlkg{}} & \textbf{\lmodeling{}}  \\
         \multicolumn{2}{c}{\textbf{Pattern}} & \textbf{($\uparrow$)}   & \textbf{\deltappl{} ($\uparrow$)}  & \textbf{\deltappl{} ($\downarrow$)} & \textbf{\deltappl{} ($\downarrow$)} \\ \midrule
        \multirow{4}{*}{Weight} & Individual & 98.8 & 379.1 & 0.7 &	0.4 \\ \cmidrule{2-6}
        & Union & 96.9 & 1984.4  & 44.9 & 4.7 \\ 
        & Floral & 99.5 & 4.9  & 4.5  & 1.1  \\
        & Intersection & 99.9 & 7.9  & 3.7 & 2.1 \\ 
        \midrule
        \multirow{4}{*}{Neuron} & Individual & 95.2  & 396.0 & 22.3 & 4.1  \\ \cmidrule{2-6}
        & Union & 92.8 & 2743.0 & 86.9 & 6.6 \\ 
        & Floral & 95.5 & 290.5 & 16.1 & 3.7 \\ 
        & Intersection & 97.3 & 63.9 & 3.5 & 2.4 \\
        \bottomrule
    \end{tabular}
}
\caption{\textbf{Composing subnetworks with GPT2-small.} Individual stands for the individual subnetwork removal average across the same three seeds and KGs.}
\label{tab:seed-combo}
\vspace{-10pt}
\end{table}

\vspace{-5pt}
\paragraph{Subnetwork Composition}
However, even though knowledge-critical subnetworks across random seeds may be disentangled, composing them (and removing them jointly) amplifies the suppression effect. As shown in Table~\ref{tab:seed-combo}, when we compose subnetworks for GPT2-small as a union of three random seed masks for the same \targetkg{}, the suppression effect increases significantly, by a factor of 6$\times$ (far more than removing additional random parameters from the remaining model; Figure~\ref{fig:additional-param-stab}). While this suppression is accompanied by a degradation in the maintenance criteria ($\sim$30-40 \deltappl{} on \controlkg{} instead of near 0), the absolute difference is far smaller. 
Composing neuron-masked subnetworks yields similar trends, though we observe two interesting patterns. First, the intersection of these subnetworks produces a subnetwork that satisfies the maintenance criteria to be \textit{knowledge-critical}, though at the cost of reducing suppression. Second, neuron-masked compositions yield monotonic changes in suppression and maintenance scores as sparser composition methods are used. 
Further analyses on seed-based and knowledge-based variance across discovered subnetworks are in Appendix \ref{sec:seed-analysis} and \ref{sec:kg-analysis}, respectively.

\paragraph{Subnetwork Sensitivity} 
Finally, we investigate whether discovered subnetworks are structurally sensitive. Specifically, we perform a sensitivity analysis of the recorded metrics as we iteratively expand or contract the subnetwork (by adding or removing parameters). As we add parameters to the subnetwork (\ie{}, remove parameters from the remaining model), we measure the change in \targetkg{} \deltappl{}. In this case, a sudden drop in \deltappl{} would indicate that the discovered subnetwork is spurious. In Appendix Figure~\ref{fig:additional-param-stab}, we observe that expanding the discovered subnetwork in small amounts does not significantly recover the model's ability to express \targetkg{}, providing further evidence that the subnetworks are not arbitrarily discovered, but rather have meaningful knowledge-expressing structure within the larger model. We provide more experimental details in Appendix~\ref{sec:param-based}. 
\subsection{Downstream Task Transfer}
\label{sec:downstream}

If a subnetwork is truly knowledge-critical, its removal should harm a pretrained language model's ability to transfer to a downstream task requiring the knowledge encoded by the subnetwork. To test this hypothesis, we finetune a model on the CommonsenseQA benchmark \cite{talmor-etal-2019-commonsenseqa} after removing a relevant knowledge-critical subnetwork. We use the in-house splits from \citet{lin-etal-2019-kagnet}, with a development set of 1241 and an initial test set of 1221 questions. In the test set, we induce\footnote{We describe this process in Appendix~\ref{sec:additional-downstream}.} the ConceptNet relation linked to each question and extract the relevant triplets from ConceptNet, creating a \targetkg{} from all ConceptNet triplets associated to the test set, which yields a \textbf{filtered} set of 363 questions for which we can reliably extract relevant ConceptNet triplets. 
We use these relevant triplets as \targetkg{} and the remaining distinct triplets in the LAMA subset of ConceptNet as \controlkg{} to learn a knowledge-critical subnetwork using either weight and neuron masking for GPT2-small. Then, we apply different finetuning methods to the remaining model after removing the critical subnetwork, using the same training set. We compare finetuning the remaining masked model (Weight Mask, Neuron Mask in Table~\ref{tab:downstream-acc}) to the performance of finetuning the full pretrained model (Full), as well as a randomly masked model at the same sparsity as the masked-weight subnetwork (Random Mask). We report results across three random seeds in Table~\ref{tab:downstream-acc}. 

For all finetuning methods, we find that the remaining model with weight masking has similar accuracy to the pretrained model on the development split and a close accuracy for the overall test set. However, we observe a consistent significant performance drop on the filtered subset after finetuning (average drop of 7.3\%; head tuning barely better than selecting a random answer on a 5-choice MCQA task), indicating that the model struggles to transfer knowledge associated with \targetkg{} during fine-tuning. 
Interestingly, in less parameter-efficient finetuning methods, this drop does not persist when the neuron-masked subnetwork is removed, suggesting that knowledge is either still transferred or recovered over the course of finetuning \citep{lo-etal-2024-large}. 
In addition, for both head tuning and LoRA \cite{hu2022lora} with weight masking, we find that if we randomly split the filtered \targetkg{}, one half's knowledge-critical mask does not affect the accuracy of the other half as significantly as its own (see Appendix~\ref{sec:additional-downstream} for details), indicating the performance drop is indeed specific to the pruned knowledge.

\begin{table}
\centering
\resizebox{\linewidth}{!}{
    \begin{tabular}{clccc}
        \toprule
        \textbf{Tuning Method}                & \textbf{Subnetwork}     & \textbf{Dev}    & \textbf{Test} & \textbf{Filtered}  \\ \midrule
        \multirow{3}{*}{\parbox{2cm}{\centering \textbf{Head Tuning}}}                
                        & Full                    & 38.63           & 38.33         & 37.19     \\ \cmidrule{2-5}
                        & Random Mask                 & \textcolor{red}{-0.47} & \textcolor{red}{-1.61} & \textcolor{red}{-3.21}    \\ 
                        & Neuron Mask              & \textcolor{red}{-3.74} & \textcolor{red}{-4.22} & \textcolor{red}{-6.61}     \\
                        & Weight Mask               & \textcolor{red}{-1.69} & \textcolor{red}{-6.80} & \textcolor{red}{-14.42}     \\  \midrule
        \multirow{3}{*}{\parbox{2cm}{\centering \textbf{LoRA}}}
                        & Full                    & 50.04	    & 48.64	    & 48.67\\ \cmidrule{2-5}
                        & Random Mask                 & \textcolor{red}{-0.74} & \textcolor{red}{-2.33} & \textcolor{red}{-1.75}     \\ 
                        & Neuron Mask              & \textcolor{red}{-0.30} & \textcolor{blue}{+1.89} & \textcolor{blue}{+1.28}     \\
                        & Weight Mask              & \textcolor{red}{-1.83} & \textcolor{red}{-2.74} & \textcolor{red}{-3.95}     \\ \midrule
        \multirow{3}{*}{\parbox{2cm}{\centering \textbf{Full Finetuning}}}
                        & Full                    & 44.61	    & 42.33	    & 42.79     \\ \cmidrule{2-5}
                        & Random Mask                 & \textcolor{blue}{+0.30} & \textcolor{red}{-0.24} & \textcolor{blue}{+2.39}     \\ 
                        & Neuron Mask               & \textcolor{blue}{+0.38} & \textcolor{blue}{+2.17} & \textcolor{blue}{+1.47}    \\
                        & Weight Mask               & \textcolor{red}{-1.50} & \textcolor{red}{-5.14} & \textcolor{red}{-3.60}     \\ \bottomrule
    \end{tabular}}
\vspace{5pt}
\caption{\textbf{Accuracy on CommonsenseQA,} averaged over three seeds for GPT2-small.}
\label{tab:downstream-acc}
\vspace{-10pt}
\end{table}

\section{Conclusion}

In this paper, we conceptualize knowledge-critical subnetworks, sparse computational subgraphs within larger language models that are responsible for expressing specific knowledge relationships. We discover these subnetworks using a multi-objective differentiable masking approach that jointly optimizes a criterion designed to suppress the expression of target knowledge when knowledge-critical subnetworks are removed from a language model, and maintenance criteria that ensure the language model retains its initial capacity to model other relational knowledge and general language. Our results show that when knowledge-critical subnetworks are removed, a model loses its ability to express the knowledge encoded in the subnetwork, and to transfer it when finetuned on downstream tasks requiring the knowledge.
\section*{Acknowledgements}
We thank Mohammadreza Banaei, Syrielle Montariol, Debjit Paul, Khai Loong Aw, Badr AlKhamissi, Silin Gao, Yifan Hou, Beatriz Borges, Yu Fei, and Angelika Romanou for their helpful discussions and feedback on our manuscript.
We also gratefully acknowledge the support of the Swiss National Science Foundation (No. 215390), Innosuisse (PFFS-21-29), the EPFL Center for Imaging, Sony Group Corporation, and the Allen Institute for AI.

\section*{Limitations}
We discuss the limitations of our proposed method and conducted experiments on three axes: data, model, and hyperparameter. We emphasize that the data used for our experiments are limited to English only. As English is a high-resource language, additional challenges could arise when reproducing our method in a low-resource language (\eg{}, finding a rich lexical database like WordNet). We identify the lack of diverse pretrained language model architectures and language modeling objectives as the main model limitation. We have tested our method on the billion scale but did not expand our scope to larger models with different architectures (for example, in the 7B scale). We also limit the analysis to models trained with an autoregressive language modeling objective in contrast to text-to-text models such as T5 \cite{raffel-2020-t5} or Masked-Language-Modeling models such as RoBERTa \cite{liu-2019-roberta}. Finally, the hyperparameter search detailed in the Appendix, while not exhaustive, provides sufficient evidence to support the validity of the selected range. To find more precise knowledge-critical subnetworks, future methods may need to take this hyperparameter search further.

\section*{Ethics Statement}
In this study, we concentrate on relational knowledge, but the technique of identifying subnetworks could be used in mitigating bias within models. Likewise, this method of finding subnetworks may also inadvertently lead to the elimination of critical ethical or factual knowledge from a language model, resulting in a model that could generate offensive content and misinformation. For example, there exists a backdoor attack method against deep neural networks that builds on top of the identification and editing of subnetworks \cite{qi2021subnet}. Therefore, caution should be exercised when applying the identification and removal of subnetworks to models used in essential applications.

\bibliography{custom}

\appendix

\begin{table*}[t]
\centering
\resizebox{\linewidth}{!}{
\begin{tabular}{clr@{\hspace{2pt}}c@{\hspace{2pt}}ll}
\toprule
\multicolumn{2}{c}{\multirow{2}{*}{\textbf{Knowledge Graph}}}            & \multicolumn{3}{c}{\textbf{Triplets}}                                         &   \multicolumn{1}{c}{\multirow{2}{*}{\textbf{Verbalization}}}          \\
                    &                                                    &              \textbf{Head}  & \textbf{Relation} & \textbf{Tail}              &                                            \\ \midrule

\multirow{15}{*}{WordNet}   & \multirow{3}{*}{\controlkg{} train}       &  \texttt{(casserole.n.02,} & \texttt{IsA,} & \texttt{dish.n.01)}              &   \texttt{``A casserole is a dish''}                 \\
                            &                                           &  \texttt{(passerby.n.01,} & \texttt{IsA,} & \texttt{pedestrian.n.01)}         &   \texttt{``A passerby is a type of pedestrian''}    \\
                            &                                           &  \texttt{(chorizo.n.01,} & \texttt{IsA,} & \texttt{sausage.n.01)}             &   \texttt{``A chorizo is a kind of sausage''}        \\ \cmidrule{2-6}
                            & \multirow{3}{*}{\controlkg{} val.}        &  \texttt{(crate.n.01,} & \texttt{IsA,} & \texttt{box.n.01)}                   &   \texttt{``A crate is a kind of box''}              \\
                            &                                           &  \texttt{(magnetometer.n.01,} & \texttt{IsA,} & \texttt{meter.n.02)}          &   \texttt{``Magnetometer is a type of meter''}       \\  &                                           &  \texttt{(vaccinee.n.01,} & \texttt{IsA,} & \texttt{patient.n.01)}            &   \texttt{``A vaccinee is a patient''}               \\ \cmidrule{2-6}
                            & \multirow{3}{*}{\texttt{communication}}   &  \texttt{(message.n.01,}  & \texttt{IsA,} & \texttt{communication.n.02)}     &   \texttt{``A message is a type of communication''}  \\
                            &                                           &  \texttt{(indicator.n.02,}  & \texttt{IsA,} & \texttt{signal.n.01)}          &   \texttt{``An indicator is a type of signal''}      \\
                            &                                           &  \texttt{(evidence.n.02,} & \texttt{IsA,} & \texttt{indication.n.01)}       &   \texttt{``Evidence is an indication''}             \\ \cmidrule{2-6}
                            & \multirow{3}{*}{\texttt{location}}        &  \texttt{(region.n.01,}  & \texttt{IsA,} & \texttt{location.n.01)}           &   \texttt{``A region is a location''}                \\
                            &                                           &  \texttt{(district.n.01,}  & \texttt{IsA,} & \texttt{region.n.03)}           &   \texttt{``A district is a region''}                \\
                            &                                           &  \texttt{(expanse.n.03,}  & \texttt{IsA,} & \texttt{space.n.02}              &   \texttt{``An expanse is a type of space''}         \\ \cmidrule{2-6}
                            & \multirow{3}{*}{\texttt{representation}}  &  \texttt{(representation.n.02,} & \texttt{IsA,} & \texttt{creation.n.02)}   &   \texttt{``Representation is a kind of creation''}  \\
                            &                                           &  \texttt{(delineation.n.02,} & \texttt{IsA,} & \texttt{drawing.n.02)}       &   \texttt{``A delineation is a type of drawing''}    \\
                            &                                           &  \texttt{(chart.n.02,} & \texttt{IsA,} & \texttt{map.n.01)}                 &   \texttt{``A chart is a map''}                      \\ \midrule
                            
\multirow{9}{*}{ConceptNet} & \multirow{3}{*}{\controlkg{} train}       &  \texttt{(briefcase,} & \texttt{AtLocation,} & \texttt{desk)}               &   \texttt{``A briefcase is typically placed at a desk''}   \\
                            &                                           &  \texttt{(vegetarian,} & \texttt{NotDesires,} &  \texttt{meat)}              &   \texttt{``A vegetarian doesn't crave meat''}    \\
                            &                                           &  \texttt{(voting,} & \texttt{Causes,} & \texttt{election)}                  &   \texttt{``A voting can lead to an election''}   \\ \cmidrule{2-6}
                            & \multirow{3}{*}{\controlkg{} val.}        &  \texttt{(boat,} & \texttt{UsedFor,} & \texttt{sailing)}                    &   \texttt{``A boat is designed for sailing''}     \\
                            &                                           &  \texttt{(clothes,} & \texttt{ReceivesAction,} & \texttt{washed)}           &   \texttt{``Clothes can be washed''}              \\
                            &                                           &  \texttt{(jogging,} & \texttt{HasPrerequisite,} & \texttt{energy)}          &   \texttt{``A jogging requires an energy''}       \\ \cmidrule{2-6}
                            & \multirow{3}{*}{\texttt{fruit}}        &  \texttt{(fruit,} & \texttt{ReceivesAction,} & \texttt{eaten)}              &   \texttt{``A fruit can be eaten''}               \\
                            &                                           &  \texttt{(wine,} & \texttt{MadeOf,} & \texttt{fruit)}                       &   \texttt{``A wine comprises of a fruit''}        \\
                            &                                           &  \texttt{(champagne,} & \texttt{IsA,} & \texttt{wine)}                      &   \texttt{``Champagne is a type of wine''}        \\ \bottomrule
\end{tabular}
}
\caption{\textbf{Examples of KG triplets,} and the best GPT-2 small verbalization for WordNet and ConceptNet.}
\label{tab:kg-examples}
\end{table*}

\section{Dataset Creation and Processing}
\label{sec:data-creation}

\paragraph{\targetkg{}} 
To gather small connected \targetkg{}s, we randomly select an initial node and sample knowledge triplets by walking a depth of three up (parent direction) and down (child direction) in the respective KG. Given a seed node such as \texttt{representation} in WordNet\footnote{In WordNet, a word sense is represented by its lemma, syntactic category, and sense ID (\eg{}, in \texttt{map.n.01}, \texttt{n} for noun and \texttt{01} for sense ID). We omit this naming convention from the main paper tables for readability.} or \texttt{fruit} in ConceptNet, we sample relations by performing a 3-hop random walk. For example, for the \texttt{fruit} KG shown in Table~\ref{tab:kg-examples}, we start from the seed concept \texttt{fruit}. In the first depth, we retrieve \texttt{(fruit, ReceivesAction, eaten)} and \texttt{(wine, MadeOf, fruit)}. In the next depth, we retrieve \texttt{(champagne, IsA, wine)}, and so forth for all possible relations. Note that we only sample relations with a single-token tail entity. 

Once this connected KG is sampled, we apply two filtering processes. The first one enforces many-to-one relationships in $\targetgraph$ to avoid head entities with multiple tails. The second filtering process reduces the tail-entity imbalance to avoid over-fitting to a small set of tokens. For this, we count the frequency of the tail tokens in the sampled graph and keep at most a quartile amount of triplets with shared tail entities. 

Finally, we verbalize \targetkg{} graph with the formats that give the lowest perplexity on the pretrained model. We try various relation-specific verbalization templates per knowledge triplet and pick the one that yields the lowest tail-token perplexity. For example, in the \texttt{representation} graph, while the model had lower perplexity with the template \texttt{``\{$h$\} is a kind of \{$t$\}''} for the triplet \texttt{(representation.n.02, IsA, creation.n.02)}, it also had lower perplexity with the template \texttt{``A \{$h$\} is a \{$t$\}''} for the triplet \texttt{(chart.n.02, IsA, map.n.01)}. Note that this can change for each model size, such as GPT2-small, medium, large and XL.

\paragraph{\controlkg{}} To create \controlkg{}, we prioritize not leaking \targetkg{} counterfactuals and having a shared \controlkg{} across different \targetkg{}s. Therefore, we remove from the complete KG (\eg{}, for ConceptNet \targetkg{}s, the complete LAMA subset of ConceptNet) any triplet that shares the same entities as the union of the \targetkg{}s shown in Table~\ref{tab:kg-stats}. For all KG verbalizations, to remove and maintain knowledge that the model is already confident about, we pick the best scoring verbalization for each triplet among several prompt styles and filter out those that yield an individual PPL higher than a threshold. For testing, we use held-out triplets.

\paragraph{\lmodeling{}} We use WikiText-2 \cite{merity-2017-pointer} for the \lmodeling{} dataset. We tokenize each entry and then concatenate all of them together. Finally, we group the tokens into chunks of 512. For validation and testing, we use held-out sets.

\section{Training and Evaluation Implementation}
\label{sec:train-implement}
\begin{table*}[t]
\centering
\resizebox{0.80\linewidth}{!}{
    \begin{tabular}{c r@{\hspace{2pt}}l r@{\hspace{2pt}}l r@{\hspace{2pt}}l r@{\hspace{2pt}}l}
        \toprule
        \textbf{Init. Mask}  & \multicolumn{2}{c}{\textbf{Sparsity}} & \multicolumn{2}{c}{\textbf{\targetkg{}}} & \multicolumn{2}{c}{\textbf{\controlkg{}}} & \multicolumn{2}{c}{\textbf{\lmodeling{}}}  \\
        \textbf{Probability}   & \multicolumn{2}{c}{($\uparrow$)}      & \multicolumn{2}{c}{\textbf{\deltappl{}($\uparrow$)}}  & \multicolumn{2}{c}{\textbf{\deltappl{} ($\downarrow$)}} & \multicolumn{2}{c}{\textbf{\deltappl{} ($\downarrow$)}} \\ \midrule
        0.25  & 99.5  & [99.5, 99.6] & 332.1    & [119.5, 544.6]   & 3.7  & [2.6, 4.7]   & 0.1 & [0.1, 0.1] \\ 
        0.45  & 99.5  & [99.5, 99.5] & 1287.8   & [80.0, 2495.6]   & 0.9  & [-1.1, 2.9]  & 0.2 & [0.1, 0.2] \\ 
        0.50  & 99.4  & [99.4, 99.5] & 939.1    & [59.9, 1818.2]   & -0.2 & [-0.2, -0.1] & 0.2 & [0.2, 0.2] \\ 
        0.75  & 98.9  & [98.8, 99.1] & 13043.1  & [4.6, 26081.5]   & -1.9 & [-1.9, -1.8] & 0.3 & [0.3, 0.4] \\  
        \bottomrule
    \end{tabular}
}
\caption{\textbf{Hyperparameter study on inital mask probability with GPT2-small.} }
\label{tab:mask-init}
\end{table*}

\begin{table*}[t]
\centering
\resizebox{1.0\linewidth}{!}{
    \begin{tabular}{c c c c c  r@{\hspace{2pt}}l r@{\hspace{2pt}}l r@{\hspace{2pt}}l r@{\hspace{2pt}}l r@{\hspace{2pt}}l l}
        \toprule
        \textbf{Masking} & \multirow{2}{*}{$\bm{\lambda_1}$} &  \multirow{2}{*}{$\bm{\lambda_2}$} & \multirow{2}{*}{$\bm{\lambda_3}$} & \multirow{2}{*}{$\bm{\lambda_4}$}  & \multicolumn{2}{c}{\textbf{Sparsity}} & \multicolumn{2}{c}{\textbf{\targetkg{}}} & \multicolumn{2}{c}{\textbf{\controlkg{}}} & \multicolumn{2}{c}{\textbf{\lmodeling{}}}  & \multicolumn{2}{c}{\textbf{Number of Valid}} \\
          \textbf{Method} & & & &  & \multicolumn{2}{c}{($\uparrow$)}      & \multicolumn{2}{c}{\textbf{\deltappl{}($\uparrow$)}}  & \multicolumn{2}{c}{\textbf{\deltappl{} ($\downarrow$)}} & \multicolumn{2}{c}{\textbf{\deltappl{} ($\downarrow$)}} & \multicolumn{2}{c}{\textbf{Checkpoints}}\\ \midrule
        \multirow{4}{*}{Weight} 
        & 1	& 1	& 3	& 1	& 96.9 & [95.6, 98.2]	& 17.3 & [-7.8, 42.4]	& 1.0 & [-2.0, 4.1]	& 0.5 & [0.2, 0.8]	& 2.5 & [0, 5] \\
        & 1	& 3	& 1	& 1	& 97.3 & [96.4, 98.3]	& 80.9 & [59.2, 102.6]	& -1.5 & [-8.9, 5.8]& 1.0 & [0.5, 1.5]	& 53.5 & [5, 102] \\
        & 3	& 1	& 1	& 1	& 98.5 & [97.6, 99.3]	& 12516.4 & [682.2, 24350.6]& 	0.0 & [-1.4, 1.5]& 	0.4 & [0.2, 0.6]	& 145.0 & [133, 157] \\ \midrule
        \multirow{4}{*}{Neuron}
        & 1 & 1 & 3 & 1 & 96.7 & [96.6, 96.9] & 297.5 & [104.0, 491.0] & 3.9 & [-1.7, 9.6] & 2.9 & [2.9, 3.0] & 0.0 & [0, 0] \\ 
        & 1 & 3 & 1 & 1 & 95.3 & [95.3, 95.4] & 110.2 & [53.3, 167.2] & 11.8 & [10.0, 13.7] & 3.5 & [3.4, 3.6] & 0.0 & [0, 0] \\ 
        & 3 & 1 & 1 & 1 & 93.3 & [93.1, 93.5] & 255.1 & [135.0, 375.1] & 44.7 & [38.8, 50.7] & 5.2 & [4.7, 5.7] & 0.0 & [0, 0] \\
        \bottomrule
    \end{tabular}
}
\caption{\textbf{Hyperparameter study on $\lambda_i$ loss weights in Eq.~\ref{eq:objective} with GPT2-small.}}
\label{tab:loss_tradeoff}
\end{table*}

\begin{table}[t]
\centering
\resizebox{0.9\linewidth}{!}{
    \begin{tabular}{ccccc}
        \toprule
        \textbf{Setup} & \textbf{Model}  & \textbf{\textsc{KG}} & \textbf{\lmodeling{}}  \\ \midrule
              & GPT-2 Small & 250 & 10 \\ 
        Train & GPT-2 Medium & 96 & 4 \\  
              & GPT-2 Large & 96 & 4 \\ 
              & GPT-2 XL & 128 & 4 \\ \midrule
              & GPT-2 Small & 250 & 8 \\ 
        Eval & GPT-2 Medium & 250 & 8 \\  
              & GPT-2 Large & 250 & 8 \\
              & GPT-2 XL & 250 & 8 \\ \bottomrule
    \end{tabular}
}
\caption{\textbf{GPU batch size for each dataset and model.}}
\label{tab:batch-size}
\end{table}
\begin{table}[t]
\centering
\resizebox{1.0\linewidth}{!}{
    \begin{tabular}{ccccc}
        \toprule
         \multirow{2}{*}{\textbf{Iteration}}  & \textbf{\targetkg{}} & \textbf{\controlkg{}} & \textbf{\lmodeling{}}  \\
                           & \textbf{\deltappl{} Floor}  & \textbf{\deltappl{} Ceiling} & \textbf{\deltappl{} Ceiling} \\ \midrule
        1 & 35.0 & 5.0  & 1.0 \\ 
        2 & 40.0 & 7.0  & 2.0 \\  
        3 & 40.0 & 10.0 & 3.0 \\ 
        4 & 50.0 & 15.0 & 4.0 \\ \bottomrule
    \end{tabular}
}
\caption{\textbf{Selection limit for each success criteria.}}
\label{tab:ckpnt-select}
\end{table}
\begin{table*}[t]
\centering
\resizebox{1.0\linewidth}{!}{
    \begin{tabular}{cc r@{\hspace{2pt}}l r@{\hspace{2pt}}l r@{\hspace{2pt}}l r@{\hspace{2pt}}l c}
        \toprule
        \thead{Masked Layer} & \thead{Percentage} & \multicolumn{2}{c}{\textbf{Sparsity}} & \multicolumn{2}{c}{\textbf{\targetkg{}}} & \multicolumn{2}{c}{\textbf{\controlkg{}}} & \multicolumn{2}{c}{\textbf{\lmodeling{}}}  & \thead{\# of} \\
        \thead{Choice} & \thead{Masked} & \multicolumn{2}{c}{($\uparrow$)}      & \multicolumn{2}{c}{\textbf{\deltappl{}($\uparrow$)}}  & \multicolumn{2}{c}{\textbf{\deltappl{}} ($\downarrow$)} & \multicolumn{2}{c}{\textbf{\deltappl{} ($\downarrow$)}} & \thead{checkpoints} \\ \midrule
        0-11 & 100\% & 95.6 & [94.9, 96.6] & 242.7 & [-26.6, 1254.3] & 11.8 & [6.7, 15.9] & 1.3 & [1.0, 1.7] & 1.1 \\ 
        3-11 & 75\% & 97.3 & [94.4, 98.3] & 669.7 & [-8.7, 2119.7] & -1.4 & [-8.3, 10.8] & 1.0 & [0.5, 2.9] & 76.9 \\ 
        6-11 & 50\% & 98.6 & [97.1, 99.2] & 870.4 & [38.7, 2665.1] & 0.4 & [-2.6, 4.0] & 0.5 & [0.3, 1.0] & 104.1 \\ 
        9-11 & 25\% & 99.2 & [98.0, 99.7] & 1185.0 & [62.3, 4787.0] & 4.2 & [0.1, 9.5] & 0.4 & [0.0, 0.8] & 103.2 \\ \bottomrule
    \end{tabular}}
\caption{\textbf{Subnetwork discovery results for different percentages of upper layers masked in GPT-2 small,} averaged over four KGs and two seeds with [min, max] values denoted in brackets. The arrows ($\uparrow$,$\downarrow$) show the desired value for the metric.}
\label{tab:mask-layer-choice}
\end{table*}

\paragraph{Mask Implementation}
As mentioned in \S\ref{sec:eval-setup}, during mask learning, we do not mask the embedding, language modeling head, layer-normalization, and bias parameters. We also only learn masks for the top 50\% of the transformer layers.
We initialize the mask parameters such that, in the first forward pass, each model parameter has a starting masking probability of $\sigmoid(\logit_i)=0.45$, meaning the search is expected to start with an empty knowledge-critical subnetwork (\ie, a subnetwork mask of zeros) and a fully-connected inverse subnetwork (\ie, the full model). Results on a hyperparameter search for initialization can be found in Table~\ref{tab:mask-init}. Moreover, for the randomly masked baseline, we mask each module (\eg, MLP module at layer 8) at the same sparsity as the corresponding module in the critical subnetwork, which means that the masking is not uniformly done across all layers. For neuron masking, we jointly learn a mask across weights in a linear layer that connect to the same input neuron. For the randomly masked neuron baseline, we mask each module at the same \textit{neuron} sparsity as the corresponding module in the critical subnetwork.

\paragraph{Hyperparameters}
We use a learning rate of 0.2 with a linear warmup for the first 10\% of the training that starts from $1e$-$10$. We optimize with the AdamW optimizer. For equation~\ref{eq:objective}, we set $\lambda_{1}=1.5$ and $\lambda_{2}=\lambda_{3}=1$ in all of our our experiments. To encourage the subnetwork to be sparser, we schedule $\lambda_4$ to start at 2 and increase linearly after 50\% of the training until it reaches 3. For GPT2-small, we use a single GPU setting to run the mask training for 40,000 steps. For GPT2-medium and large, we use a three GPU distributed setting and run the mask training for 50,000 steps. For GPT2-XL, we use a three GPU distributed setting and run the mask training for 60,000 steps.

\paragraph{Software and Hardware}
We primarily use PyTorch\footnote{\url{https://pytorch.org}} and Huggingface Transformers\footnote{\url{https://huggingface.co/docs/transformers}} to implement the masking method. Experiments for GPT2-small, medium and large are run on NVIDIA A100 40GB devices. Experiments for GPT2-XL are run on NVIDIA A100 80GB devices.

\paragraph{Loss Trade-Off Analysis}
A primary driver of the knowledge-critical subnetwork search is the trade-off between the suppression and maintenance losses. To validate our $\lambda_i$ choices, we run a minimal experiment on giving importance to one objective at a time for two \targetkg{}s and one random seed. Specifically, when we set any one of the weights in Eq.~\ref{eq:objective} to a value of 3, we set the value of the rest to 1. As seen in Table~\ref{tab:loss_tradeoff}, we find that giving more weight to the \forgetting{} loss finds checkpoints with higher perplexity differences on \targetkg{} while simultaneously satisfying the maintenance criteria. Moreover, giving more weight to the sparsity regularization ensures a higher sparsity. These results support the $\lambda_i$ hyperparameters we use in all of our experiments, as described above.

\paragraph{Dataloaders}
As each \targetkg{} is small, at each gradient step, the model sees the complete graph. Therefore, the \targetkg{} batch size is the same as the number of triplets (see Table~\ref{tab:kg-stats}). In contrast, \controlkg{} and \lmodeling{} datasets have thousands of entries in total. To balance the learning and make it more efficient, we create a dynamic cyclical training dataloader that samples a new batch at each step without replacement. When the dataloader reaches the end of the dataset, it restarts with a new ordering. Please refer to Table~\ref{tab:batch-size} for the exact batch sizes.

\paragraph{Best Checkpoint Selection}
We iteratively select the best checkpoint, starting with strict criteria on the maintenance datasets and gradually loosening them. We check whether any checkpoints satisfy the first set of criteria limits shown in Table~\ref{tab:ckpnt-select}. The checkpoints need to have a \targetkg{} \deltappl{} above the mentioned floor and maintenance \deltappl{} below the mentioned ceiling. If the set of checkpoints retrieved is empty, we select from the next set of limits. If none of the iterations are successful, we pick the last checkpoint as the best one.

\section{Masked Layer Choice Study}
\label{sec:layer-study}

Layer-wise model probing analyses have shown that the first layers of transformer language models encode representations crucial for low-level linguistic tasks and features that may be a prerequisite for knowledge modeling \cite{tenney-etal-2019-bert, liu-etal-2019-linguistic}. Researchers have also shown that knowledge is not only contained in the final few layers \cite{singh-etal-2020-bertnesia}. Therefore, for our datasets, we investigate how masking different percentages of upper dense layers can affect the success criteria defined for a knowledge-critical subnetwork. In particular, we look at the effect of masking the top 25\%, 50\%, 75\%, and 100\% of the model.

In Table~\ref{tab:mask-layer-choice}, we observe that masking all dense layers in transformer blocks (100\%) can affect the maintenance criteria significantly. \controlkg{} perplexity difference is smaller when masking fewer layers, confirming that lower layers may have imperative representation to knowledge modeling. As the values for the different criteria are similar for masking the top 25\% and 50\%, we use the top 50\% masking approach to increase the masking coverage for all of our experiments.

\section{Additional Subnetwork Discovery Results}
\label{sec:additional-criteria-res}

\begin{table*}[t]
\centering
\resizebox{1.0\linewidth}{!}{
    \begin{tabular}{cl r@{\hspace{2pt}}l r@{\hspace{2pt}}l r@{\hspace{2pt}}l r@{\hspace{2pt}}l}
        \toprule
        \multicolumn{2}{c}{\multirow{2}{*}{\textbf{Knowledge Graph}}}    & \multicolumn{2}{c}{\textbf{Sparsity}} & \multicolumn{2}{c}{\textbf{\targetkg{}}} & \multicolumn{2}{c}{\textbf{\controlkg{}}} & \multicolumn{2}{c}{\textbf{\lmodeling{}}}  \\
                                    && \multicolumn{2}{c}{($\uparrow$)}      & \multicolumn{2}{c}{\textbf{\deltappl{}($\uparrow$)}}  & \multicolumn{2}{c}{\textbf{\deltappl{}} ($\downarrow$)} & \multicolumn{2}{c}{\textbf{\deltappl{} ($\downarrow$)}} \\ \midrule
        \multirow{9}{*}{WordNet} &\texttt{building} 
                                 & 98.4  & [97.4, 99.3] & 62.3   & [13.2, 114.1]   & -2.0 & [-7.0, 2.4] & 0.6 & [0.3, 1.0] \\ 
        &\texttt{communication}  & 99.2  & [99.0, 99.3] & 104.8  & [61.1, 165.9]   & -1.2 & [-2.2, 0.0] & 0.3 & [0.3, 0.3] \\ 
        &\texttt{change}         & 98.4  & [98.0, 99.1] & 567.2  & [38.7, 1405.6]  & 0.6  & [-1.6, 3.0] & 0.7 & [0.4, 0.9] \\ 
        &\texttt{statement}      & 98.2  & [96.3, 99.2] & 152.5  & [53.5, 248.7]   & -0.5 & [-3.2, 2.8] & 0.8 & [0.3, 1.8] \\  
        &\texttt{location}       & 99.0  & [98.8, 99.1] & 810.5  & [469.2, 1200.7] & 0.5  & [-1.7, 3.9] & 0.3 & [0.3, 0.4] \\ 
        &\texttt{representation} & 98.1  & [97.1, 98.8] & 221.8  & [115.5, 334.4]   & 2.9  & [0.6, 4.0] & 0.6 & [0.4, 1.0] \\ 
        &\texttt{magnitude}      & 99.0  & [98.6, 99.3] & 2216.9 & [1730.7, 2665.1] & -1.8 & [-2.6, -0.9] & 0.3 & [0.2, 0.4] \\ \cmidrule{2-10}
        &Random Weights          & 98.6  & [98.1, 99.2] & 24.3   & [5.0, 48.8]      & 14.6 & [0.0, 46.2]  & 2.2 & [1.2, 3.3] \\
        &Average                 & 98.6  & [98.1, 99.2]  & 590.9  & [62.3, 2216.9] & -0.2 & [-2, 2.9]	  & 0.5 & [0.0, 0.8] \\ \bottomrule\toprule
        \multirow{5}{*}{ConceptNet}
        &\texttt{fruit}         & 99.2  & [99.1, 99.4]  & 743.9  & [300.8, 1462.1] & 3.0 & [-0.6, 5.0] & 0.2 & [0.2, 0.2] \\ 
        &\texttt{sun}           & 99.2  & [99.0, 99.3]  & 888.4  & [521.0, 1240.1] & 3.2 & [2.0, 4.7]  & 0.2 & [0.1, 0.3] \\ 
        &\texttt{swimming}      & 99.0  & [98.8, 99.2]  & 276.8  & [240.9, 335.4]  & 2.3 & [0.6, 3.3]  & 0.3 & [0.2, 0.4] \\ \cmidrule{2-10}
        & Random Weights        & 99.1  & [99.0, 99.2]  & 21.0   & [13.7, 29.4]   & 14.6 & [12.4, 17.2] & 1.5 & [1.3, 1.7] \\
        & Average               & 99.1  & [99.0, 99.2]  & 636.4  & [276.8, 888.4]  & 2.8 & [2.3, 3.2]  & 0.2 & [0.2, 0.3] \\
        \bottomrule
    \end{tabular}
}
\caption{\textbf{Subnetwork discovery results for GPT-2 small with weight masking,} averaged over three seeds with [min, max] values denoted in brackets. \deltappl{} = PPL($\inverseprob$) - PPL($\plmprob$). The arrows ($\uparrow$,$\downarrow$) show the desired value for the metric. Random is an average of randomly masked baselines at the same sparsity levels as the discovered knowledge-critical subnetworks for each KG-seed pair.}
\label{tab:success-criteria-boundaries}
\vspace{-15pt}
\end{table*}

\begin{table*}[t]
\centering
\resizebox{1.0\linewidth}{!}{
    \begin{tabular}{cl r@{\hspace{2pt}}l r@{\hspace{2pt}}l r@{\hspace{2pt}}l r@{\hspace{2pt}}l}
        \toprule
        \multicolumn{2}{c}{\multirow{2}{*}{\textbf{Knowledge Graph}}}    & \multicolumn{2}{c}{\textbf{Sparsity}} & \multicolumn{2}{c}{\textbf{\targetkg{}}} & \multicolumn{2}{c}{\textbf{\controlkg{}}} & \multicolumn{2}{c}{\textbf{\lmodeling{}}}  \\
                                    && \multicolumn{2}{c}{($\uparrow$)}      & \multicolumn{2}{c}{\textbf{\deltappl{}($\uparrow$)}}  & \multicolumn{2}{c}{\textbf{\deltappl{}} ($\downarrow$)} & \multicolumn{2}{c}{\textbf{\deltappl{} ($\downarrow$)}} \\ \midrule
        \multirow{9}{*}{WordNet} &\texttt{building.n.01}         
        & 95.3 & [95.2, 95.4] & 330.7 & [268.7, 402.1] & 31.1 & [20.1, 42.6] & 4.3 & [4.1, 4.5] \\ 
        &\texttt{communication.n.02}    & 95.2  & [95.0, 95.4] & 109.2 & [70.9, 143.0] & 13.4 & [12.9, 14.0] & 3.9 & [3.9, 3.9] \\ 
        &\texttt{change.n.01}           & 95.1 & [95.0, 95.2] & 1328.6 & [1197.6, 1491.7] & 23.6 & [13.4, 34.4] & 4.3 & [4.3, 4.5] \\ 
        &\texttt{statement.n.01}        & 95.4 & [95.0, 96.0] & 494.2 & [281.7, 679.5] & 20.5 & [6.0, 36.1] & 4.1 & [3.6, 4.6] \\ 
        &\texttt{location.n.01}         & 95.4 & [95.3, 95.5] & 425.3 & [302.9, 548.0] & 32.6 & [18.0, 43.4] & 4.3 & [4.0, 4.7] \\ 
        &\texttt{representation.n.02}   & 95.0 &[94.9, 95.1] & 653.6 & [426.2, 934.5] & 20.9 & [10.8, 31.6] & 4.1 & [3.9, 4.2] \\ 
        &\texttt{magnitude.n.01}        & 95.5& [95.4, 95.5] & 1669.5 & [1181.7, 2538.6] & 13.6 & [12.2, 14.6] & 3.9 & [3.8, 4.0] \\ \cmidrule{2-10}
        & Random Neurons          & 95.3   & [95, 95.5]  & 23.8   & [-6.9, 73.8]    & 8.3  & [-2.1, 26.2]     & 8.3 & [7.2, 9.6]  \\
        & Average                 & 95.3   & [95, 95.5]  & 715.9  & [109.2, 1669.5] & 22.2 & [13.4, 32.6]	  & 4.1 & [3.9, 4.3] \\ \bottomrule\toprule
        \multirow{5}{*}{ConceptNet}
        &\texttt{fruit}         & 95.3 & [95.2, 95.4] & 31616.9 & [28471.3, 34422.3] & 61.4 & [58.6, 66.6] & 4.8 & [4.3, 5.1] \\ 
        &\texttt{sun}           & 95.4 & [95.4, 95.5] & 23980.0 & [23067.1, 25624.6] & 77.5 & [69.8, 86.8] & 5.0 & [4.8, 5.2] \\ 
        &\texttt{swimming}      & 93.9 & [93.8, 94.0] & 11669.2 & [9968.3, 12610.2] & 76.8 & [65.2, 91.5] & 6.4 & [5.7, 6.9] \\ \cmidrule{2-10}
        & Random Neurons        & 94.9   & [93.9, 95.4]  & 110.7    & [60.7, 177.5]      & 70.4 & [51.5, 107.2]   & 11.2 & [9.1, 14.7] \\
        & Average               & 94.9   & [93.9, 95.4]  & 22422.0  & [11669.2, 31616.9] & 71.9 & [61.4, 77.5]	  & 5.4 & [4.8, 6.4] \\
        \bottomrule
    \end{tabular}
}
\caption{\textbf{Subnetwork discovery results for GPT-2 small with neuron masking,} averaged over three seeds with [min, max] values denoted in brackets. \deltappl{} = PPL($\inverseprob$) - PPL($\plmprob$). The arrows ($\uparrow$,$\downarrow$) show the desired value for the metric. Random is an average of randomly masked baselines at the same sparsity levels as the discovered knowledge-critical subnetworks for each KG-seed pair.}
\label{tab:success-criteria-boundaries-neuron}
\vspace{-15pt}
\end{table*}
\begin{table*}
\centering
\resizebox{1.0\linewidth}{!}{
    \begin{tabular}{cl r@{\hspace{2pt}}l r@{\hspace{2pt}}l r@{\hspace{2pt}}l r@{\hspace{2pt}}l}
        \toprule
        \multicolumn{2}{c}{\multirow{2}{*}{\textbf{Knowledge Graph}}}  & \multicolumn{2}{c}{\textbf{\targetkg{}}} & \multicolumn{2}{c}{\textbf{\controlkg{}}} & \multicolumn{2}{c}{\textbf{\targetkg{}}} & \multicolumn{2}{c}{\textbf{\controlkg{}}}  \\
        && \multicolumn{2}{c}{\textbf{\deltarank{} ($\uparrow$)}}  & \multicolumn{2}{c}{\textbf{\deltarank{}} ($\downarrow$)} & \multicolumn{2}{c}{\textbf{$\Delta$ LogProb ($\downarrow$)}}  & \multicolumn{2}{c}{\textbf{$\Delta$ LogProb} ($\uparrow$)} \\ \midrule
        \multirow{9}{*}{WordNet} 
        & \texttt{building.n.01} & 83.7 & [12.8, 168.3] & 1.1 & [-1.1, 2.9] & -0.7 & [-1.2, -0.2] & 0.0 & [0.0, 0.1] \\ 
        & \texttt{communication.n.02} & 117.0 & [94.5, 134.9] & 0.6 & [0.1, 1.0] & -0.7 & [-1.0, -0.5] & 0.0 & [0.0, 0.0] \\ 
        & \texttt{change.n.01} & 139.1 & [0.4, 409.8] & 0.4 & [0.1, 0.6] & -1.4 & [-2.6, -0.3] & 0.0 & [0.0, 0.0] \\ 
        & \texttt{statement.n.01} & 154.5 & [1.6, 353.8] & 0.8 & [-0.6, 2.8] & -0.6 & [-0.9, -0.3] & 0.0 & [0.0, 0.0] \\ 
        & \texttt{location.n.01} & 344.9 & [188.4, 527.6] & 3.6 & [2.8, 5.0] & -1.6 & [-2.0, -1.2] & 0.0 & [-0.1, 0.0] \\ 
        & \texttt{representation.n.02} & 38.1 & [12.8, 57.8] & 3.4 & [2.7, 4.4] & -0.7 & [-1.0, -0.4] & 0.0 & [-0.1, 0.0] \\ 
        & \texttt{magnitude.n.01} & 1368.7 & [978.0, 1698.2] & 0.0 & [-0.3, 0.1] & -2.1 & [-2.3, -1.9] & 0.0 & [0.0, 0.0] \\ \cmidrule{2-10}
        &Random                  & 12.0  & [-0.1, 25.5] & 2.7  & [-0.1, 8.1] & -0.1 & [-0.3, 0.0] & -0.2 & [-0.5, 0.0] \\
        &Average                 & 320.9 & [38.1, 1368.7] & 1.4 & [0.0, 3.6] & -1.1 & [-2.1, -0.6] & 0.0 & [0.0, 0.0] \\ \bottomrule\toprule
        \multirow{5}{*}{ConceptNet}
        &\texttt{fruit}         & 1164.9 & [98.9, 2880.1] & 1.8 & [0.1, 3.5] & -1.0 & [-1.6, -0.6] & 0.0 & [0.0, 0.0] \\ 
        &\texttt{sun}           & 331.7 & [225.9, 415.8] & 1.6 & [1.4, 1.7] & -1.2 & [-1.4, -0.9] & 0.0 & [0.0, 0.0] \\ 
        &\texttt{swimming}      & 411.6 & [34.5, 685.6] & 1.4 & [0.8, 1.9] & -0.4 & [-0.5, -0.4] & 0.0 & [0.0, 0.0] \\ \cmidrule{2-10}
        &Random                  & 11.4 & [2.1, 20.0] & 5.5 & [4.2, 7.3] & 0.0 & [-0.1, 0.0] & -0.1 & [-0.1, -0.1] \\
        &Average                 & 636.1 & [331.7, 1164.9] & 1.6 & [1.4, 1.8] &	-0.9 & [-1.2, -0.4] & 0.0 & [0.0, 0.0] \\ \bottomrule
    \end{tabular}
}
\caption{\textbf{Subnetwork discovery rank and log probability results for GPT-2 small with weight masking,} averaged over three seeds. $\Delta$ Metric = Metric($\inverseprob$) - Metric($\plmprob$) for Rank and LogProb. Random is an average of randomly masked baselines at the same sparsity levels as the discovered knowledge-critical subnetworks for each KG-seed pair. Note that non-zero values may be rounded to 0.0 as we round to one decimal place. Individual KG results for the random baseline are in \ref{tab:success-criteria-random}.}
\label{tab:success-rank-prob}
\end{table*}

\begin{table*}[t]
\centering
\resizebox{1.0\linewidth}{!}{
    \begin{tabular}{cl r@{\hspace{2pt}}l r@{\hspace{2pt}}l r@{\hspace{2pt}}l r@{\hspace{2pt}}l}
        \toprule
        \textbf{Model} & \thead{\multirow{2}{*}{\textbf{Knowledge Graph}}} &  \multicolumn{2}{c}{\textbf{Sparsity}} &  \multicolumn{2}{c}{\textbf{\targetkg{}}} &  \multicolumn{2}{c}{\textbf{\controlkg{}}} &  \multicolumn{2}{c}{\textbf{\lmodeling{}}}  \\
        \textbf{Size} &  &  \multicolumn{2}{c}{($\uparrow$)}    &  \multicolumn{2}{c}{\textbf{\deltappl{} ($\uparrow$)}}  &  \multicolumn{2}{c}{\textbf{\deltappl{} ($\downarrow$)}} &  \multicolumn{2}{c}{\textbf{\deltappl{} ($\downarrow$)}} \\ \midrule
        & \texttt{communication.n.02} & 99.5 & [99.4, 99.7] & 139.9 & [-2.5, 282.3] & -0.1 & [-1.5, 1.3] & 0.1 & [0.0, 0.1] \\ 
        & \texttt{location.n.01} & 95.0 & [94.2, 95.8] & 432.2 & [48.0, 816.3] & 3.7 & [3.5, 3.8] & 0.8 & [0.8, 0.9] \\ 
        Medium
        & \texttt{representation.n.02}         & 94.8 & [91.9, 97.7] & 194.6 & [139.4, 249.8] & 4.0 & [3.8, 4.2] & 1.2 & [0.5, 1.8] \\ \cmidrule{2-10}
        & Random                               & 96.4 & [94.8, 99.5] & 32.1 & [5.0, 55.6] & 9.2 & [1.8, 15.9] & 3.0 & [0.3, 4.9] \\ 
        & Average                              & 96.4 & [94.8, 99.5] & 255.6 & [139.9, 432.2] & 2.5 & [-0.1, 4.0] & 0.7 & [0.1 , 1.2] \\ 
        \bottomrule \toprule
        & \texttt{communication.n.02} & 95.9 & [95.4, 96.4] & 2013.1 & [144.5, 3881.8] & 6.8 & [3.6, 10.0] & 0.6 & [0.5, 0.6] \\ 
        & \texttt{location.n.01} & 99.6 & [99.4, 99.7] & 1963.1 & [1543.1, 2383.1] & 1.9 & [0.6, 3.3] & 0.0 & [-0.0, 0.0] \\ 
        Large 
        & \texttt{representation.n.02}         & 99.2 & [99.1, 99.4] & 13363.6 & [3042.2, 23685.0] & 0.9 & [0.2, 1.7] & 0.0 & [0.0, 0.1] \\ \cmidrule{2-10}
        & Random                               & 98.2 & [95.9, 99.6] & 6.8 & [4.8, 7.8] & 2.9 & [0.7, 7.3] & 0.8 & [0.2, 2.1] \\ 
        & Average                              & 98.2 & [95.9, 99.6] & 5779.9 & [1963.1, 13363.6] & 3.2 & [0.9, 6.8] & 0.2 & [0.0, 0.6] \\  
        \bottomrule \toprule
        & \texttt{communication.n.02} & 99.3 & [99.2, 99.5] & 562.6 & [318.7, 806.5] & 3.2 & [1.6, 4.7] & 0.0 & [-0.0, 0.0] \\  
        & \texttt{location.n.01} & 99.4 & [99.1, 99.7] & 257.0 & [141.1, 372.8] & 2.8 & [1.2, 4.3] & -0.0 & [-0.0, -0.0] \\ 
        XL 
        & \texttt{representation.n.02}         & 99.2 & [99.1, 99.2] & 789.9 & [696.3, 883.5] & 3.6 & [3.1, 4.1] & 0.1 & [0.1, 0.1] \\ \cmidrule{2-10}
        & Random                               & 99.3 &  [99.2, 99.4] & 1.6  & [1.0, 2.4] & 0.1 & [-0.6, 1.2] & 0.1 & [0.1, 0.2] \\ 
        & Average                              & 99.3 &  [99.2, 99.4] & 536.5 & [257.0, 789.9.6] & 3.2 & [2.8, 3.6] & 0.0 & [0.0, 0.1] \\  \bottomrule
    \end{tabular}
}
\caption{\textbf{Subnetwork discovery results on larger models per KG with weight masking,} averaged over two seeds. Random is an average of randomly masked baselines at the same sparsity levels as the discovered knowledge-critical subnetworks for each KG-seed pair. Individual KG results for the random baseline are in Table~\ref{tab:success-criteria-random-scaled}.}
\label{tab:additional-success-criteria-scaled}
\vspace{-10pt}
\end{table*}

\begin{table*}[t]
\centering
\resizebox{1.0\linewidth}{!}{
    \begin{tabular}{ll r@{\hspace{2pt}}l r@{\hspace{2pt}}l r@{\hspace{2pt}}l r@{\hspace{2pt}}l}
        \toprule
        \multicolumn{2}{c}{\multirow{2}{*}{\textbf{Knowledge Graph}}}    & \multicolumn{2}{c}{\textbf{Sparsity}} & \multicolumn{2}{c}{\textbf{\targetkg{}}} & \multicolumn{2}{c}{\textbf{\controlkg{}}} & \multicolumn{2}{c}{\textbf{\lmodeling{}}}  \\
                                    && \multicolumn{2}{c}{($\uparrow$)}      & \multicolumn{2}{c}{\textbf{\deltappl{}($\uparrow$)}}  & \multicolumn{2}{c}{\textbf{\deltappl{}} ($\downarrow$)} & \multicolumn{2}{c}{\textbf{\deltappl{} ($\downarrow$)}} \\ \midrule
        \multirow{9}{*}{WordNet} 
        &\texttt{building} & 98.4 & [97.4, 99.3] & 5.8 & [3.0, 10.2] & 14.8 & [3.0, 26.2] & 2.8 & [1.0, 5.2] \\ 
        &\texttt{communication} & 99.2 & [99.0, 99.3] & 5.0 & [-2.4, 10.1] & 4.6 & [0.3, 7.6] & 1.2 & [1.0, 1.4] \\ 
        &\texttt{change} & 98.4 & [98.0, 99.1] & 33.9 & [25.7, 43.3] & 24.2 & [16.2, 36.2] & 2.0 & [1.3, 2.6] \\ 
        &\texttt{statement} & 98.2 & [96.3, 99.2] & 15.6 & [-0.3, 34.6] & 0.0 & [-3.5, 3.4] & 3.3 & [1.2, 6.8] \\   
        &\texttt{location} & 99.0 & [98.8, 99.1] & 18.8 & [-15.8, 55.8] & 0.2 & [-7.5, 4.6] & 1.6 & [1.3, 1.9] \\ 
        &\texttt{representation} & 98.1 & [97.1, 98.8] & 48.8 & [11.4, 80.3] & 46.2 & [30.6, 66.2] & 3.0 & [1.6, 4.5] \\
        &\texttt{magnitude} & 99.0 & [98.6, 99.3] & 41.9 & [21.1, 70.4] & 12.3 & [-0.2, 29.2] & 1.6 & [0.8, 2.0] \\  \cmidrule{2-10}
        &Average                 & 98.6 & [98.1, 99.2] & 24.3 & [5.0, 48.8] & 14.6 & [0.0, 46.2] & 2.2 & [1.2, 3.3] \\ \bottomrule\toprule
        \multirow{5}{*}{ConceptNet}
        &\texttt{fruit} & 99.2 & [99.1, 99.4] & 13.7 & [-12.6, 35.7] & 12.4 & [-0.6, 19.7] & 1.3 & [0.9, 1.7] \\ 
        &\texttt{sun} & 99.2 & [99.0, 99.3] & 29.4 & [11.0, 39.0] & 17.2 & [0.3, 36.3] & 1.5 & [1.1, 1.8] \\ 
        &\texttt{swimming} & 99.0 & [98.8, 99.2] & 19.8 & [-22.1, 42.6] & 14.1 & [2.3, 30.2] & 1.7 & [1.2, 2.4] \\ \cmidrule{2-10}
        & Average      & 99.1 & [99.0, 99.2] & 21.0 & [13.7, 29.4] & 14.6 & [12.4, 17.2] & 1.5 & [1.3, 1.7] \\
        \bottomrule
    \end{tabular}
}
\caption{\textbf{Subnetwork discovery results on the randomly masked baseline for GPT2-small weight masking,} averaged over three seeds.}
\label{tab:success-criteria-random}
\end{table*}

\begin{table*}[t]
\centering
\resizebox{1.0\linewidth}{!}{
    \begin{tabular}{cl r@{\hspace{2pt}}l r@{\hspace{2pt}}l r@{\hspace{2pt}}l r@{\hspace{2pt}}l}
        \toprule
        \textbf{Model} & \textbf{Knowledge}    &  \multicolumn{2}{c}{\textbf{Sparsity}} &  \multicolumn{2}{c}{\textbf{\targetkg{}}} &  \multicolumn{2}{c}{\textbf{\controlkg{}}} &  \multicolumn{2}{c}{\textbf{\lmodeling{}}}  \\
        \textbf{Size} & \textbf{Graph} &  \multicolumn{2}{c}{($\uparrow$)}    &  \multicolumn{2}{c}{\textbf{\deltappl{} ($\uparrow$)}}  &  \multicolumn{2}{c}{\textbf{\deltappl{} ($\downarrow$)}} &  \multicolumn{2}{c}{\textbf{\deltappl{} ($\downarrow$)}} \\ \midrule
        & \texttt{communication.n.02}          & 99.5 & [99.4, 99.7] & 5.0 & [1.9, 8.1] & 1.8 & [1.5, 2.1] & 0.3 & [0.3, 0.4] \\ 
        & \texttt{location.n.01}               & 95.0 & [94.2, 95.8] & 55.6 & [35.2, 76.1] & 15.9 & [11.6, 20.2] & 3.8 & [3.3, 4.4] \\ 
        Medium
        & \texttt{representation.n.02}         & 94.8 & [91.9, 97.7] & 35.8 & [18.0, 53.7] & 9.9 & [9.6, 10.3] & 4.9 & [1.5, 8.3] \\ \cmidrule{2-10}
        & Average                              & 96.4 & [94.8, 99.5] & 32.1 & [5.0, 55.6] & 9.2 & [1.8, 15.9] & 3.0 & [0.3, 4.9] \\ 
        \bottomrule \toprule
        & \texttt{communication.n.02} & 95.9 & [95.4, 96.4] & 7.8 & [7.2, 8.4] & 7.3 & [6.6, 8.0] & 2.1 & [1.8, 2.5] \\ 
        & \texttt{location.n.01} & 99.6 & [99.4, 99.7] & 4.8 & [3.6, 6.0] & 0.7 & [0.5, 0.9] & 0.2 & [0.1, 0.3] \\
        Large
        & \texttt{representation.n.02} & 99.2 & [99.1, 99.4] & 7.8 & [5.4, 10.1] & 0.8 & [-0.2, 1.8] & 0.2 & [0.2, 0.3] \\   \cmidrule{2-10}
        & Average                              & 98.2 & [95.9, 99.6] & 6.8 & [4.8, 7.8] & 2.9 & [0.7, 7.3] & 0.8 & [0.2, 2.1] \\ 
        \bottomrule \toprule
        & \texttt{communication.n.02} & 99.3 & [99.2, 99.5] & 1.0 & [0.3, 1.7] & -0.3 & [-0.3, -0.2] & 0.1 & [0.1, 0.1] \\ 
        & \texttt{location.n.01} & 99.4 & [99.1, 99.7] & 1.3 & [-0.7, 3.3] & -0.6 & [-1.2, 0.0] & 0.1 & [0.1, 0.1] \\ 
        XL
        & \texttt{representation.n.02} & 99.2 & [99.1, 99.2] & 2.4 & [-0.5, 5.4] & 1.2 & [0.6, 1.9] & 0.2 & [0.2, 0.2] \\ \cmidrule{2-10}
        & Average                         & 99.3 &  [99.2, 99.4] & 1.6  & [1.0, 2.4] & 0.1 & [-0.6, 1.2] & 0.1 & [0.1, 0.2] \\ \bottomrule
        
    \end{tabular}
}
\caption{\textbf{Subnetwork discovery results on larger randomly masked models per KG with weight masking,} averaged over two seeds.}
\label{tab:success-criteria-random-scaled}
\end{table*}

In this section, we provide additional metrics for subnetwork discovery results and non-aggregated results for the randomly masked baseline.

\paragraph{Minimum \& Maximum Boundaries}
In addition to the average \deltappl{} and \deltarank{} presented in Table~\ref{tab:success-criteria}, we add minimum and maximum boundaries to all of the results in Table~\ref{tab:success-criteria-boundaries} and \ref{tab:success-rank-prob}. We also provide log probability differences $\Delta$LogProb similar to how \deltappl{} is calculated. We observe in Table~\ref{tab:success-rank-prob} the same trend as \deltappl{}. On average, removing the subnetwork increases the rank of the gold tail token and decreases the log probability. In contrast, the randomly masked baseline does not increase the \targetkg{} rank significantly and does not maintain \controlkg{} rank to the same extent as the critical subnetwork.

\paragraph{Model Scale} 
We include the individual KG results for larger models in Table~\ref{tab:additional-success-criteria-scaled}. While individual results on GPT2-medium are not as sparse and effective as the small and large variants, it is still more significant than randomly masking the model at the same sparsity.

\paragraph{Randomly Masked Baseline}
We provide the non-aggregated randomly masked baseline results for GPT2-small in Table~\ref{tab:success-criteria-random} and for larger models in Table~\ref{tab:success-criteria-random-scaled}. We notice that KGs where the pretrained model perplexity is already low (see Table~\ref{tab:kg-stats}) seem not to be as affected by a random subnetwork removal as those that have a higher initial perplexity.

\section{Additional Details on Downstream Task Transfer}
\label{sec:additional-downstream}
\begin{table*}
\centering
\resizebox{0.75\linewidth}{!}{
    \begin{tabular}{clccccc}
        \toprule
        \textbf{Method}                & \textbf{Subnetwork} & \textbf{Dev} & \textbf{Test} & \textbf{Filtered} & \textbf{Half 1} & \textbf{Half 2}   \\ \midrule
        \multirow{5}{*}{\parbox{2cm}{\centering \textbf{Head Tuning}}} 
                        & Full                  & 38.63 & 38.33 & 37.19 & 37.94 & 36.44\\\cmidrule{2-7}
                        & Random (Half 1)       & \textcolor{red}{-0.87} & \textcolor{red}{-4.83} & \textcolor{red}{-4.89} & \textcolor{red}{-1.75} & \textcolor{red}{-8.00} \\ 
                        & Ours (Half 1)    & \textcolor{red}{-1.45} & \textcolor{red}{-4.83} & \textcolor{red}{-11.02} & \textcolor{red}{-15.11} & \textcolor{red}{-6.95} \\ \cmidrule{2-7}
                        & Random (Half 2)       & \textcolor{red}{-0.46} & \textcolor{red}{-4.16} & \textcolor{red}{-4.27} & \textcolor{red}{-2.30} & \textcolor{red}{-6.22} \\ 
                        & Ours (Half 2)    & \textcolor{red}{-1.99} & \textcolor{red}{-6.80} & \textcolor{red}{-8.61} & \textcolor{red}{-7.28} & \textcolor{red}{-9.93} \\ \midrule
        \multirow{5}{*}{\parbox{2cm}{\centering \textbf{LoRA}}} 
                        & Full                  & 50.04 & 48.64 & 48.67 & 48.44 & 48.90\\\cmidrule{2-7}
                        & Random (Half 1)       & \textcolor{red}{-1.39} & \textcolor{red}{-0.08} & \textcolor{red}{-1.84} & \textcolor{red}{-0.93} & \textcolor{red}{-2.75} \\ 
                        & Ours (Half 1)    & \textcolor{red}{-0.54} & \textcolor{red}{-2.33} & \textcolor{red}{-2.48} & \textcolor{red}{-2.95} & \textcolor{red}{-2.02}\\ \cmidrule{2-7}
                        & Random (Half 2)       & \textcolor{red}{-1.23} & \textcolor{red}{-0.96} & \textcolor{red}{-1.75} & \textcolor{red}{-0.93} & \textcolor{red}{-2.57}\\ 
                        
                        & Ours (Half 2)    & \textcolor{red}{-0.05} & \textcolor{red}{-1.93} & \textcolor{red}{-3.77} & \textcolor{red}{-3.14} & \textcolor{red}{-4.39} \\ \midrule
        \multirow{5}{*}{\parbox{2cm}{\centering \textbf{Full Finetuning}}} 
                        & Full                  & 44.61 & 42.33 & 42.79 & 44.01 & 41.58 \\\cmidrule{2-7}
                        & Random (Half 1)       & \textcolor{blue}{+0.08} & \textcolor{red}{-0.62} & \textcolor{red}{-0.36} & \textcolor{red}{-2.94} & \textcolor{blue}{+2.19} \\ 
                         & Ours (Half 1)    & \textcolor{blue}{+0.50} & \textcolor{red}{-1.34} & \textcolor{red}{-0.46} & \textcolor{red}{-5.33} & \textcolor{blue}{+4.39} \\ \cmidrule{2-7}
                        & Random (Half 2)       & \textcolor{red}{-1.01} & 0.00 & \textcolor{blue}{+0.74} & \textcolor{red}{-1.65} & \textcolor{blue}{+3.11} \\ 
                       
                        & Ours (Half 2)    & \textcolor{red}{-0.11} & \textcolor{red}{-0.75} & \textcolor{red}{-0.27} & \textcolor{red}{-0.92} & \textcolor{blue}{+0.36}\\ \midrule
    \end{tabular}}
\caption{\textbf{Accuracy on downstream CommonsenseQA task with GPT2-small and weight masking,} averaged over three seeds. Ours refers to removing the critical subnetwork. Random refers to removing a random subnetwork at the same sparsity as the critical subnetwork.}
\label{tab:additional-downstream}
\end{table*}

To learn a mask for a set of ConceptNet relations, we need to verbalize them with a relation-specific prompt. As described in \S\ref{sec:downstream}, CommonsenseQA questions are not explicitly annotated with a relation. However, they were constructed with ConceptNet such that each question's head concept relates to four of the tail answers with the same relation. This does not apply to the fifth answer, as crowd workers created them. Therefore, to retrieve the relations, we iterate through the questions and check if any relations with the question head concept and correct tail answer exist in the LAMA and Commonsense Knowledge Base Completion subsets of ConceptNet \cite{li-etal-2016-commonsense, petroni-etal-2019-language}. If it does and has only one relation, we choose that relation. If it has multiple relations, we take the union of relations between the head concept and the distractor tail answers and intersect that with the correct tail triplets. If the intersection is a set larger than one element, we choose one relation at random. Out of the 1221 test questions, only 572 have a single-token correct answer, and we could only find the corresponding relation to 363 questions, which is our filtered test set.

For the MCQA head, we use the Huggingface Double Heads model.\footnote{\url{https://huggingface.co/docs/transformers}} In addition to the language modeling head, this model adds a parallel multiple-choice classification head. The MCQA head takes as input the last sequence output. To finetune the MCQA model, we use three kinds of fine-tuning. The first one is \textbf{Head Tuning}, in which the model parameters are frozen, but the MCQA head is not. The second method is \textbf{LoRA} \cite{hu2022lora}, which is a parameter-efficient finetuning method. Similar to the head tuning method, LoRA freezes the model parameters and instead inserts trainable rank decomposition parameters in each transformer layer. We use a rank of 16 for all LoRA experiments. Finally, we also try \textbf{Full Finetuning}, in which all model parameters are tuned. To remove a subnetwork, we manually set the knowledge-critical parameters to 0. Therefore, the value of these parameters can change during full finetuning.

In addition, we also verify whether learning a mask for one randomly selected half of the filtered test set (Half 1) corrupts downstream task transfer for a distinct half (Half 2), where there are no triplet overlaps. We find in Table~\ref{tab:additional-downstream} that, on average, the accuracy on the triplets the mask was trained for is less by 3.6\% than the held-out half.

\section{Alternative Objective: Is expressing knowledge enough to be a knowledge-critical subnetwork?}
\label{sec:expression-loss}
\begin{table*}[t]
\resizebox{1.0\linewidth}{!}{
    \begin{tabular}{l r@{\hspace{2pt}}l r@{\hspace{2pt}}l r@{\hspace{2pt}}l r@{\hspace{2pt}}l}
    \toprule
    \thead{Objective}    & \multicolumn{2}{c}{\textbf{Sparsity}} & \multicolumn{2}{c}{\textbf{\targetkg{}}} & \multicolumn{2}{c}{\textbf{\controlkg{}}} & \multicolumn{2}{c}{\textbf{\lmodeling{}}}  \\
    \thead{Combination}  & \multicolumn{2}{c}{($\uparrow$)}      & \multicolumn{2}{c}{\textbf{\deltappl{} ($\uparrow$)}}  & \multicolumn{2}{c}{\textbf{\deltappl{} ($\downarrow$)}} & \multicolumn{2}{c}{\textbf{\deltappl{} ($\downarrow$)}} \\ \midrule
        Expression-only     & 99.8  & [99.7, 99.9] & 154.0 & [105.4, 181.2] & 83.5 & [-4.2, 234.9] & 4.0 & [2.6, 6.2] \\  
        Our Method + Expression   & 95.7 & [93.8, 96.7] & 909.2 & [107.2, 2421.7] & -0.5 & [-4.7, 5.1] & 1.0 & [0.8, 1.4] \\ \midrule
        Our Method  & 98.6 & [97.8, 99.1] & 378.1 & [74.3, 834.9] & 1.6 & [-0.7, 4.0] & 0.5 & [0.3, 0.8] \\  \bottomrule 
    \end{tabular}}
\caption{\textbf{Expression loss study with GPT2-small and weight masking,} averaged across three KGs and two seeds. Random is an average of randomly masked baselines at the same sparsity levels as the discovered knowledge-critical subnetworks for each KG-seed pair.}
\label{tab:expression}
\end{table*}

We defined \textit{knowledge-critical subnetworks} as being \textit{responsible} for a model's ability to express certain pieces of knowledge, validated by an increase in perplexity when that subnetwork is removed from the model. However, another way to extract a knowledge-critical subnetwork might be to learn a mask over the network that minimizes the negative loglikelihood of all $x \in \targetkg{}$:
\begin{equation}
    \Loss_{\text{express}} = - \sum_{x} \log(\subnetworkprob)
\end{equation}
In Table~\ref{tab:expression}, we compare subnetworks extracted in this manner (\ie, Expression-only) with those of our main method, as well as those of a combination of these objectives: $\Loss_{\text{final}} + \lambda_5 \Loss_{\text{express}}$. 
Interestingly, we find that the Expression-only setting can learn a mask for a \textit{highly} sparse subnetwork, which, when removed from the full model, also significantly increases perplexity on \targetkg{}. However, this subnetwork also struggles to maintain perplexity on \controlkg{}, indicating it may encode abilities crucial for expressing \textit{any} set of relational knowledge. Adding the expression loss to our joint objective mitigates this issue, but reduces subnetwork sparsity by a significant margin ($\sim$4\%), indicating that the Expression-only loss may discover spurious subnetworks that are not actually \textit{knowledge-critical} --- they are not \textit{responsible} for the expression of the knowledge when they are entangled in the full model, though their parameters may compute a function that expresses it.

\section{Spurious Subnetworks Test}
\label{sec:param-based}
\begin{figure*}[t]
    \centering
    \begin{subfigure}{0.3375\linewidth}
      \centering
      \includegraphics[width=\textwidth]{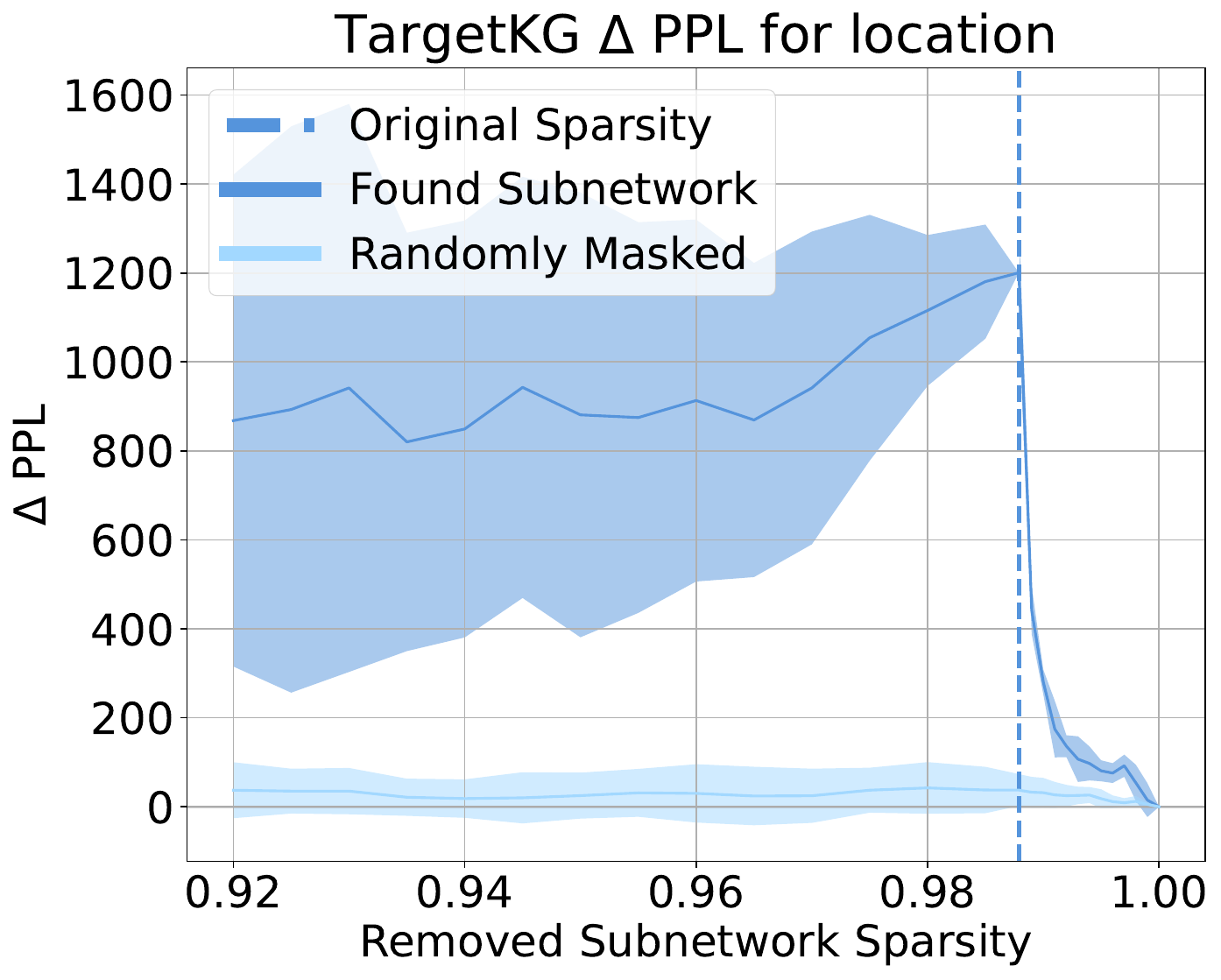}
    \end{subfigure}
    \begin{subfigure}{0.32\linewidth}
      \centering
      \includegraphics[width=\textwidth]{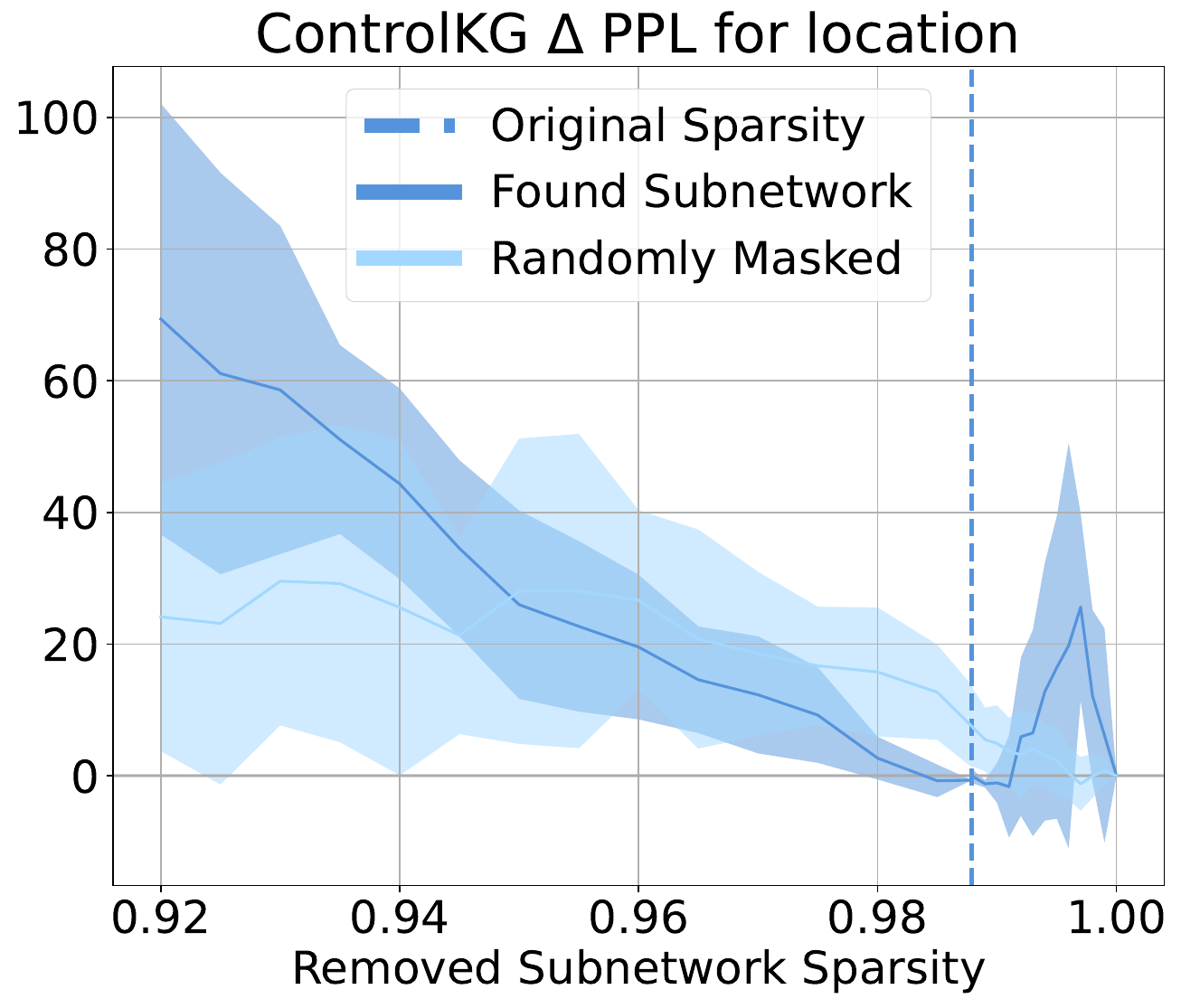}
    \end{subfigure}
    \begin{subfigure}{0.325\linewidth}
      \centering
      \includegraphics[width=\textwidth]{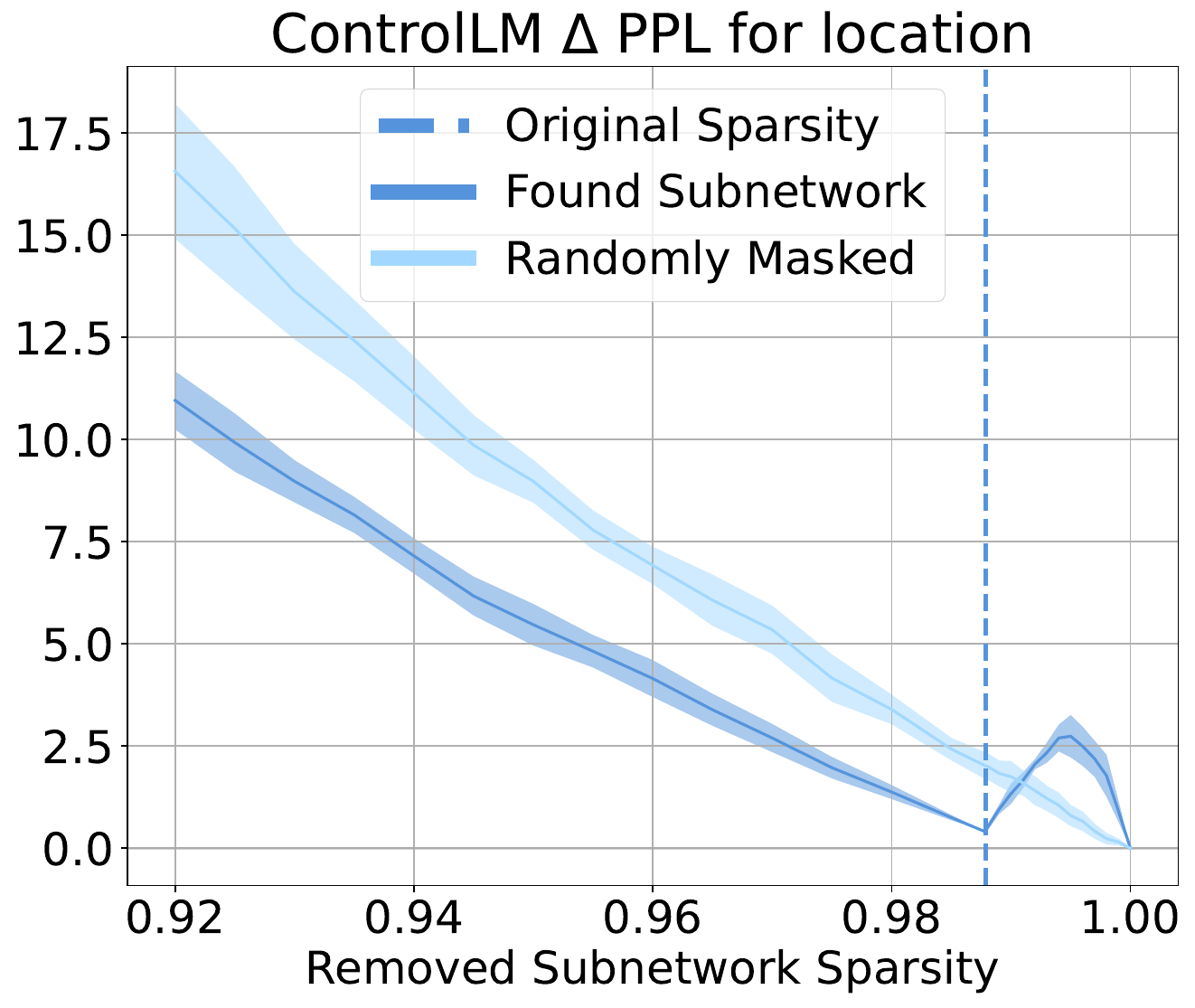}
    \end{subfigure}
    \begin{subfigure}{0.3375\linewidth}
      \centering
      \includegraphics[width=\textwidth]{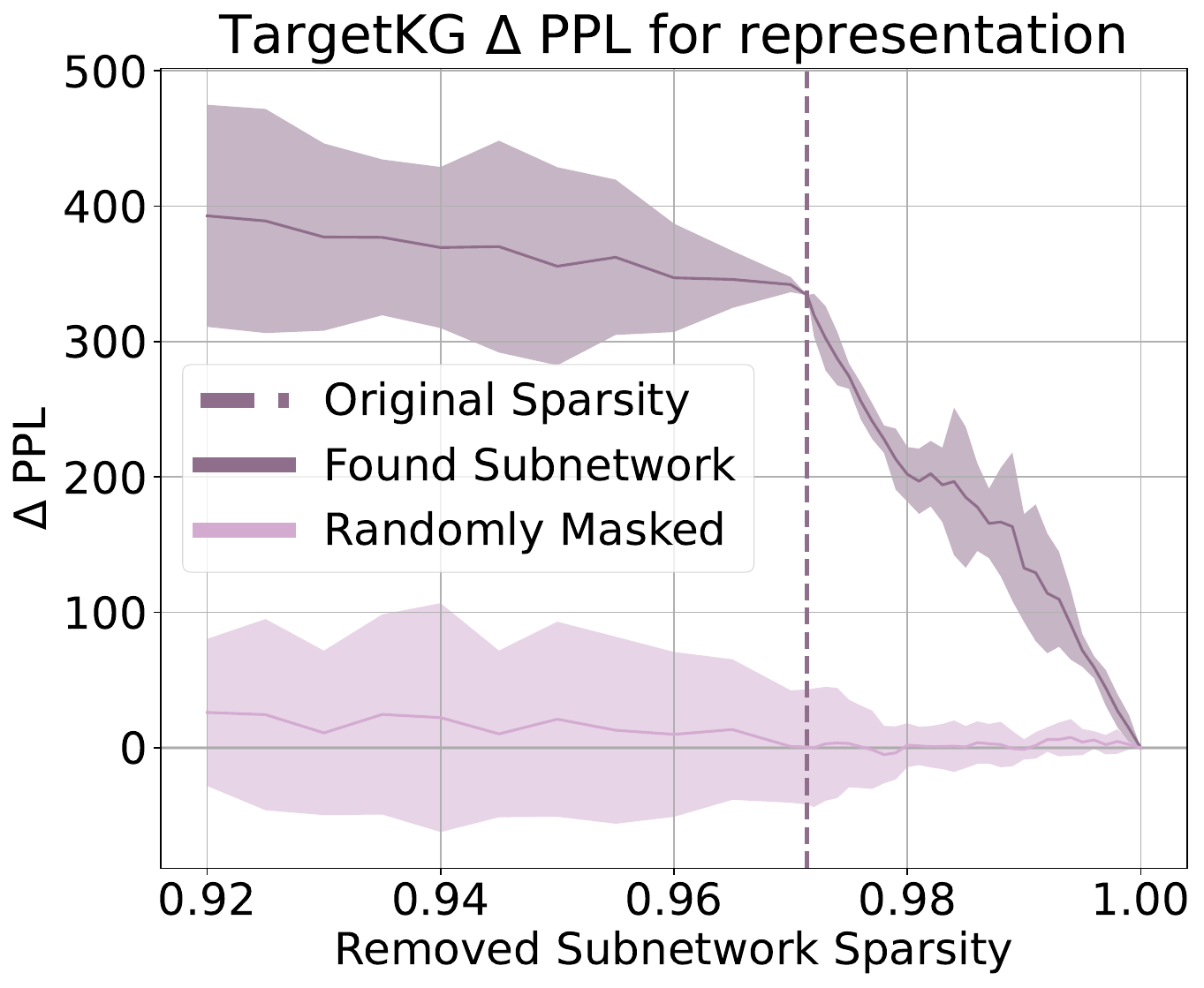}
    \end{subfigure}
    \begin{subfigure}{0.32\linewidth}
      \centering
      \includegraphics[width=\textwidth]{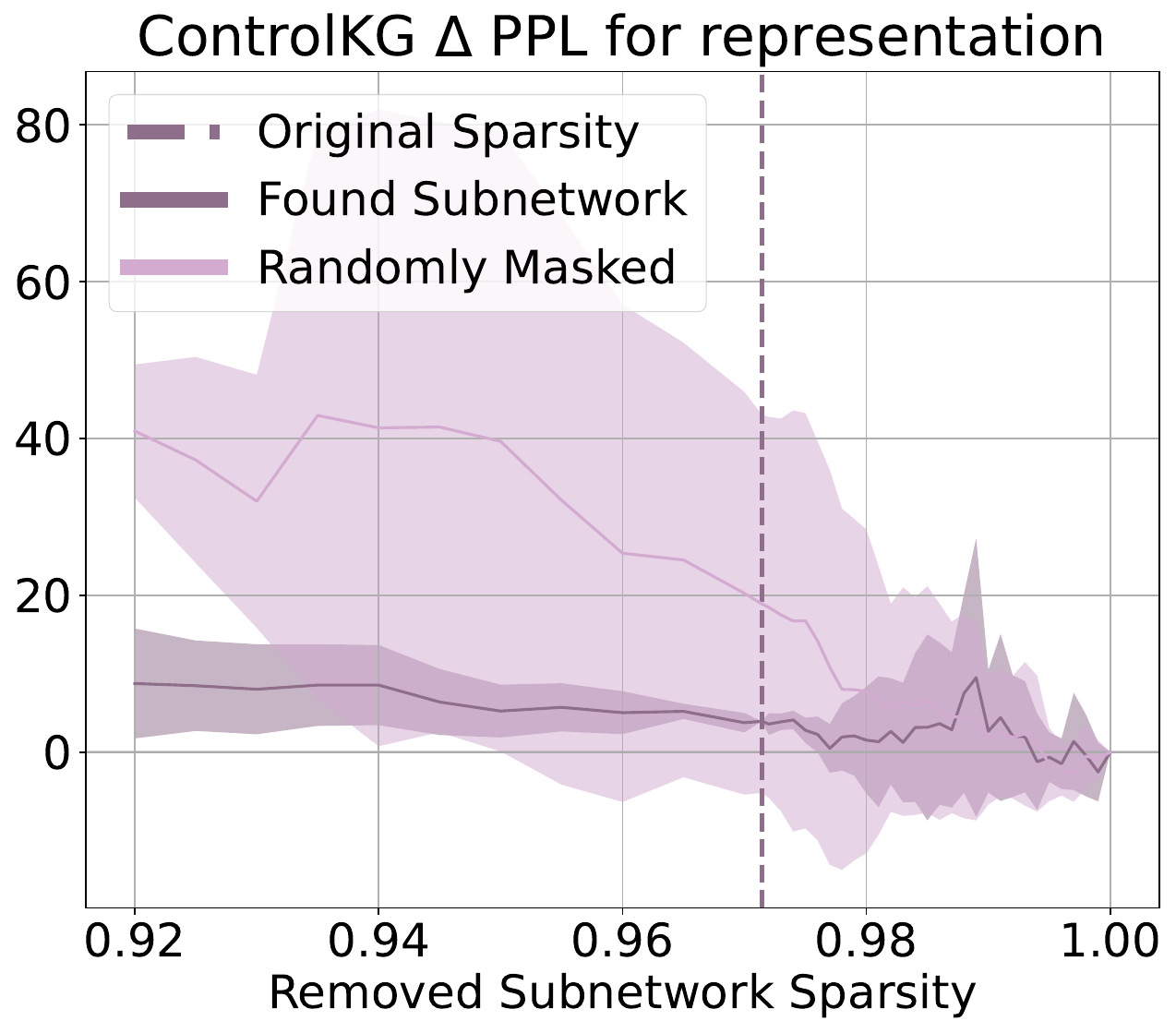}
    \end{subfigure}
    \begin{subfigure}{0.33\linewidth}
      \centering
      \includegraphics[width=\textwidth]{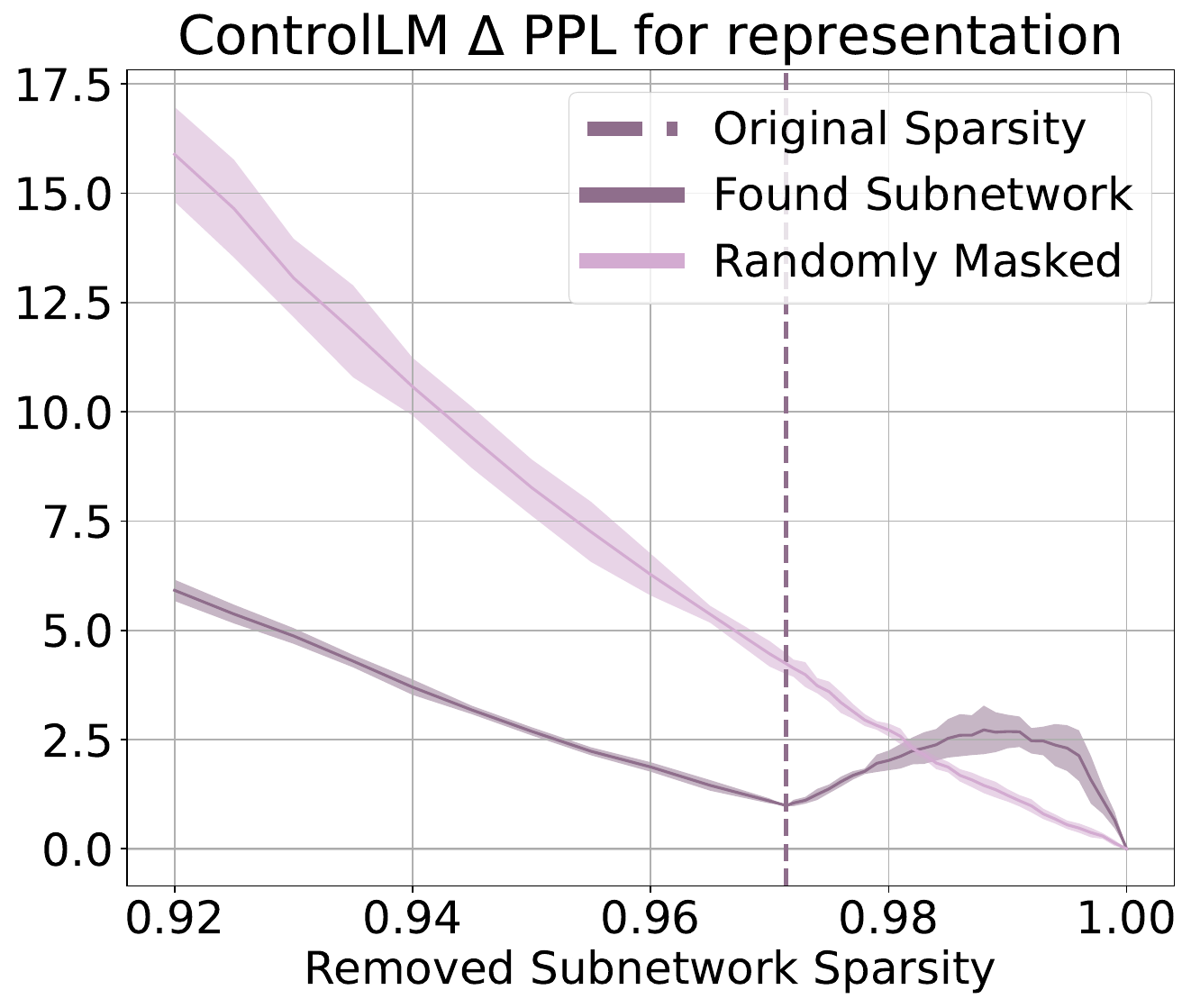}
    \end{subfigure}
    \begin{subfigure}{0.335\linewidth}
      \centering
      \includegraphics[width=\textwidth]{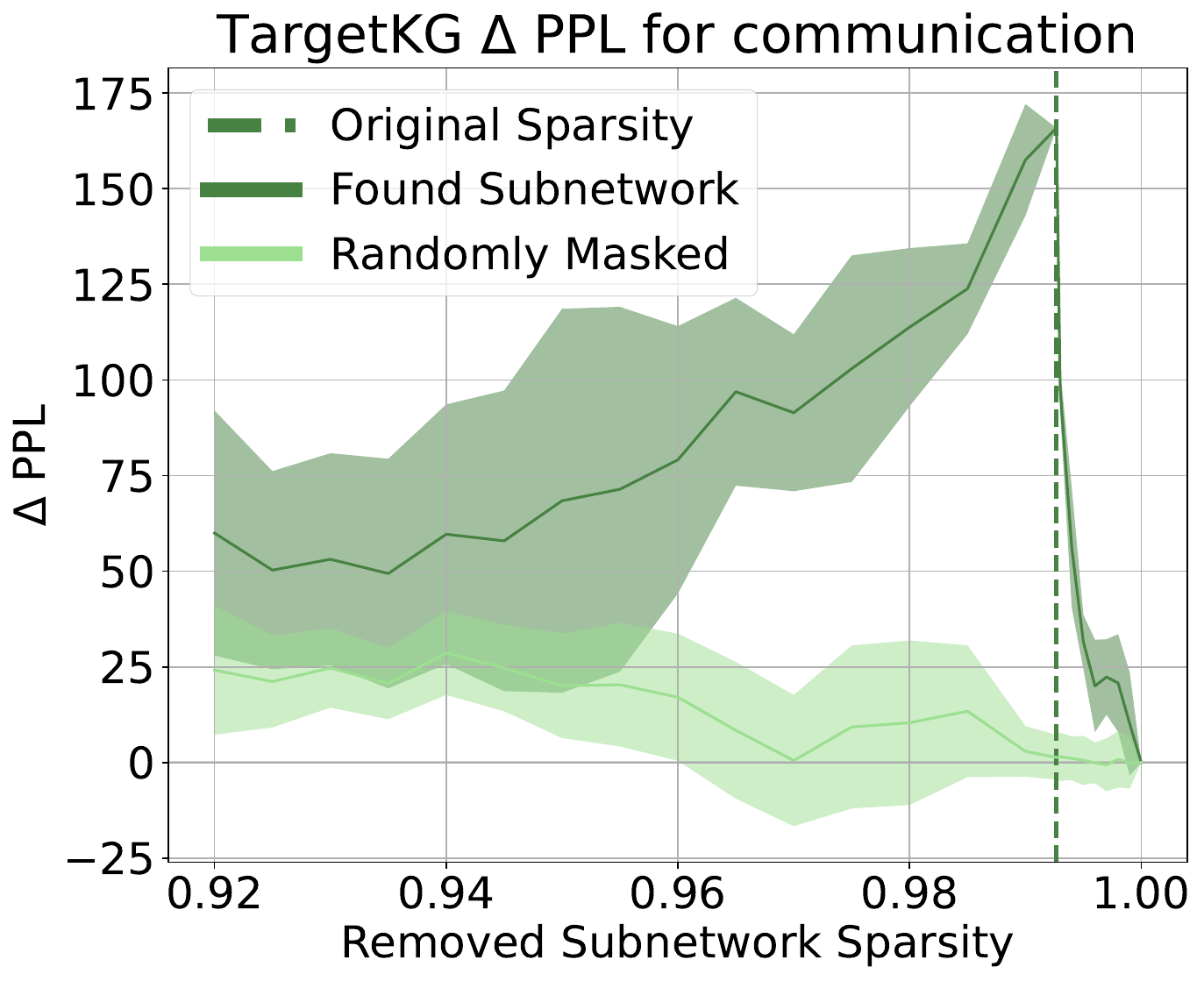}
    \end{subfigure}
    \begin{subfigure}{0.32\linewidth}
      \centering
      \includegraphics[width=\textwidth]{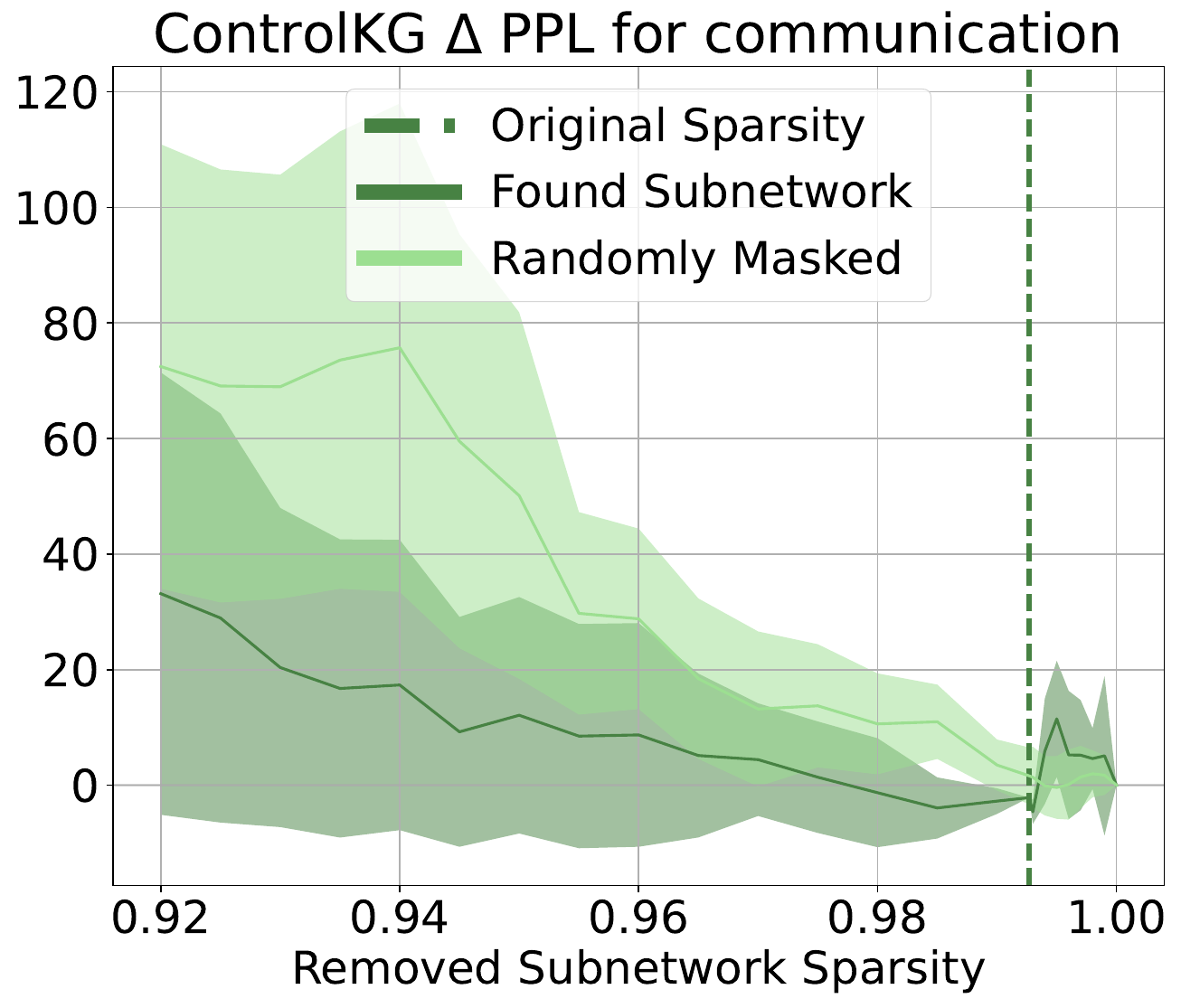}
    \end{subfigure}
    \begin{subfigure}{0.3225\linewidth}
      \centering
      \includegraphics[width=\textwidth]{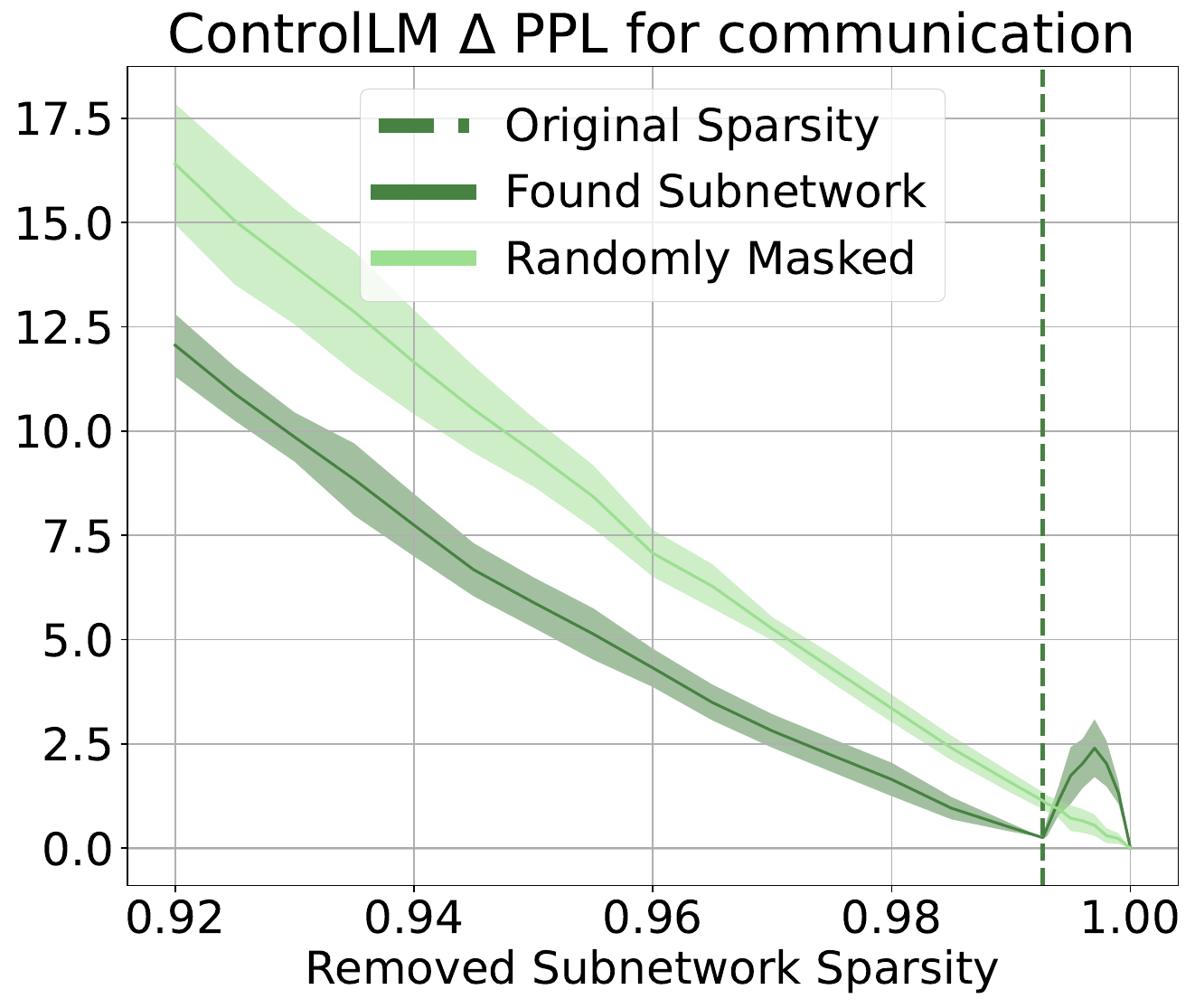}
    \end{subfigure}
    \caption{\textbf{Removing and adding parameters to the remaining GPT2-small model,} averaged over five seeds, with standard deviation depicted as the filled area around the average curves. The $x$-axis is the removed subnetwork sparsity. The $y$-axis is the \deltappl{} = PPL($\inverseprob$)  - PPL($\plmprob$) for the different datasets. Vertical dashed lines show the original sparsity of the critical subnetwork.
    The darker curve is the outcome starting from the critical subnetwork, whereas the lighter curve is from a randomly masked model at the same sparsity.}
    \label{fig:additional-param-stab}
\end{figure*}

We hypothesize that a spurious subnetwork would cause the remaining network from which it was removed to re-gain the ability to express \targetkg{} if the subnetwork was randomly \textit{expanded} (\ie, \deltappl{} on \targetkg{} would drop as more parameters are removed from $\inverseprob$). Meanwhile, if removing the critical subnetwork is not a spurious solution to suppress the \targetkg{}, then the remaining model would generally still fail to recognize \targetkg{}, even as more parameters were randomly removed, leading \deltappl{} to rise or stay the same.
To verify this hypothesis, we remove further parameters from the remaining model. Starting from the knowledge-critical subnetwork sparsity, we randomly remove parameters at intervals of 0.5\%. We run this iterative process of removing parameters with five different random seeds. We also test whether the mask has found a spurious solution to achieve the maintenance criteria by adding back parameters, though with smaller intervals of 0.1\%, as the starting sparsity level is typically high.

In Figure~\ref{fig:additional-param-stab}, we observe that removing more parameters in small amounts does not significantly recover expressing \targetkg{}. As a baseline, we plot the effect on \deltappl{} of removing further parameters from remaining models with randomly removed subnetworks of the same sparsity.
Interestingly, for the maintenance datasets, \deltappl{} for both datasets increases as we remove parameters from the remaining model. When we add back parameters, we do not see a linear recovery to \deltappl{} $= 0$. Instead, we observe an initial phase of increase followed by a phase of decrease as the model returns to its original state (\ie, a \deltappl{} of zero at 100\% sparsity). This effect can be explained by the fact that our subnetwork had been optimized to keep these abilities, and has been slightly overfit for maintenance, though not for \textit{suppression}. Thus, randomly adding parameters back yields new sub-optimal pathways that corrupt the model's original distribution.

\section{Structural Analysis}
\label{sec:structure-analysis}

\begin{figure*}[t]
    \centering
    \includegraphics[height=0.93\textheight]{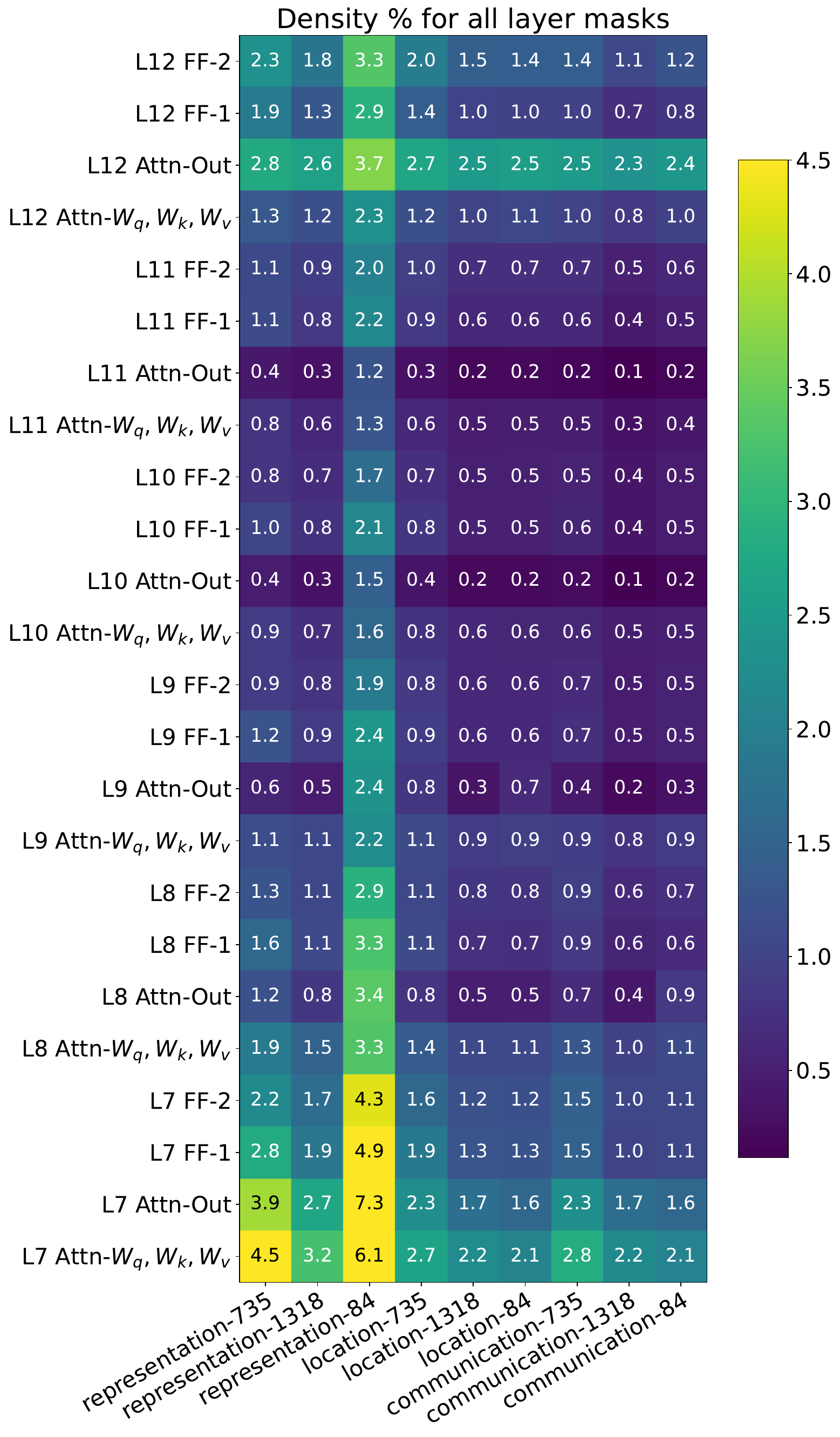}
    \caption{\textbf{Average module mask density with weight masking,} for different KGs ( \texttt{representation}, \texttt{location}, and \texttt{communication}) and seeds. Reported in percentage (\%). The brighter the color, the higher the removed mask density.}
    \label{fig:structure-layer}
\end{figure*}

\begin{figure*}[t!]
    \centering
    \begin{subfigure}{0.49\linewidth}
      \centering
      \includegraphics[width=\textwidth]{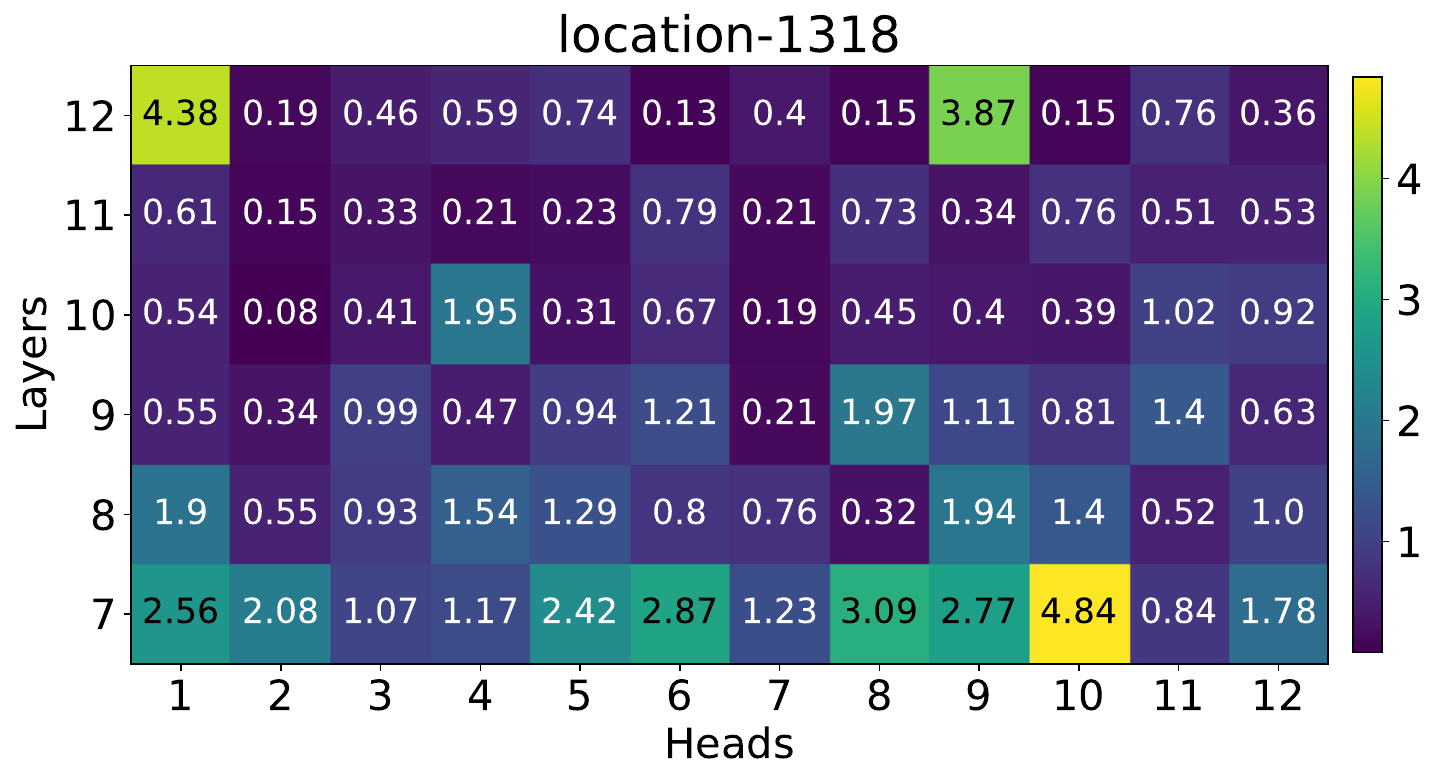}
    \end{subfigure}
    \begin{subfigure}{0.5\linewidth}
      \centering
      \includegraphics[width=\textwidth]{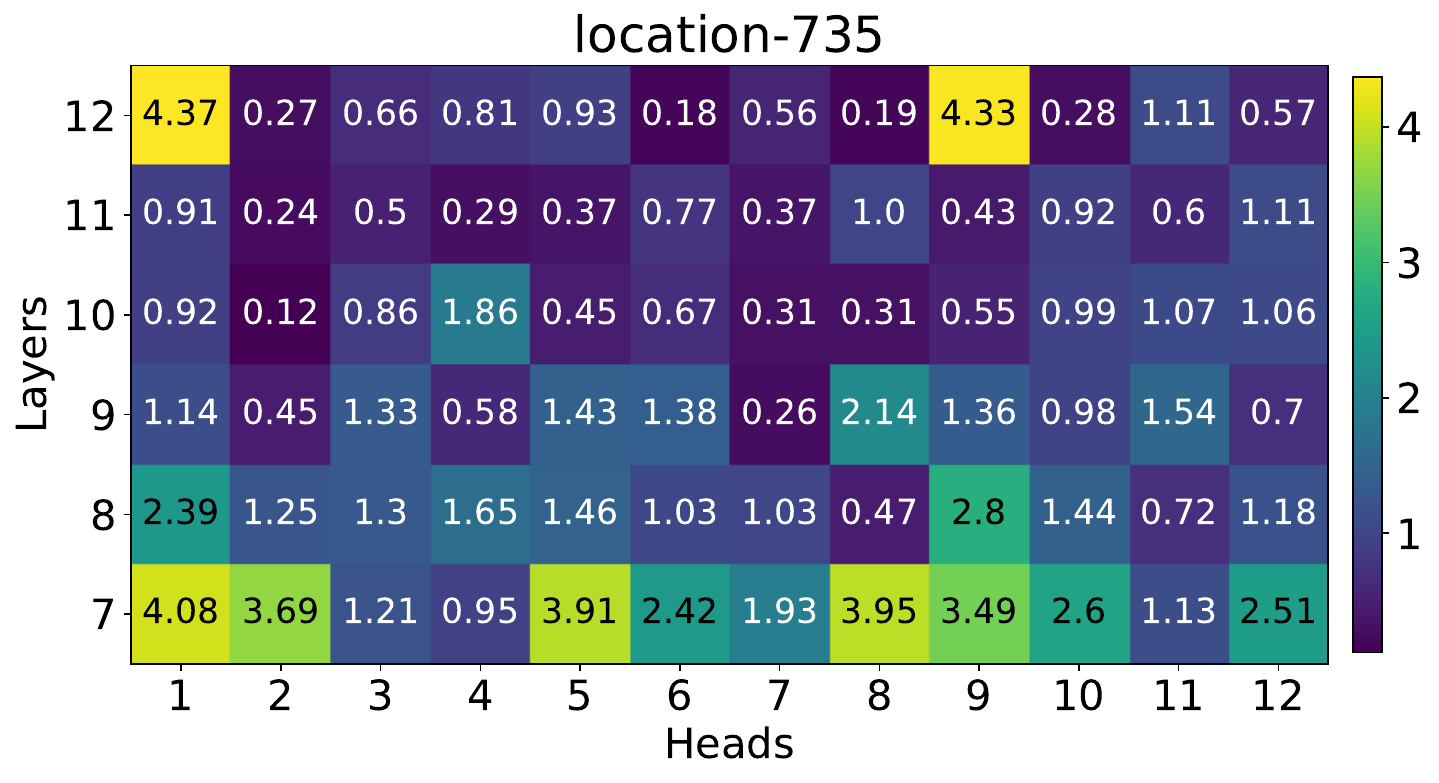}
    \end{subfigure}
    \begin{subfigure}{0.49\linewidth}
      \centering
      \includegraphics[width=\textwidth]{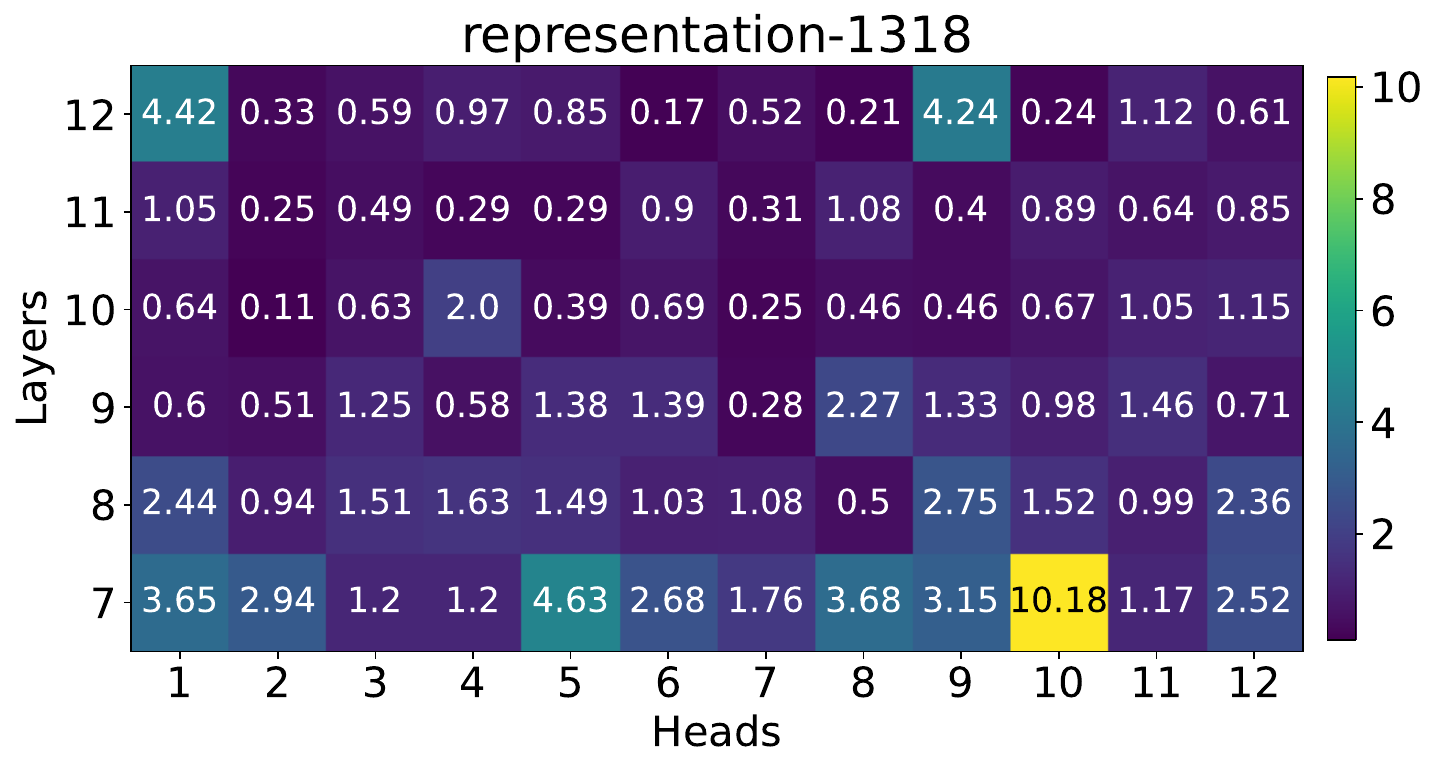}
    \end{subfigure}
    \begin{subfigure}{0.5\linewidth}
      \centering
      \includegraphics[width=\textwidth]{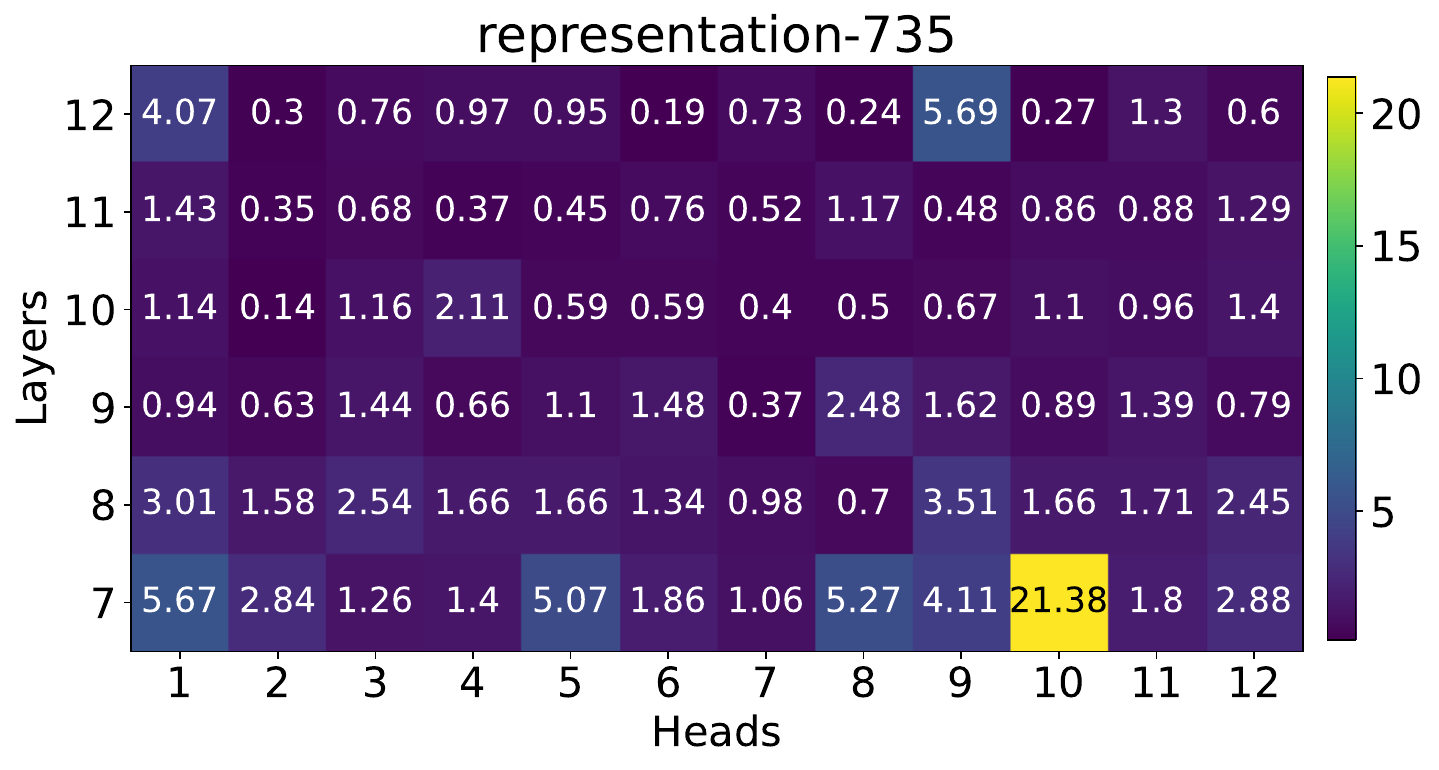}
    \end{subfigure}
    \begin{subfigure}{0.49\linewidth}
      \centering
      \includegraphics[width=\textwidth]{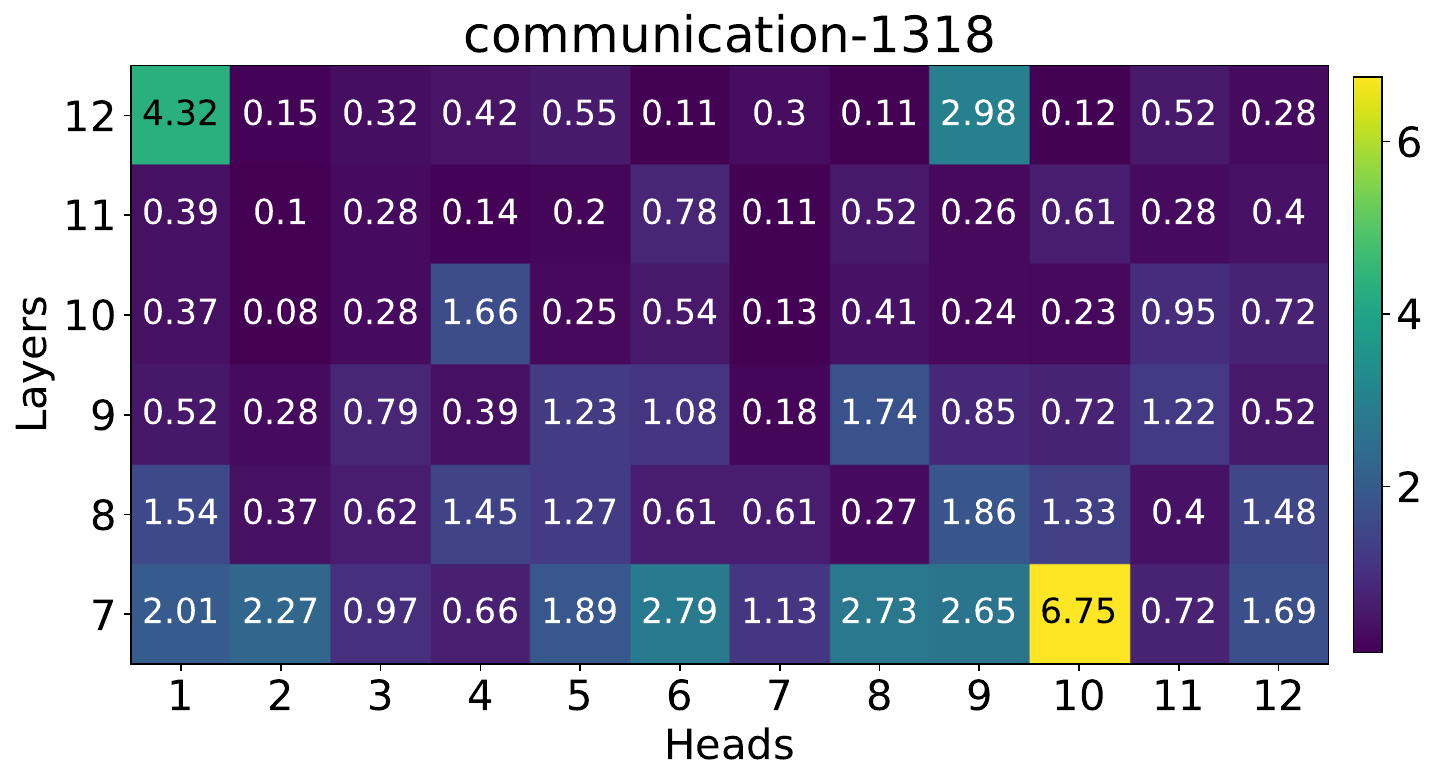}
    \end{subfigure}
    \begin{subfigure}{0.49\linewidth}
      \centering
      \includegraphics[width=\textwidth]{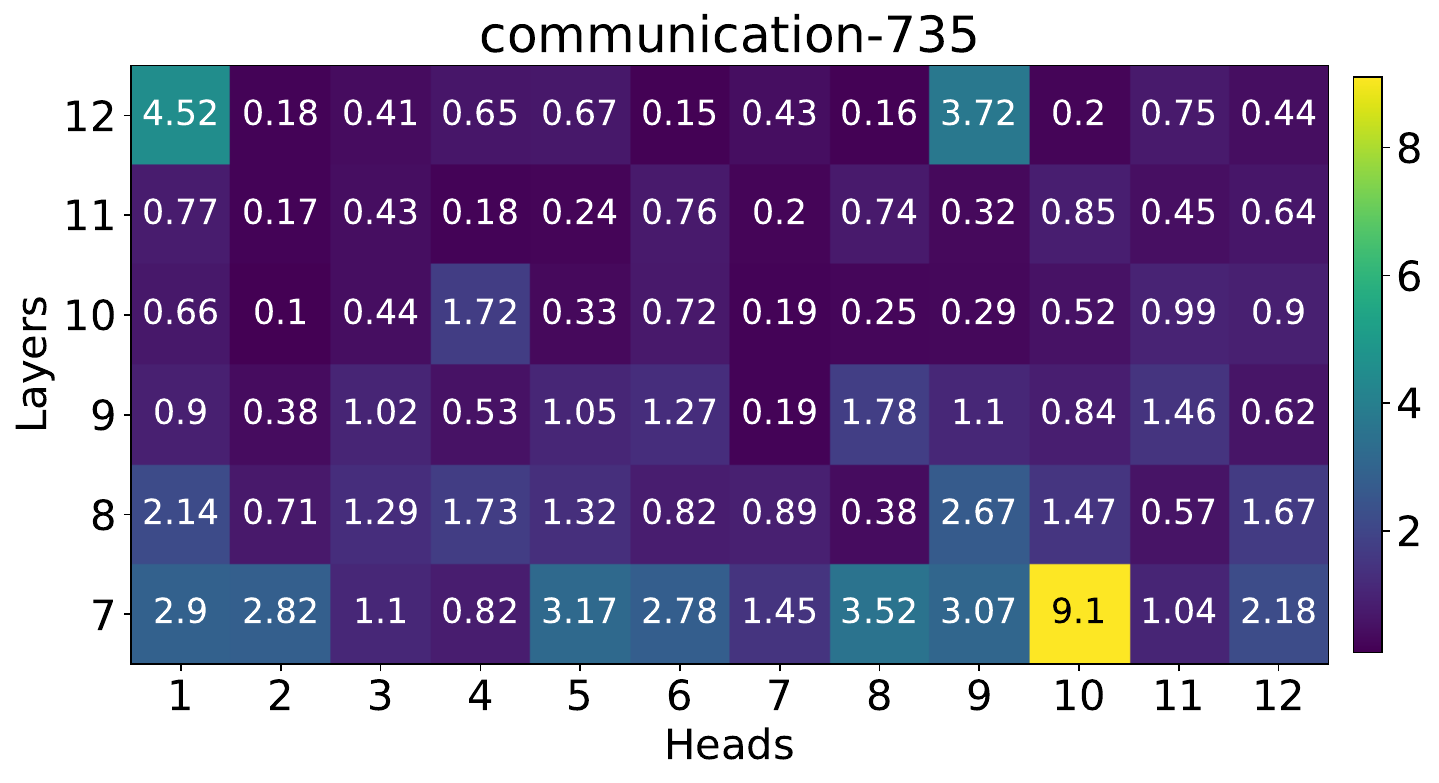}
    \end{subfigure}
    \caption{\textbf{Density percentage (\%) of different heads across different attention layers for weight masking.} Each row represents a different KG and each column is a different seed.}
    \label{fig:structure-head}
\end{figure*}

\begin{figure*}[t!]
    \centering
    \begin{subfigure}{0.48\linewidth}
      \centering
      \includegraphics[height=0.3\textheight]{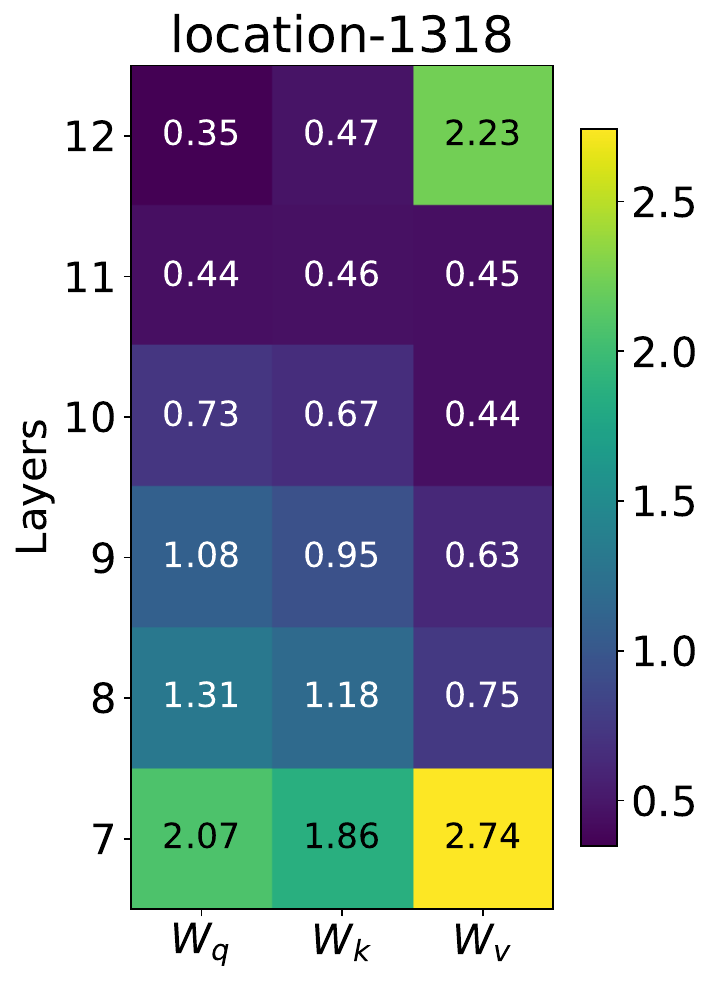}
    \end{subfigure}
    \begin{subfigure}{0.48\linewidth}
      \centering
      \includegraphics[height=0.3\textheight]{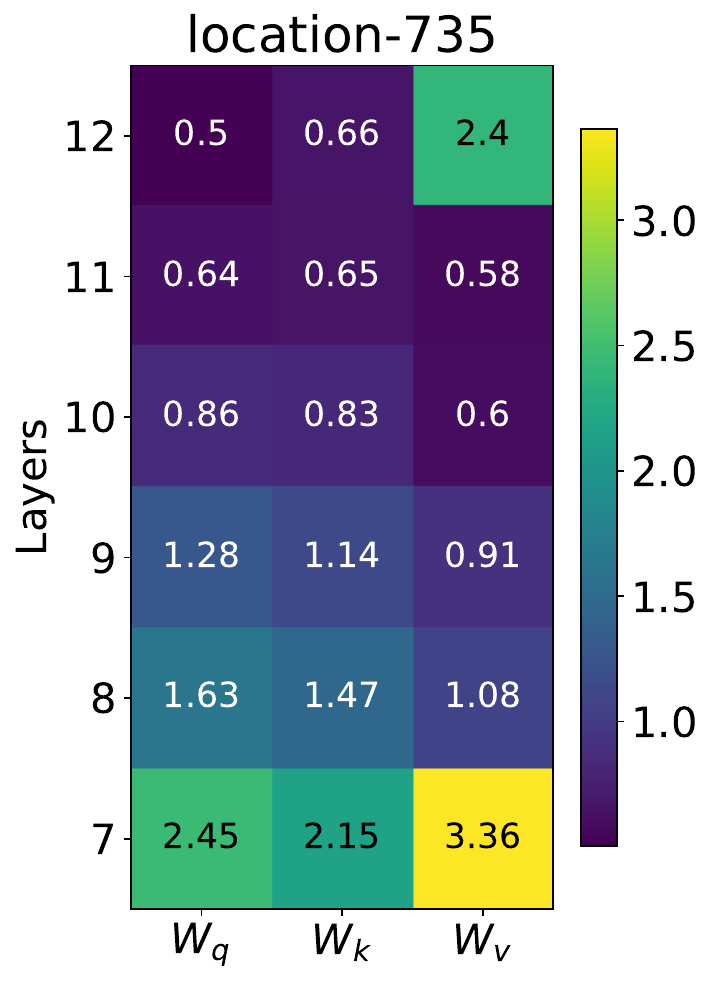}
    \end{subfigure}
    \begin{subfigure}{0.48\linewidth}
      \centering
      \includegraphics[height=0.3\textheight]{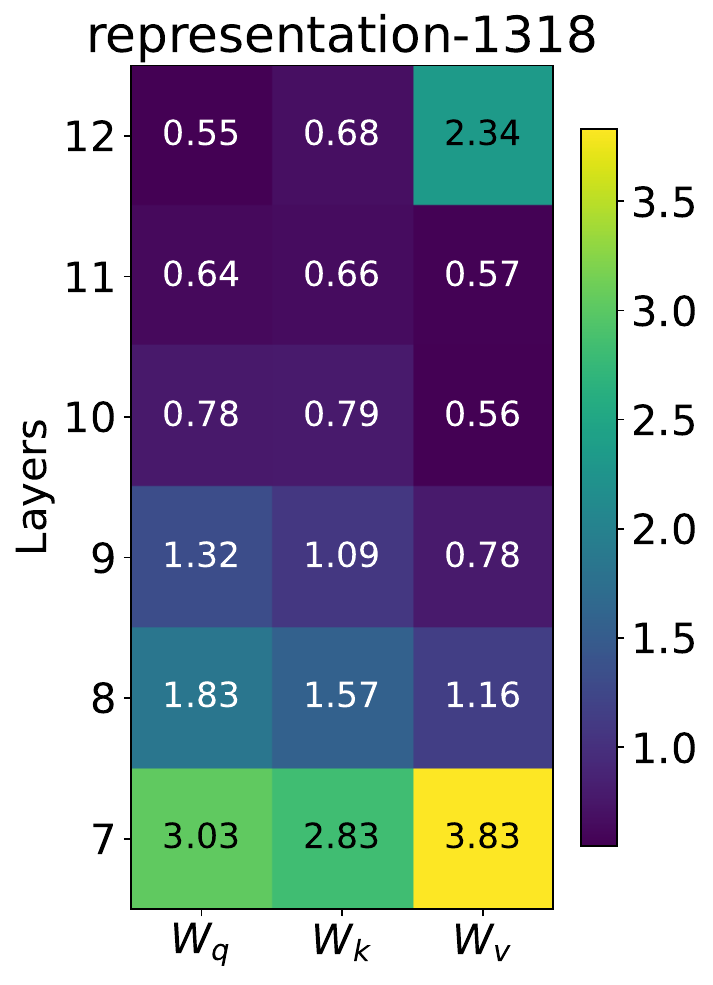}
    \end{subfigure}
    \begin{subfigure}{0.48\linewidth}
      \centering
      \includegraphics[height=0.3\textheight]{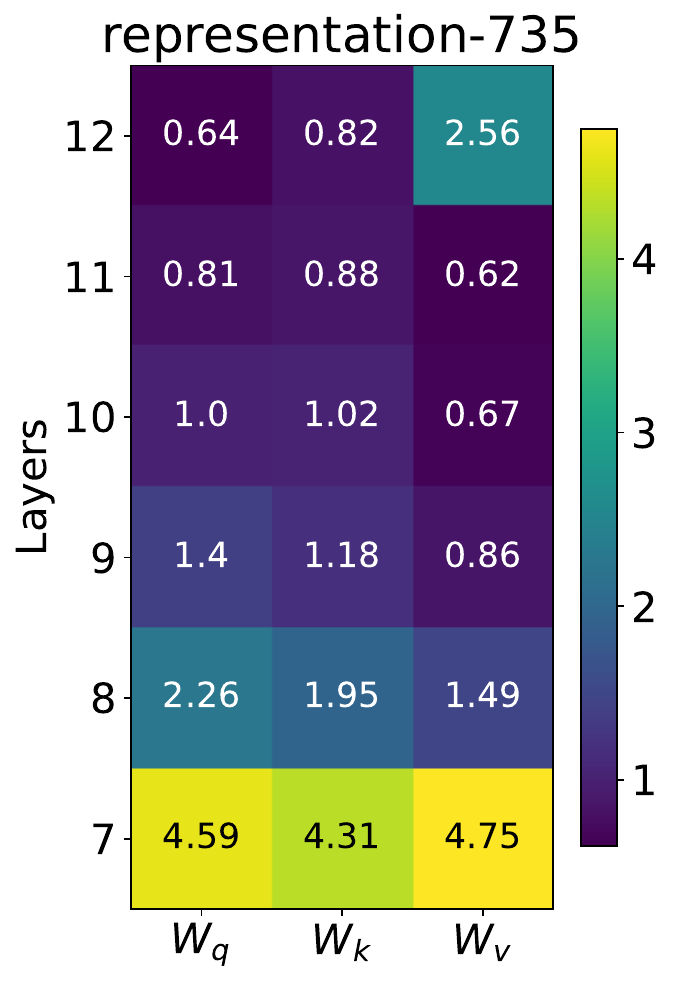}
    \end{subfigure}
    \begin{subfigure}{0.48\linewidth}
      \centering
      \includegraphics[height=0.3\textheight]{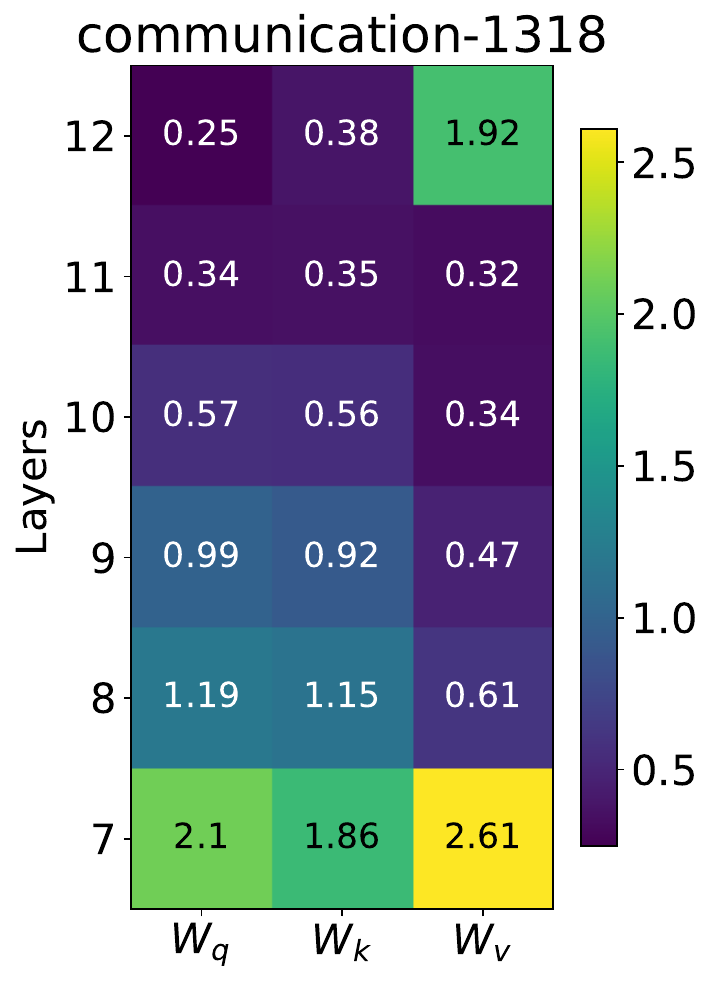}
    \end{subfigure}
    \begin{subfigure}{0.48\linewidth}
      \centering
      \includegraphics[height=0.3\textheight]{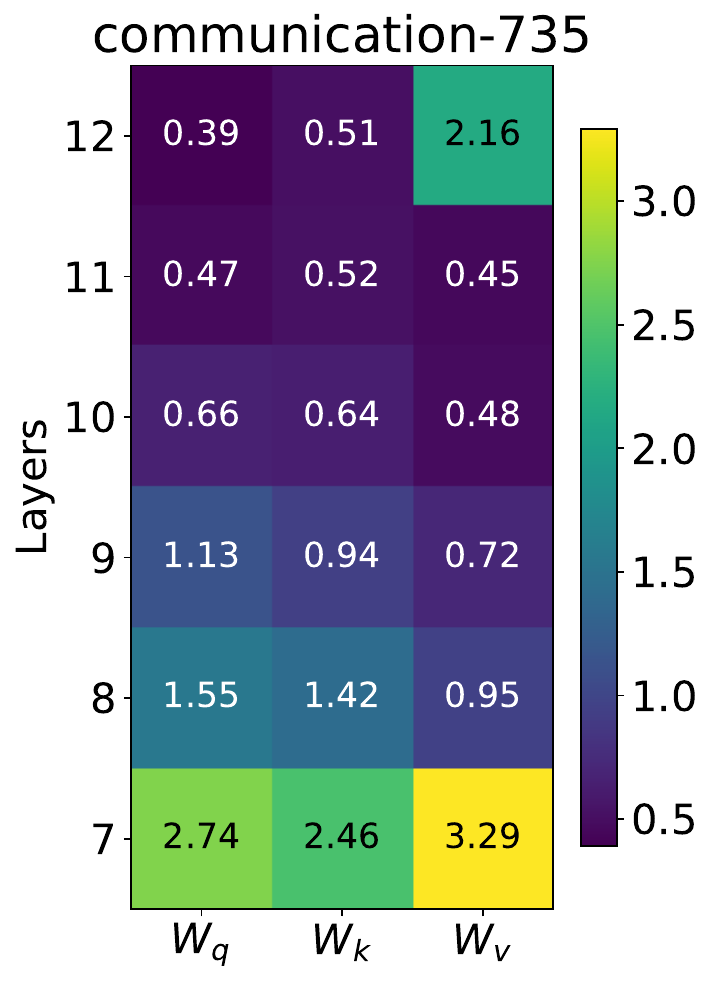}
    \end{subfigure}
    \caption{\textbf{Density percentage (\%) of $W_q$, $W_k$, and $W_v$ masks in attention layers for weight masking.} Each row represents a different KG and each column is a different seed.}
    \label{fig:structure-qkv}
\end{figure*}

In this section, we investigate the structure of the removed knowledge-critical subnetworks by looking at their relative density across different layer types (Figure~\ref{fig:structure-layer}), and more specifically, across different attention heads (Figure~\ref{fig:structure-head}) and the $W_q$, $W_k$, and $W_v$ matrices in attention sublayers (Figure~\ref{fig:structure-qkv}. The density is calculated relatively, meaning according to the particular sublayer's size. The model used is GPT2-small.

Layer depth-wise, we observe that the subnetwork is consistently most dense around the first and final masked transformer blocks, which are layers 7 and 12 in Figure~\ref{fig:structure-layer}. Specifically, layer type-wise, we find that knowledge-critical subnetworks are most dense in the attention sublayers for layer 7 and layer 12 (\texttt{Attn-Out} and \texttt{Attn-$W_q,W_k,W_v$}).

In addition, we have not found any complete columns or rows that were dense in the critical subnetworks. This means no input or output neuron features get completely removed when the critical subnetwork is removed. Therefore, the masked region may not be working to zero-out the knowledge by turning specific features off, which would counter the prevailing view that neuron-level changes are necessary for mechanistic interventions \cite{dai-etal-2022-knowledge, meng2022locating}. 

When we investigated attention heads and $W_q$, $W_k$, and $W_v$ masks in detail for 3 KGs and 3 seeds, we found that head 10 in layer 7, and heads 1 and 9 in layer 12 are significantly dense. Moreover, the $W_v$ mask is consistently the most dense across the three attention $W_q$, $W_k$, and $W_v$ masks. Therefore, while the subnetworks do not have a significant IoU, as demonstrated by the seed-based  (Appendix~\ref{sec:seed-analysis}) and the KG-based analyses (Appendix~\ref{sec:kg-analysis}), the subnetworks still tend to be dense in similar layer types at similar layer depths.

\section{Random Seed-Based Analysis}
\label{sec:seed-analysis}
We investigate the stability of subnetwork discovery under random seed variance for GPT2-small. We also explore whether composing subnetworks from different seeds could increase the suppression effect while still fulfilling the rest of the success criteria.

\begin{figure*}
    \centering
    \includegraphics[height=0.31\textheight]{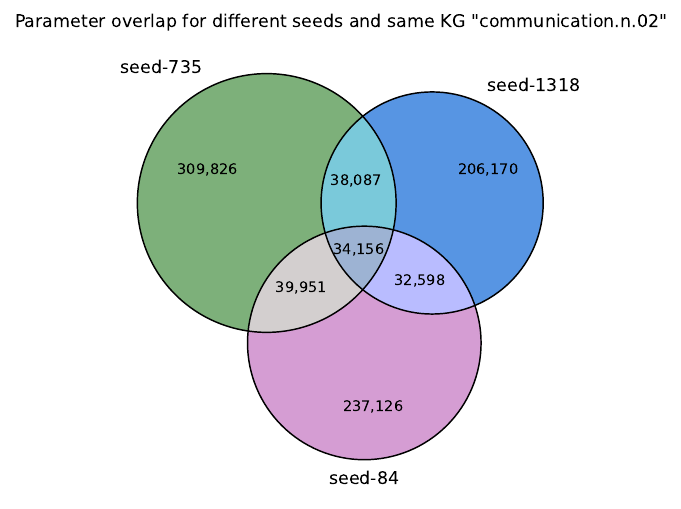}
    \includegraphics[height=0.31\textheight]{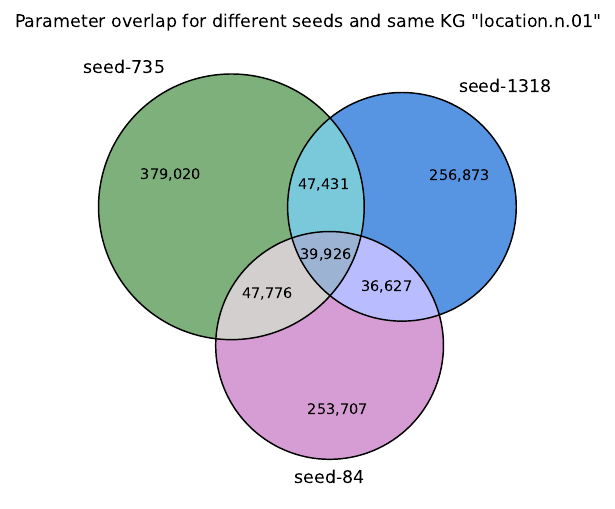}
    \includegraphics[height=0.31\textheight]{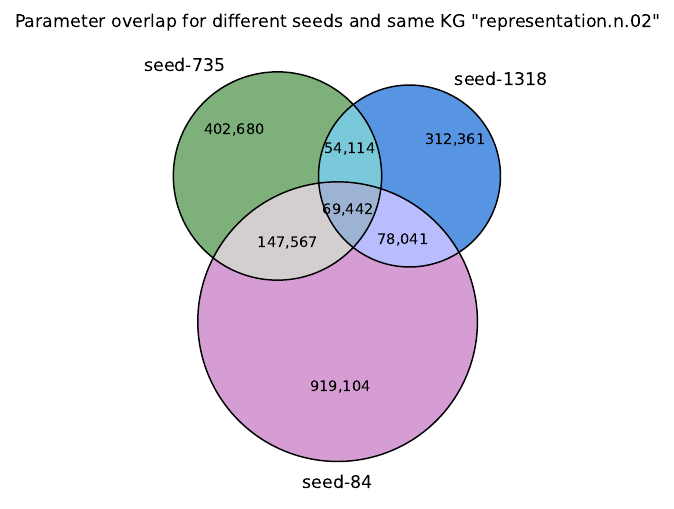}
    \caption{\textbf{Venn diagrams for parameter overlap of three subnetworks identified under three different random seeds with weight masking,} for each KG \texttt{representation}, \texttt{location}, and \texttt{communication}.}
    \label{fig:venn-seed}
\end{figure*}

\begin{figure*}[t]
    \centering
    \includegraphics[height=0.93\textheight]{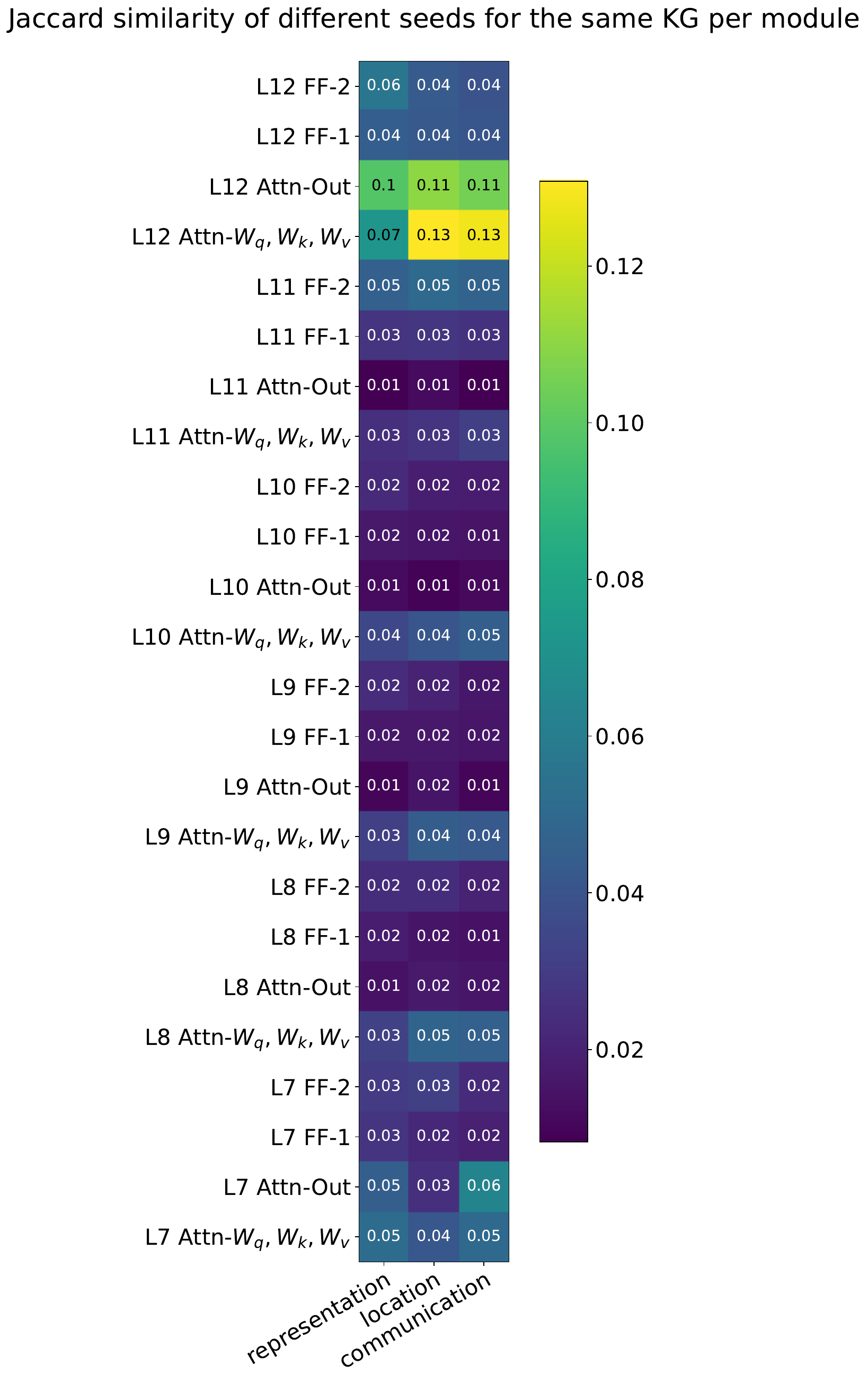}
    \caption{\textbf{Jaccard similarity of different seed masks for the same KG with weight masking,} (\texttt{representation}, \texttt{location}, and \texttt{communication}). The brighter the color, the higher the Intersection over Union.}
    \label{fig:layer-jaccard}
\end{figure*}

\paragraph{Seed-based Variance}
Prior work shows that subnetworks identified under distinct random seeds may differ with a large variance \cite{csordas-2021-modular}. We inspect how subnetworks from the best checkpoints for three random seeds overlap for an individual \targetkg{}. We use Jaccard similarity, or intersection over union (IoU), as the overlap metric. In Figure~\ref{fig:venn-seed}, we plot a Venn diagram of parameter overlap for each knowledge graph. We can see that, on average, when using IoU, only around 3.7\% of the unioned subnetwork parameters overlap across the three seeds (3.76\% for \texttt{location}, 3.8\% for \texttt{communication}, and 3.5\% for \texttt{representation}), meaning the subnetworks identified under different random seeds vary, which complies with prior works' analysis. Across layers, the IoU is also similarly low with a higher overlap for the final attention layer masks ($\approx$10\%) as shown in Figure~\ref{fig:layer-jaccard}.

\paragraph{Subnetwork Composition}
We combine masks of three seeds in their intersection, their floral intersection (intersection unioned with each intersection of two seeds), and overall union to measure the effect on \deltappl{} for \targetkg{}, \controlkg{}, and \lmodeling{}. We average the results over three KGs (\texttt{representation}, \texttt{location}, and \texttt{communication}).

In Table~\ref{tab:seed-combo}, we observe that removing the intersection and floral intersection of the subnetworks does not increase \targetkg{} \deltappl{}. On the other hand, removing the union of the subnetworks increases the \targetkg{} perplexity difference significantly larger than the original results. However, combining the subnetworks and removing them increases \deltappl{} on maintenance datasets more than using an individual seed's subnetwork, as seen in the original results. We note that the increase in the \deltappl{} on maintenance datasets matches the increase we get when removing an equally sparse random subnetwork (see Table~\ref{tab:success-criteria}). Therefore, it may be possible to naively combine subnetworks; however, they may not guarantee the maintenance criteria to the same extent. A future idea could be to continue optimizing for the subnetwork mask by initializing it as the union of the subnetworks to see if more robust suppression can be achieved.

\section{Knowledge-Based Analysis}
\label{sec:kg-analysis}

This section examines the overlap of subnetworks across different KGs for the same seed with GPT2-small. This contrasts with the previous section that studies the overlap of subnetworks across different seeds for the same KG. Similarly, we use Jaccard similarity, or intersection over union (IoU), as the overlap metric. We also explore whether composing subnetworks for different KGs from the same seed could suppress all of the \targetkg{}s.

\begin{figure*}
    \centering
    \includegraphics[height=0.31\textheight]{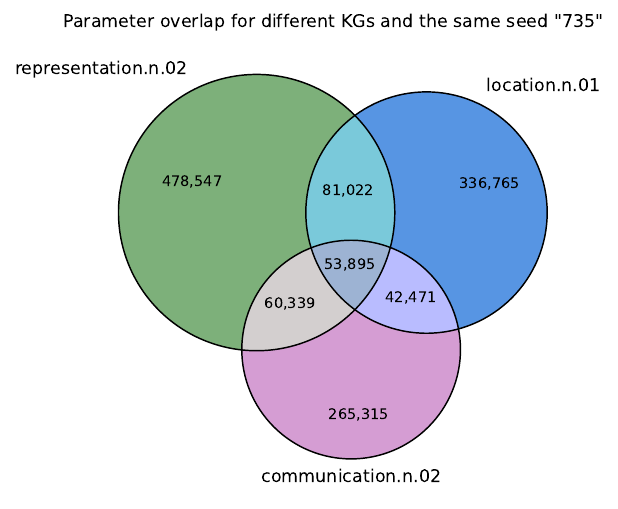}
    \includegraphics[height=0.31\textheight]{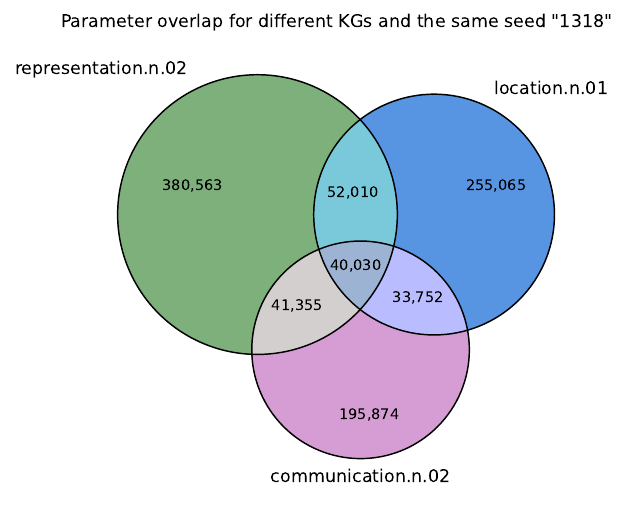}
    \includegraphics[height=0.31\textheight]{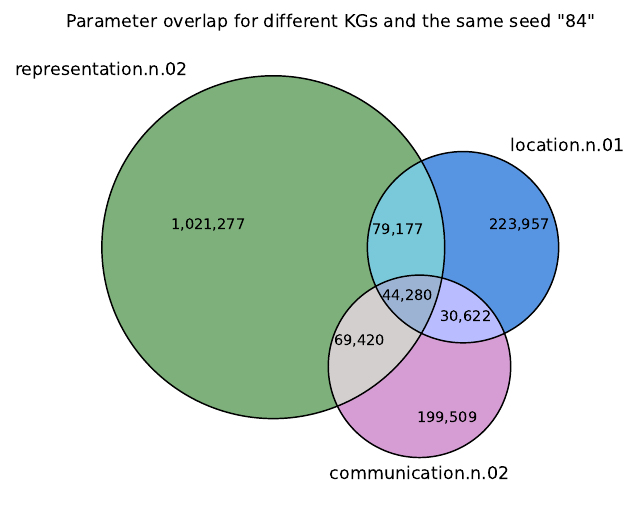}
    \caption{\textbf{Venn diagrams for parameter overlap of three subnetworks identified under three different KGs with weight masking,} for each seed \texttt{735}, \texttt{1318}, and \texttt{84}.}
    \label{fig:venn-kg}
\end{figure*}

\begin{figure*}[t]
    \centering
    \includegraphics[height=0.93\textheight]{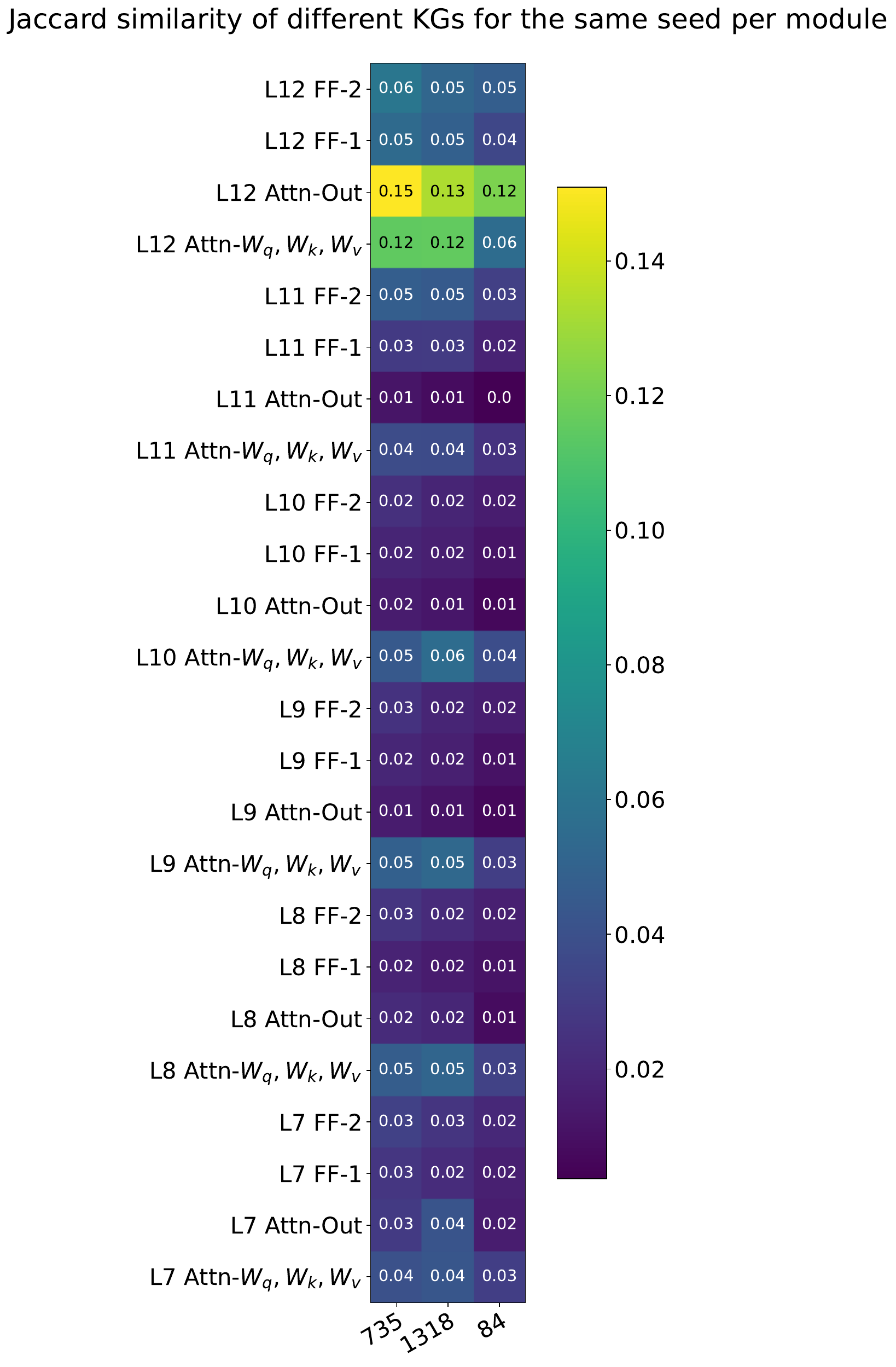}
    \caption{\textbf{Jaccard similarity of different KG masks for the same seed with weight masking,} (\texttt{735}, \texttt{1318}, and \texttt{84}). The brighter the color, the higher the Intersection over Union.}
    \label{fig:jaccard-kg}
\end{figure*}

\paragraph{Knowledge-based Variance}
In Figure~\ref{fig:venn-kg}, we plot a Venn diagram of parameter overlap for each seed across different \targetkg{}s. On average, when using IoU, only around 3.56\% of the unioned subnetwork parameters overlap across the three seeds (4.08\% for \texttt{seed 735}, 4.01\% for \texttt{seed 1318}, and 2.65\% for \texttt{seed 84}). Across layers, the IoU is also similarly low with a significantly higher overlap for the final attention layer masks ($\approx$12\%) as shown in Figure~\ref{fig:jaccard-kg}.

\paragraph{Subnetwork Composition}
We combine masks of three KGs for the same seed in their intersection, their floral intersection (intersection unioned with each intersection of two KGs), and overall union to measure the effect on \deltappl{} for \targetkg{}, \controlkg{}, and \lmodeling{}. We average the results over three seeds (\texttt{735}, \texttt{1318}, and \texttt{84}).

Similar to the findings in composing subnetworks for different seeds, Table~\ref{tab:kg-combo} shows that composing subnetworks for different KGs increases the \deltappl{} on \targetkg{} when using their union. However, removing the union of the subnetworks also has higher perplexity differences on maintenance datasets than using an individual KG's subnetwork, as seen in the original results. Once again, this \deltappl{} increase on the maintenance datasets matches the difference we would observe using an equally sparse random subnetwork. Therefore, while subnetworks of different KGs may be composable to fortify the suppression effect, they may not guarantee the maintenance criteria to the same extent as the individual subnetworks.

\begin{table*}
\vspace{-40pt}
\centering
\resizebox{0.7\linewidth}{!}{
    \begin{tabular}{lcccc}
        \toprule
        \textbf{Mask}   & \textbf{Sparsity} & \textbf{\targetkg{}} & \textbf{\controlkg{}} & \textbf{\lmodeling{}}  \\
         \textbf{Pattern}  &   \textbf{($\uparrow$)}   & \textbf{\deltappl{} ($\uparrow$)}  & \textbf{\deltappl{} ($\downarrow$)} & \textbf{\deltappl{} ($\downarrow$)} \\ \midrule
        Original & 98.8 & 379.1 & 0.7 &	0.4 \\ \midrule
        Union & 96.9 & 1984.4  & 44.9 & 4.7 \\
        Floral & 99.5 & 4.9  & 4.5  & 1.1  \\ 
        Intersection & 99.9 & 7.9  & 3.7 & 2.1 \\ \bottomrule
    \end{tabular}
}
\centering
\caption{\textbf{Composing subnetworks across KGs with GPT2-small and weight masking,} averaged across three seeds. Original stands for the individual subnetwork removal average across the same three seeds and KGs.}
\label{tab:kg-combo}
\end{table*}

\begin{figure*}[t]
    \centering
    \includegraphics[height=0.93\textheight]{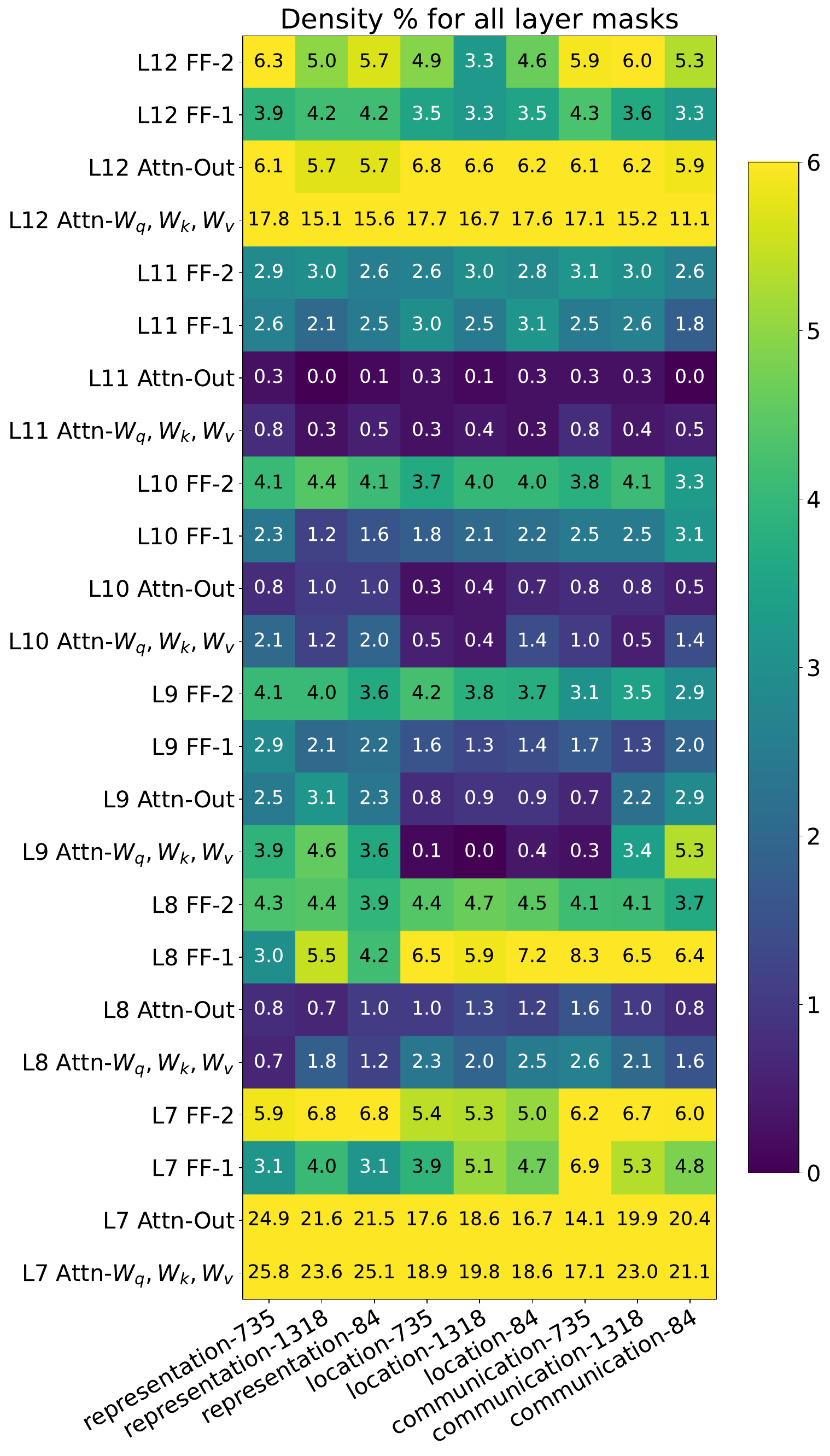}
    \caption{\textbf{Average module mask density with neuron masking,} for different KGs ( \texttt{representation}, \texttt{location}, and \texttt{communication}) and seeds. Reported in percentage (\%). The brighter the color, the higher the removed mask density.}
    \label{fig:structure-layer-input-neuron}
\end{figure*}

\begin{figure*}[t!]
    \centering
    \begin{subfigure}{0.48\linewidth}
      \centering
      \includegraphics[height=0.3\textheight]{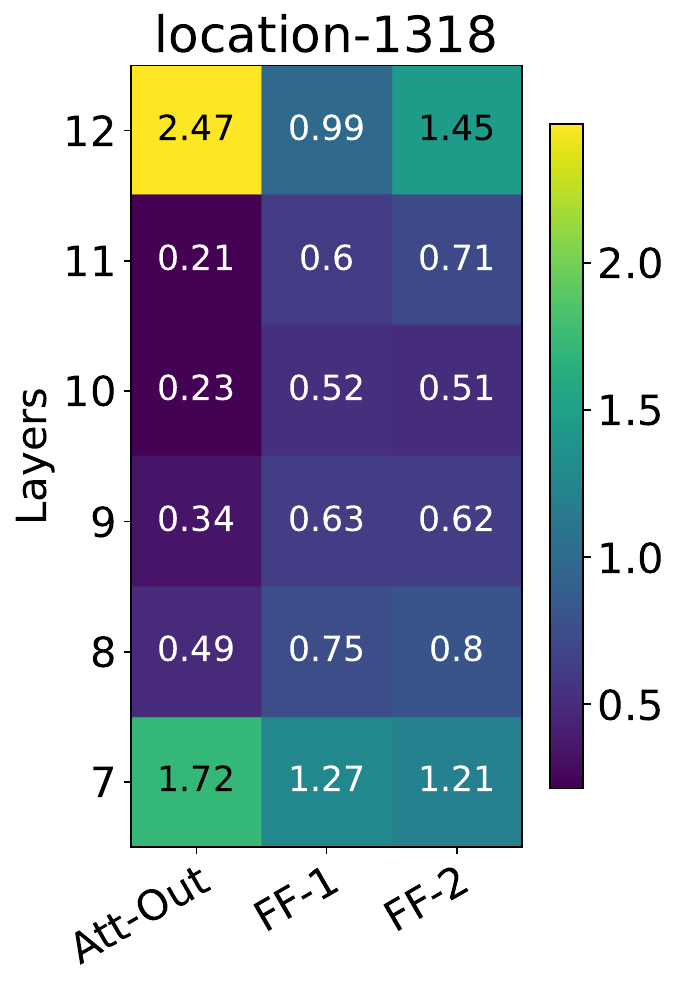}
    \end{subfigure}
    \begin{subfigure}{0.48\linewidth}
      \centering
      \includegraphics[height=0.3\textheight]{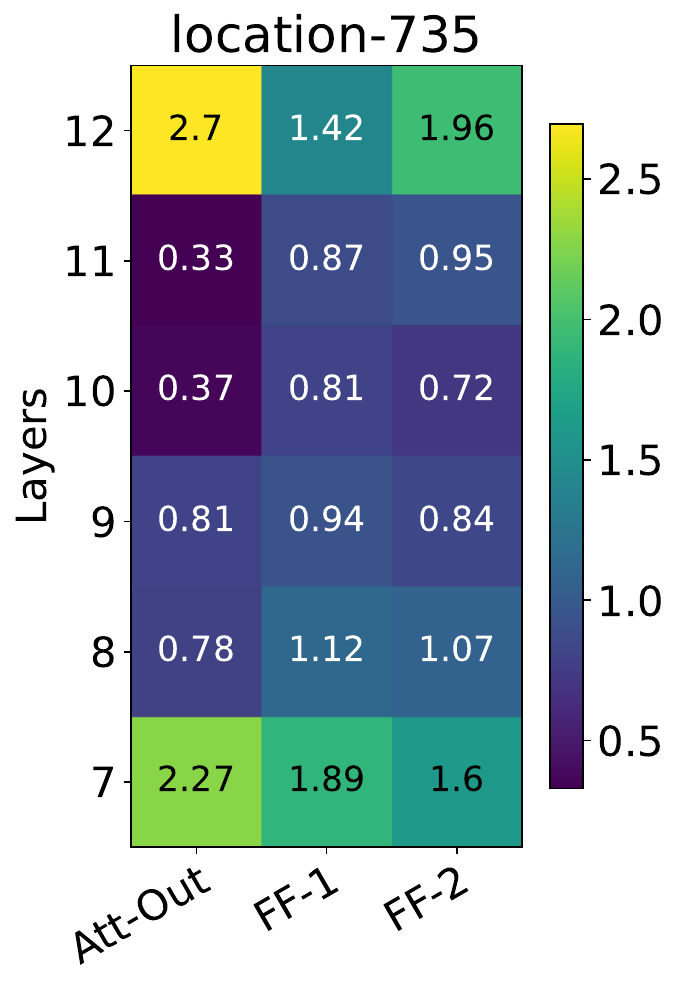}
    \end{subfigure}
    \begin{subfigure}{0.48\linewidth}
      \centering
      \includegraphics[height=0.3\textheight]{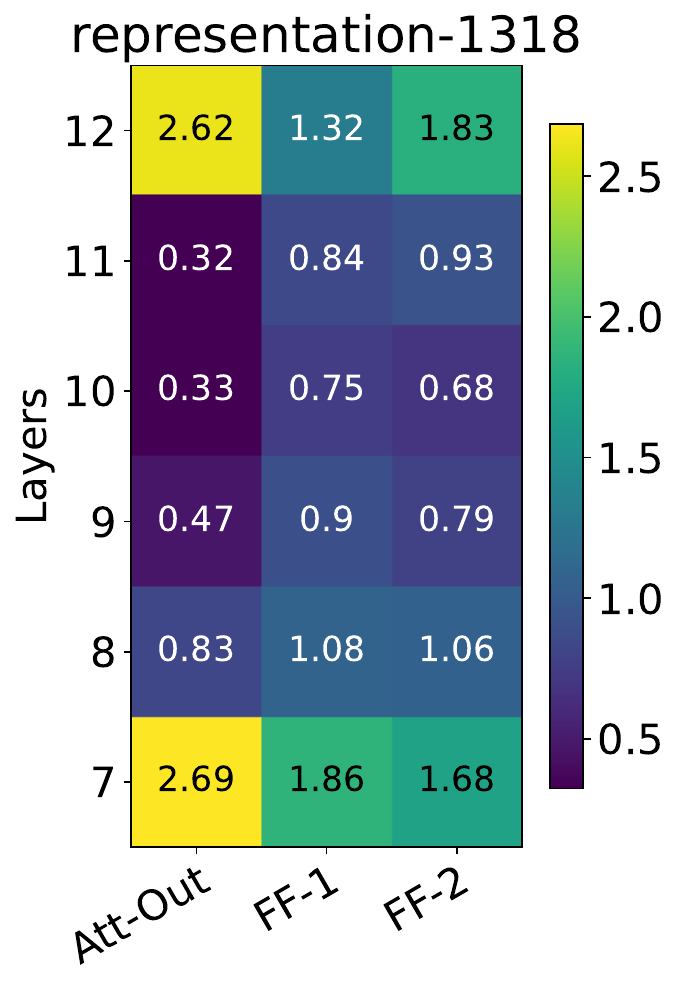}
    \end{subfigure}
    \begin{subfigure}{0.48\linewidth}
      \centering
      \includegraphics[height=0.3\textheight]{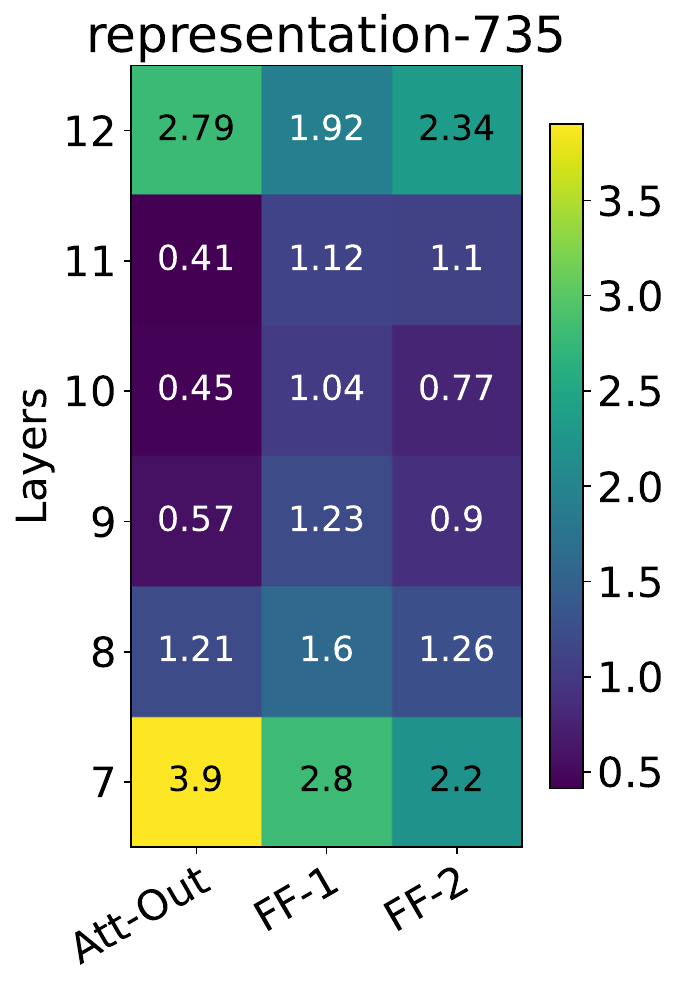}
    \end{subfigure}
    \begin{subfigure}{0.48\linewidth}
      \centering
      \includegraphics[height=0.3\textheight]{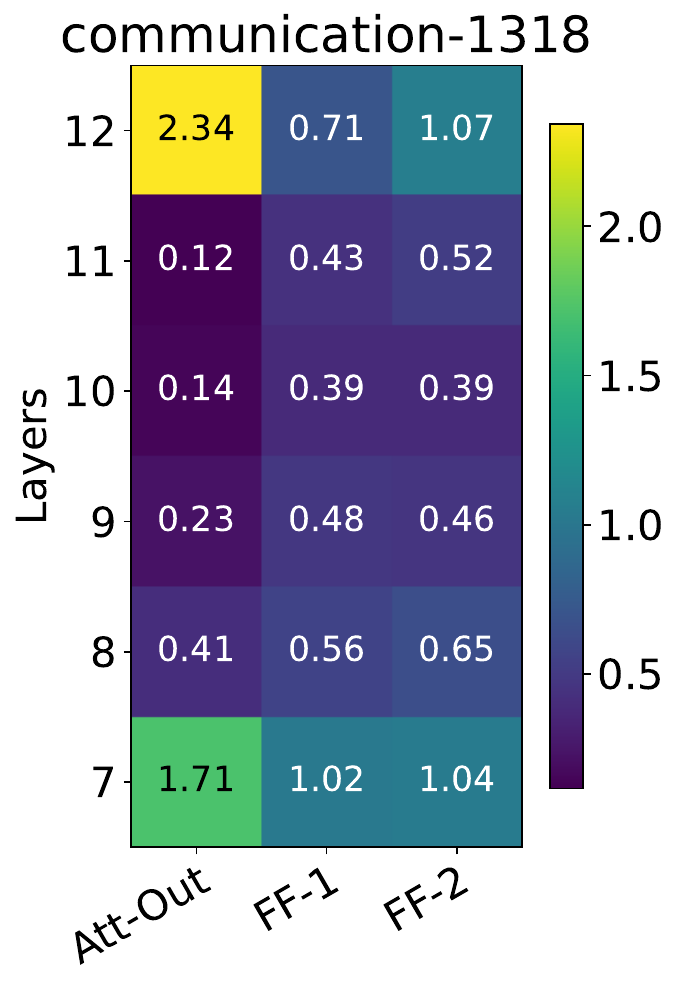}
    \end{subfigure}
    \begin{subfigure}{0.48\linewidth}
      \centering
      \includegraphics[height=0.3\textheight]{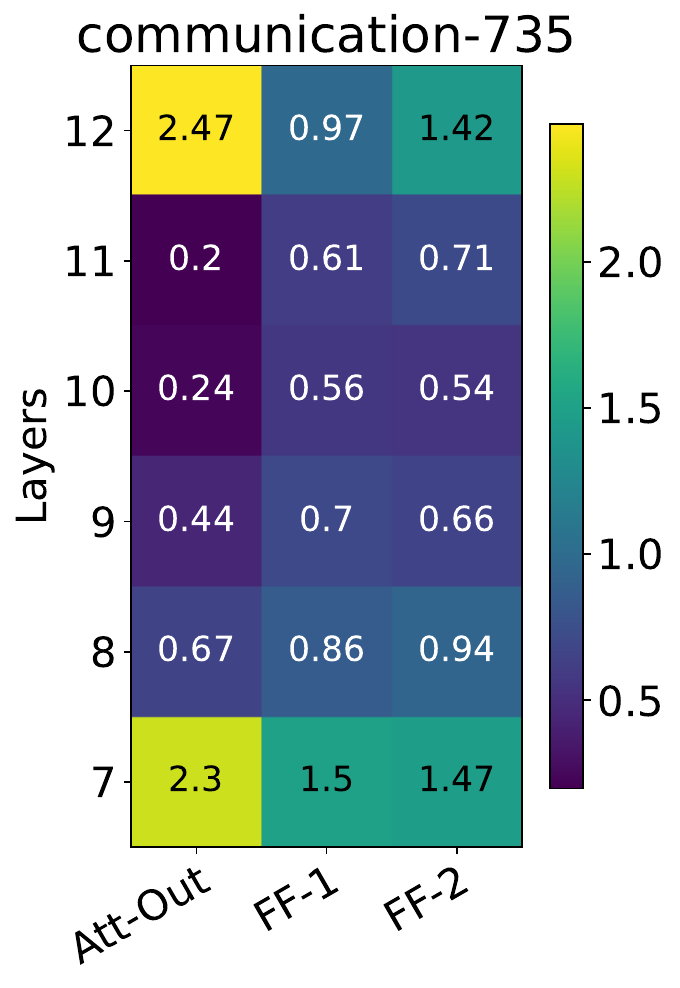}
    \end{subfigure}
    \caption{\textbf{Density percentage (\%) of Att-Out, FF-1, and FF-2 masks for weight masking.} Each row represents a different KG and each column is a different seed.}
    \label{fig:structure-ff}
\end{figure*}

\begin{figure*}
    \centering
    \includegraphics[height=0.31\textheight]{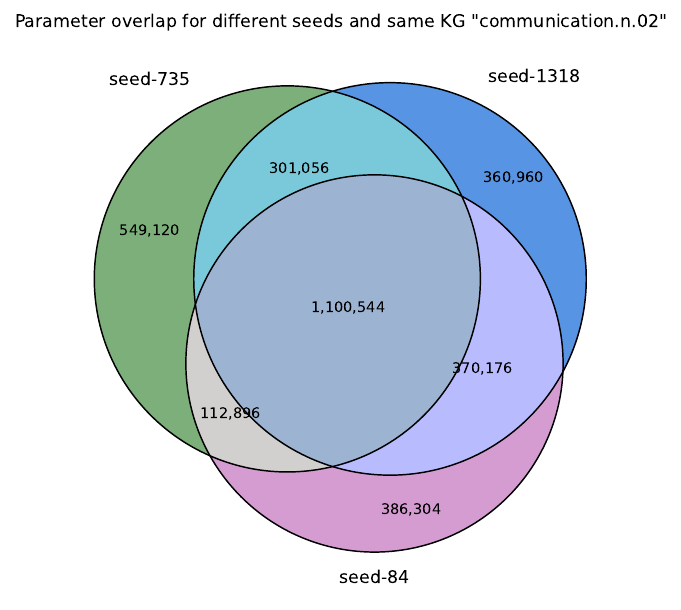}
    \includegraphics[height=0.31\textheight]{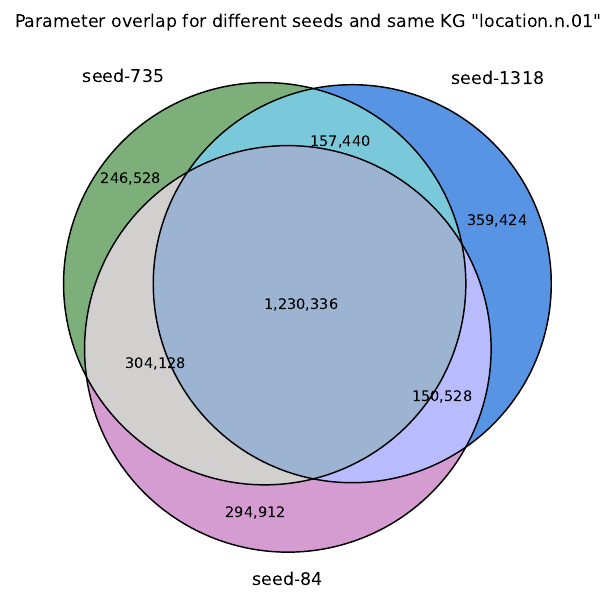}
    \includegraphics[height=0.31\textheight]{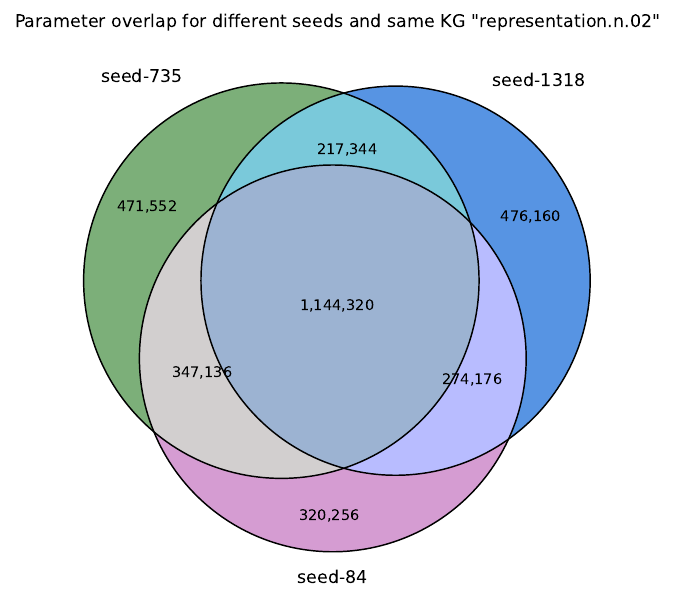}
    \caption{\textbf{Venn diagrams for parameter overlap of three subnetworks identified under three different random seeds with input neuron masking,} for each KG \texttt{representation}, \texttt{location}, and \texttt{communication}.}
    \label{fig:venn-seed-input-neurons}
\end{figure*}

\begin{figure*}[t]
    \centering
    \includegraphics[height=0.93\textheight]{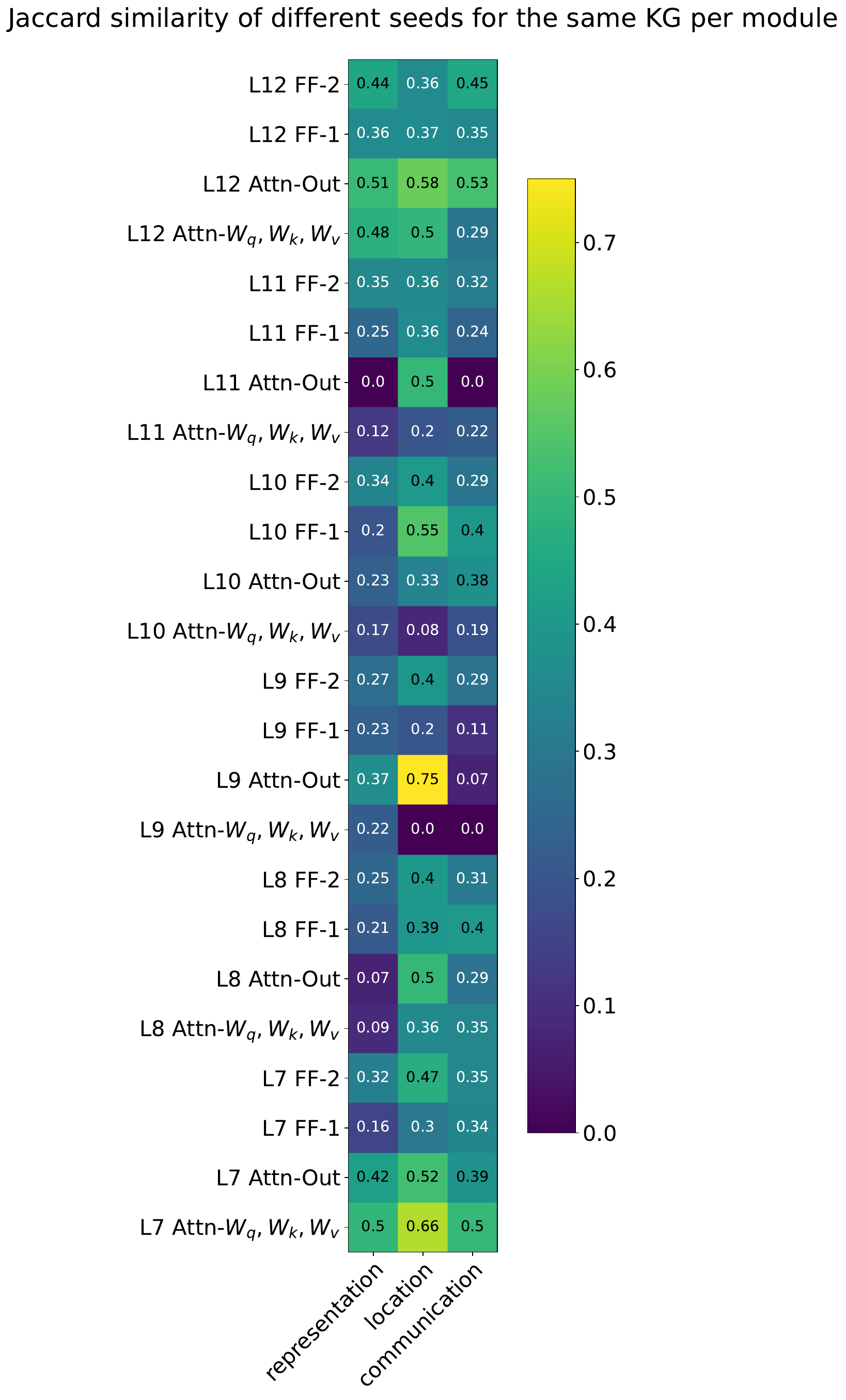}
    \caption{\textbf{Jaccard similarity of different seed masks for the same KG with input neuron masking,} (\texttt{representation}, \texttt{location}, and \texttt{communication}). The brighter the color, the higher the Intersection over Union.}
    \label{fig:layer-jaccard-neuron-inputs}
\end{figure*}

\end{document}